\documentclass[10pt,journal,compsoc]{IEEEtran}

\ifCLASSOPTIONcompsoc

\usepackage{cite}
\usepackage{balance}
\usepackage{subcaption}
\usepackage{booktabs} 
\usepackage{graphicx}
\usepackage{verbatim}
\usepackage{amssymb,amsmath}
\usepackage{gensymb}
\usepackage{mathtools}

\usepackage{algorithmic}
\usepackage{slashbox}

\usepackage{multirow} 
\usepackage{multicol}
\usepackage{rotating} 
\usepackage{epstopdf}
\usepackage{amsthm}
\usepackage{romannum}
\usepackage{gensymb}
\usepackage{tikz}
\usepackage{courier}
\usepackage{microtype}
\usepackage{makecell}
\usepackage{bbm}
\usepackage{bm}
\usepackage[linesnumbered,ruled,vlined]{algorithm2e}
\usepackage{pifont}
\else
  \usepackage{cite}
\fi

\newcommand{\cmark}{\ding{51}}%
\newcommand{\xmark}{\ding{55}}%
\newcommand{\norm}[1]{\left\lVert#1\right\rVert}

\ifCLASSINFOpdf
\else

\fi

\renewcommand{\baselinestretch}{0.982}
\begin{document}

\title{\LARGE Towards Efficient Scheduling of Federated Mobile Devices under Computational and Statistical Heterogeneity}

\author{Cong~Wang, Member, IEEE, Yuanyuan~Yang, Fellow, IEEE and Pengzhan Zhou
\IEEEcompsocitemizethanks{\IEEEcompsocthanksitem Cong Wang is with the Department
of Computer Science, Old Dominion University, Norfolk, VA, USA, E-mail: c1wang@odu.edu
\IEEEcompsocthanksitem Yuanyuan Yang and Pengzhan Zhou are with the Department of Electrical and Computer Engineering, Stony Brook University, NY, USA, E-mail: \{yuanyuan.yang, pengzhan.zhou\}@stonybrook.edu
\IEEEcompsocthanksitem Correspondence to c1wang@odu.edu.}}

\IEEEtitleabstractindextext{%
\begin{abstract}
Originated from distributed learning, federated learning enables privacy-preserved collaboration on a new abstracted level by sharing the model parameters only. While the current research mainly focuses on optimizing learning algorithms and minimizing communication overhead left by distributed learning, there is still a considerable gap when it comes to the real implementation on mobile devices. In this paper, we start with an empirical experiment to demonstrate computation heterogeneity is a more pronounced bottleneck than communication on the current generation of battery-powered mobile devices, and the existing methods are haunted by mobile stragglers. Further, non-identically distributed data across the mobile users makes the selection of participants critical to the accuracy and convergence. To tackle the computational and statistical heterogeneity, we utilize data as a tuning knob and propose two efficient polynomial-time algorithms to schedule different workloads on various mobile devices, when data is identically or non-identically distributed. For identically distributed data, we combine partitioning and linear bottleneck assignment to achieve near-optimal training time without accuracy loss. For non-identically distributed data, we convert it into an average cost minimization problem and propose a greedy algorithm to find a reasonable balance between computation time and accuracy. We also establish an offline profiler to quantify the runtime behavior of different devices, which serves as the input to the scheduling algorithms. We conduct extensive experiments on a mobile testbed with two datasets and up to 20 devices. Compared with the common benchmarks, the proposed algorithms achieve 2-100$\times$ speedup epoch-wise, 2-7\% accuracy gain and boost the convergence rate by more than 100\% on CIFAR10.
\end{abstract}

}

\maketitle

\IEEEdisplaynontitleabstractindextext

%
\IEEEpeerreviewmaketitle


\section{Introduction} \label{sec:intro}
The tremendous success of machine learning stimulates a new wave of smart applications. Despite the great convenience, these applications consume massive personal data, at the expense of our privacy. The growing concerns of privacy become one of the major impetus to shift computation from the centralized cloud to users' end devices such as mobile, edge and IoTs. The current solution supports running on-device inference from a pre-trained model in near real-time~\cite{zhang,lane}, whereas their capability to adapt to the new data and learn from each other is still limited.

\emph{Federated Learning} (FL) emerges as a practical and cost-effective solution to mitigate the risk of privacy leakage~\cite{fedavg,arm-noniid,fed-distillation,sysml,konecny,cmfl-icdcs,decentralize_liu,evolution,multi-task-nips}, which enables collaboration on an abstracted level rather than the raw data itself. Originated from distributed learning~\cite{dean}, it learns a centralized model where the training data is held privately at the end users. Local models are computed in parallel and the updates are aggregated towards a centralized \emph{parameter server}. The server takes the mean of the parameters from the users, pushes the averaged model back as the initial point for the next iteration. Previous research mainly focuses on the prominent problems left from distributed learning such as improving communication efficiency~\cite{konecny,cmfl-icdcs,decentralize_liu,evolution,multi-task-nips,tiered}, security robustness~\cite{bonawitz,byzantine}, or the learning algorithms~\cite{multi-task-nips} to address new challenges in FL.

Most of these works only contain proof-of-concept implementations on cloud/edge servers with stable, external power and proprietary GPUs. Though pioneering efforts to improve FL at the algorithm level, they still leave a gap to the system-level implementations at the mobile data source, where FL was originally targeting at. Meanwhile, the dramatic increase of mobile processing power has enabled not only on-device inference, but also moderate training (backpropagation)~\cite{sysml,zhao-edge,acm-mm}, thus providing a basis to launch FL on battery-powered mobile devices.

Unfortunately, the vast heterogeneity of mobile processing power has yet to be addressed. Even worse, the high variance among user data adds another layer of statistical heterogeneity~\cite{multi-task-nips} and makes the selection of participants a nontrivial problem. An inappropriate selection would adversely cause gradient divergence and diminish every effort to reduce computation time. From an empirical study, we first validate that the bottleneck has actually shifted from communication back to computation on consumer mobile devices. The runtime behavior depends on a complex combination of vendor-specific software implementations and the underlying hardware architecture, i.e., the System on a Chip (SoC), power management policies and the input computation intensity of the neural networks. These challenges are magnified in practices when users behave differently. For example, in activity recognition, some users may perform only a few actions (e.g., sitting for a long time), thus leading to highly skewed local distributions, which breaks the independent and identically distributed (IID) assumptions held as a basis for distributed learning. When averaged into the global model, these skewed gradients may have a damaging impact on the collaborative model. Thus, efficient scheduling of FL tasks entails an optimal selection of participants relying on both computation and data distribution.

To tackle these challenges, we propose an optimization framework to schedule training using the workloads (amount of training data) as a tuning knob and achieve near-optimal staleness in synchronous gradient aggregation. We build a performance profiler to characterize the relations between training time and data size/model parameters using a multiple linear regressor. Taking the profiles as inputs, we start with the basic case when data is IID (class-balanced), and formulate the problem into a min-max optimization problem to find optimal partitioning of data that achieves the minimum makespan. We propose an efficient $\mathcal{O}(n^2 \log n)$ algorithm~\cite{gbap} with $\mathcal{O}(n)$ analytical solution when the cost function is linear ($n$ is the number of users). For non-IID data, we introduce a new accuracy cost and a quantitative derivation from the analysis of gradient diversity~\cite{gradient-diversity}. Then we re-formulate the problem into a min average cost problem and develop a greedy $\mathcal{O}(mn)$-algorithm to assign workloads with the minimum average cost in each step using a variation of bin packing~\cite{bpf1} ($m$ is the number of data shards). The observation suggests a nontrivial trade-off between staleness and convergence. The proposed algorithm aims to leverage users' class distributions to adaptively include/exclude an outlier in order to improve model generalization and convergence speed without sacrificing the epoch-wise training time too much. Finally, the proposed algorithms are evaluated on MNIST and CIFAR10 datasets with a mobile testbed of various smartphone models.

The main contributions are summarized below. First, we motivate the design by a series of empirical studies of launching backpropagation on Android. This expands the current research of FL with new discoveries of the fundamental cause of mobile stragglers, as well as offering an explanation to the subtlety in non-IID outliers through the lens of gradient diversity. Second, we formulate the problem to find the optimal scheduling with both IID and non-IID data, and propose polynomial-time algorithms with analytical solutions when the cost profile is linear. Finally, we conduct extensive evaluations on MNIST and CIFAR10 datasets under a mobile testbed with 6 types of device combinations (up to 20 devices). Compared to the benchmarks, the results show 2-100$\times$ speedups for both IID/non-IID data while boosting the accuracy by 2-7\% on MNIST/CIFAR10 for non-IID data. The algorithms demonstrate advantages of avoiding worst-case stragglers and better utilization of the parallelled resources. In contrast to the existing works that decouple learning from system-level implementations, to the best of our knowledge, this is the first work that not only connects them, but also optimizes the overall system performance.

The rest of the paper is organized as follows. Section \ref{sec:background} discusses the related works and background. Section \ref{sec:computation} and \ref{sec:distribution} motivate this work with a series of empirical studies. Sections \ref{sec:design_iid} and \ref{sec:non_iid} optimizes training time for IID and non-IID data. Section \ref{sec:profiling} describes the profiler. Section \ref{sec:eval} evaluates the framework on the mobile testbed and dataset. Section \ref{sec:discussion} discusses the limitation and Section \ref{sec:conclusion} concludes this work.


\section{Background and Related Works} \label{sec:background}

\subsection{Deep Learning on Mobile Devices}

The continuous advance in mobile processing power, battery life and improvement of power management reaches its culminating point with the debut of AI chips~\cite{bionic,kirin980}. Their power spans from executing simple algorithms like logistical regression or supported vector machine, to the resource-intensive deep neural networks. The research community quickly embraces the idea to migrate \emph{inference} computations off the cloud to the mobile device, and develops new applications for better user interaction and experience~\cite{zhang,lane}. To fit the neural network within the memory capacity, some optimization is necessary such as pruning the near-zero parameters via compression~\cite{han} or directly learning a sparse network before deployment~\cite{lottery}. Recently, there are new efforts to incorporate the entire \emph{training} process on mobile devices for better privacy preservation and adaptation of the new patterns in data distribution~\cite{acm-mm,zhao-edge}. Their implementation attempts to close the loop of the learning process from data generation/pre-processing to decision making all on user's end devices, which has also laid the foundation of this paper.

The implementation of federated learning on mobile devices is subject to physical limitations from the memory and computation. Unlike desktop or cloud servers that the GPUs have dedicated, high-bandwidth memory, the memory on consumer mobile devices is extremely limited. The mobile SoC typically has unified memory, where the mobile GPU only has some limited on-chip buffer. Further, on the OS-level, Android has limited memory usage for each application (e.g., setting the \texttt{LargeHeap} would give the application 512 MB). Though using the native code can bypass the limit, it is still subject to the memory limits about 3-4GB on most mobile devices, where a majority is shared by other system processes. This largely constrains us from running ultra-deep models. As shown later, the bottleneck still lies in the computational side due to limited paralleled resources, since most mobile devices have about 8 CPU cores and similar number of GPU shader cores. Thus, it is expected that the computation time would increase parabolically with a more complex neural network model, especially on the lowgrade devices.

The collaboration of mobile devices brings more uncertainties to the system. Vendors typically integrate different IP blocks under the area and thermal constraints, thus resulting a highly fragmented mobile hardware market: 1) unlike the ubiquitous of CUDA to accelerate cloud GPUs, programming support is inadequate for mobile GPUs; 2) most mobile GPUs have similar computational capability compared to the multi-core CPUs~\cite{facebook-hpca}. Thus, for better programming support, we execute learning by the multi-core CPUs in this paper.

Like any other applications in the userspace, the learning process is handled by the Linux kernel of Android, which controls \texttt{cpufreq} by the CPU governor in response to the workload. For example, the default \texttt{interactive} governor in Android 8.0 scales the clockspeed over the course of a timer regarding the workload. For better energy-efficiency and performance, off-the-shelf smartphones are often embedded with asymmetric multiprocessing architectures, i.e., ARM's big.LITTLE~\cite{arm}. For instance, Nexus 6P is powered by octa-core CPUs with four big cores running at 2.0 GHz and four little ones at 1.53 GHz. The design intuition is to handle the bursty nature of user interactions with the device by placing low-intensity tasks on the small cores, and high-intensity tasks on the big cores. However, the behavior of such subsystem facing intensive, sustained workload such as backpropagation remains underexplored. Further, since vendors usually extend over the vanilla task scheduler through proprietary designs of task migration/packing, load tracking and frequency scaling, the same learning task would incur heterogenous processing time depending on the hardware and system-level implementation. Our goal is to mitigate such impact on FL while still using the default governor and scheduler for applications in the userspace.


%
%
%
%
%
%


\subsection{Federated Learning}

McMahan et al. introduce FedAvg that averages aggregated parameters from the local devices and minimizes a global loss~\cite{fedavg}. Formally, for the local loss function $l_k(\cdot) (k \in \{1,2,\cdots,N\})$ of $N$ mobile devices, FL minimizes the global loss $L$ by averaging the local weights $\bm{w}$,
\begin{equation}
\small
\min_{\bm{w}} \bigl\{L(\bm{w}) = \sum_{k=1}^{N} l_k(\bm{w}) \bigr\}.  \label{fl_main_eq}
\end{equation}
In each round, mobile device $k$ performs a number of $E$ local updates,
\begin{equation}
\small
\bm{w}_{i+1}^k = \bm{w}_i^k - \eta_i \nabla l_k(\bm{w}_i^k),  \label{local_update_eq}
\end{equation}
with learning rate $\eta_i$ and $i = \{0,1,\cdots,E\}$. The local updates $\bm{w}_{E}^k$ are aggregated towards the server for averaging, and the server broadcasts the global model to the mobile devices to initiate the next round. From the user's persecutive, one may advocate designs without much central coordination. As most of the FL approaches pursue the first-order synchronous approach~\cite{fedavg,arm-noniid,fed-distillation, sysml,konecny,cmfl-icdcs,decentralize_liu}, the starting time of the next round is determined by the \emph{straggler} in the last round, who finishes the last among all the users. Hence, from the service provider's perspective, it is far from efficient due to the straggler problem. It leads to slower convergence and generates a less performing model with low accuracy and ultimately undermines the collaborative efforts from all the users.

A solution is to switch to asynchronous update~\cite{async-icml,eric-xing}. Asynchronous methods allow the faster users to resume computation without waiting for the stragglers. However, inconsistent gradients could easily lead to divergence and amortize the savings in computation time. Though it is possible to estimate the gradient of the stragglers using second-order Taylor expansion, the computation and memory cost of the Hessian matrix become prohibitive for complex models~\cite{async-icml}. In the worst case, gradients from the stragglers that are multiple epoches old could significantly divert model convergence to a local minima. For example, the learning process typically decreases the learning rate to facilitate convergence on the course of training. The stragglers's stale gradients with a large learning rate would exacerbate such divergence~\cite{asplos-thread}. The practical solution from Google is to simply drop out the stragglers in large scale implementation~\cite{sysml}. Yet, for those small-scale federated tasks, e.g., learning from a small set of selected patients with rare disease, such hard drop-out could be detrimental to model generalization and accuracy. Gradient coding~\cite{gradient_coding} replicates copies of data between users so when there is a slow-down, any linear combination from the neighbors can still recover the global gradient. It is suitable for GPU clusters, where all the nodes are authenticated and data can be moved without privacy concerns. Nevertheless, sharing raw data among users defeats the original privacy-preserving purpose of FL, thereby rendering such pre-sharing method unsuitable for distributed mobile environments.

As a key difference from distributed learning, non-IIDness is discussed in~\cite{non-iid-converge,arm-noniid,fed-distillation}. It is shown in~\cite{non-iid-converge} that for strongly convex and smooth problems, FedAvg still retains the same convergence rate on non-IID data. However, convergence may be brittle for the rest non-convex majorities like multi-layer neural networks. As a remedy,~\cite{arm-noniid} pre-shares a subset of non-sensitive global data to mobile devices and~\cite{fed-distillation} utilizes a generative model to restore the data back to IID, but at non-negligible computation, communication and coordination efforts. Another thread of works address the common problem of communication efficiency in FL~\cite{konecny,cmfl-icdcs,decentralize_liu,evolution,multi-task-nips,tiered}. The full model is compressed and cast into a low-dimensional space for bandwidth saving in~\cite{konecny}. Local updates that diverge from the global model are identified and excluded to avoid unnecessary communication~\cite{cmfl-icdcs}. Decentralized approaches and convergence are discussed in~\cite{decentralize_liu} when users only exchange gradients with their neighbors. Evolutionary algorithm is explored to minimize communication cost and test error in a multi-objective optimization~\cite{evolution}. The challenges from system, statistics and fault tolerance are jointly formulated into a unified multi-task learning framework~\cite{multi-task-nips}. A new architecture is proposed with tiered gradient aggregation for saving network bandwidth~\cite{tiered}. Such aggregation re-weights the individual's contribution to the global model, that may unwittingly emphasize the share of non-IID users. As recommended by~\cite{google_ai_blog}, a practical way to save the monetary cost of communication is to schedule FL tasks at night when the devices are usually charging and connected to WiFi. These efforts are orthogonal to our research and can be efficiently integrated to complement our design.

Our study has fundamental difference from a large body of works in scheduling paralleled machines~\cite{schedule,optimus-eurosys,scheduling-online}. First, rather than targeting at jobs submitted by cloud users, we delve into a more microcosmic level and jointly consider partitioning a learning task and makespan minimization, where the cost function is characterized from real experimental traces. Second, FL calls for the scheduling algorithm to be aware of non-IIDness and model accuracy when workloads are partitioned. Hence, our work is among the first to address computational and statistical heterogeneity on mobile devices.

%

%

%

\section{Computation vs. Communication Time} \label{sec:computation}

\begin{figure}[ht!]
\centering
\hspace*{-0.3in}
\begin{subfigure}[b]{0.23\textwidth}
                \includegraphics[width=1.1\textwidth]{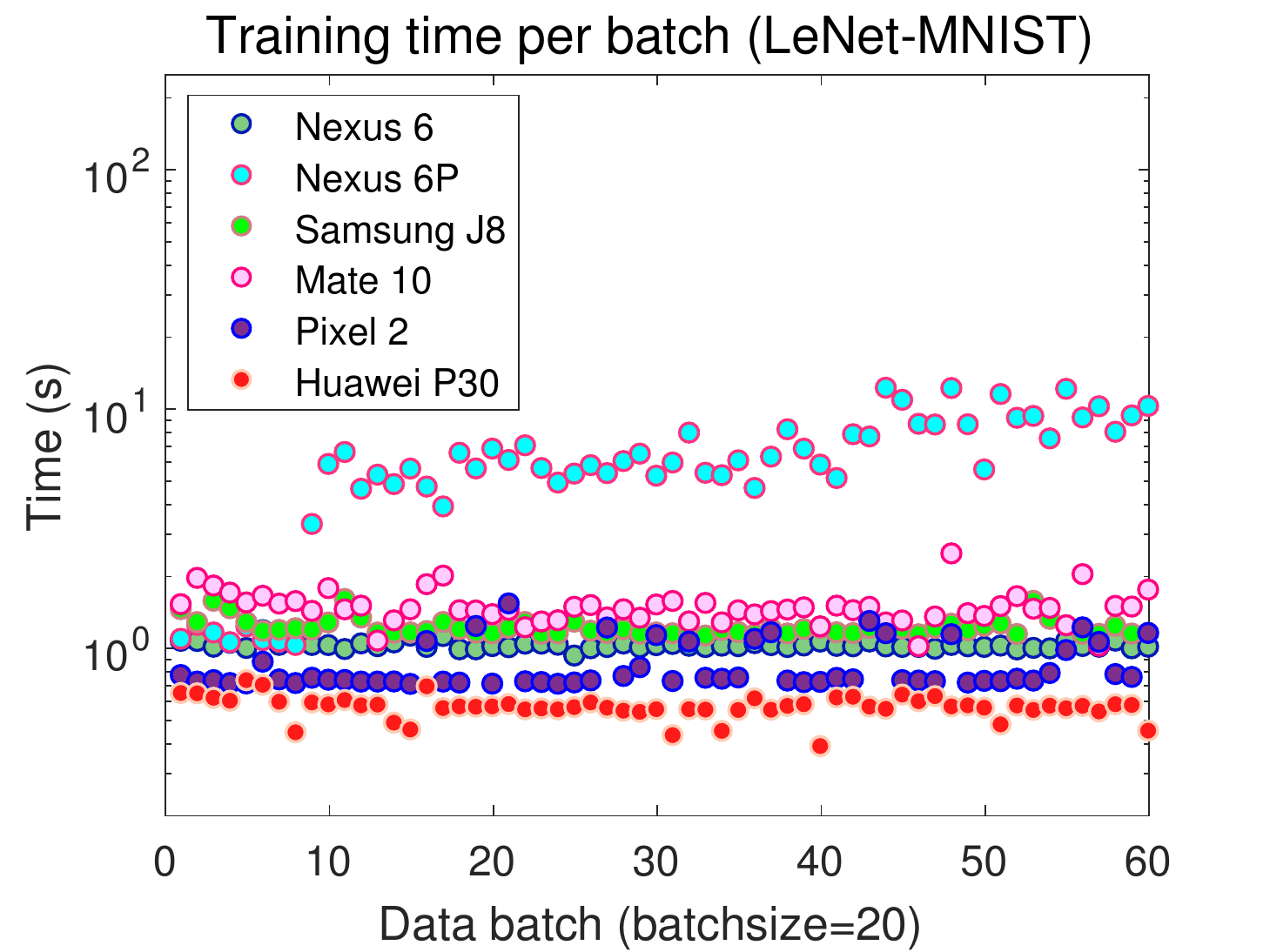}
\end{subfigure}
\hspace*{0.1in}
\begin{subfigure}[b]{0.23\textwidth}
        \includegraphics[width=1.1\textwidth]{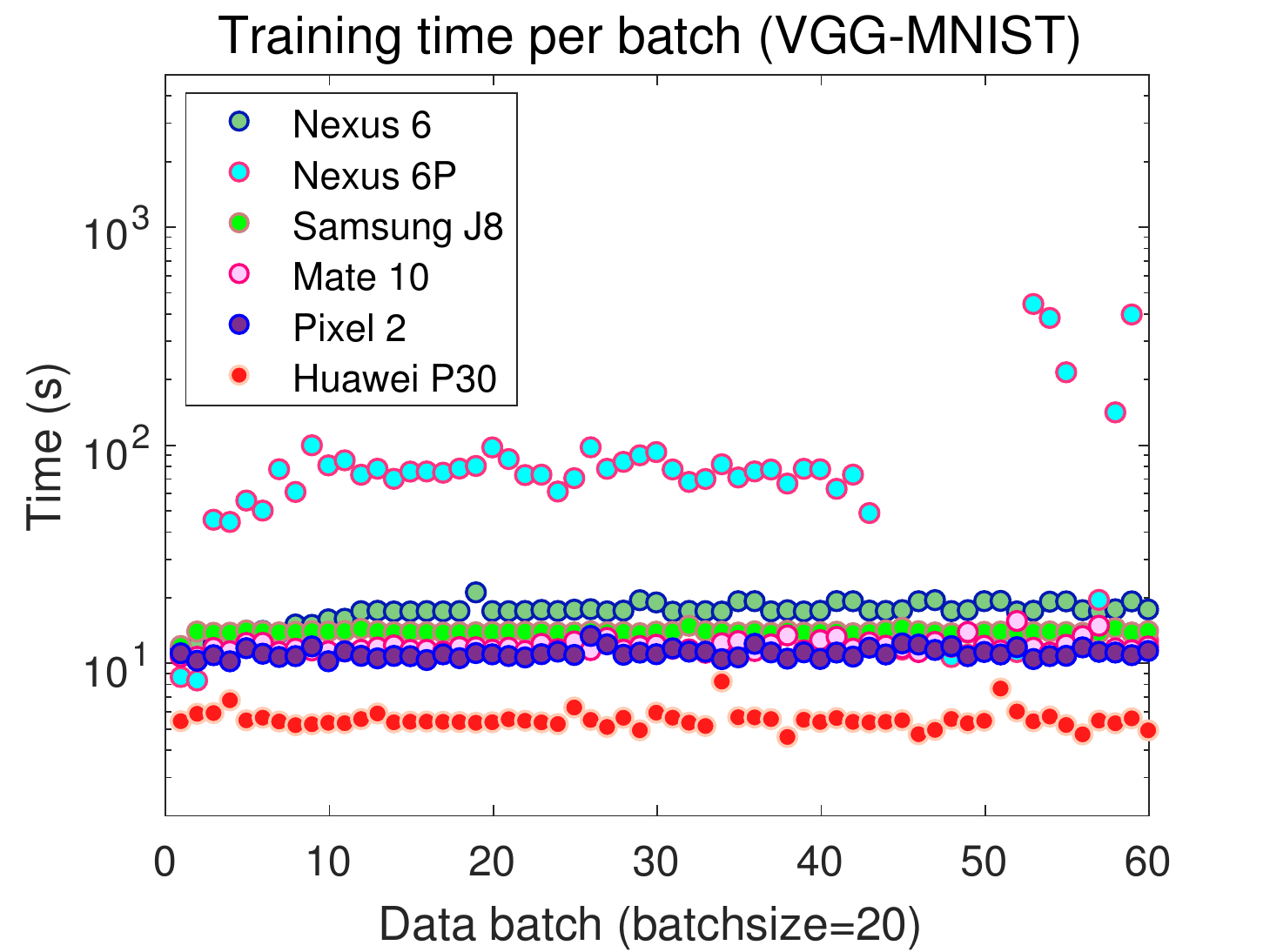}
\end{subfigure}
\vspace{-0.05in}
{\small \hspace{0.1in} (a) \hspace{1.55in} (b)}
\vspace{-0.1in}
\caption{Benchmark training time on different mobile devices (MNIST dataset) (a) LeNet (b) VGG6 (best view in color)}
\label{time_diff_fig}
\end{figure}

In this section, we are motivated to answer the basic question: \emph{how large is the gap of the computational time among mobile devices and how does it compare to communication}? We demonstrate this through an empirical study to launch a training application using neural network models of LeNet~\cite{lenet} and VGG6~\cite{vgg} on the mobile testbed (shown in Table \ref{table:testbed} in Sec. \ref{sec:eval}).

To benchmark the computation time, we trace the training time per data batch (20 samples) on different devices shown in Fig. \ref{time_diff_fig} and the average of CPU clock speed every $5$s vs. the temperature in Fig. \ref{cpu_temperature_fig}. Though the CPUs can switch frequencies much faster, this experiment shows how the frequency and temperature interact over time to reach stability under the power management policy (in \texttt{/system/etc/thermal-engine.conf}).

\begin{figure}[ht!]
\centering
\hspace*{-0.3in}
\begin{subfigure}[b]{0.23\textwidth}
                \includegraphics[width=1.1\textwidth]{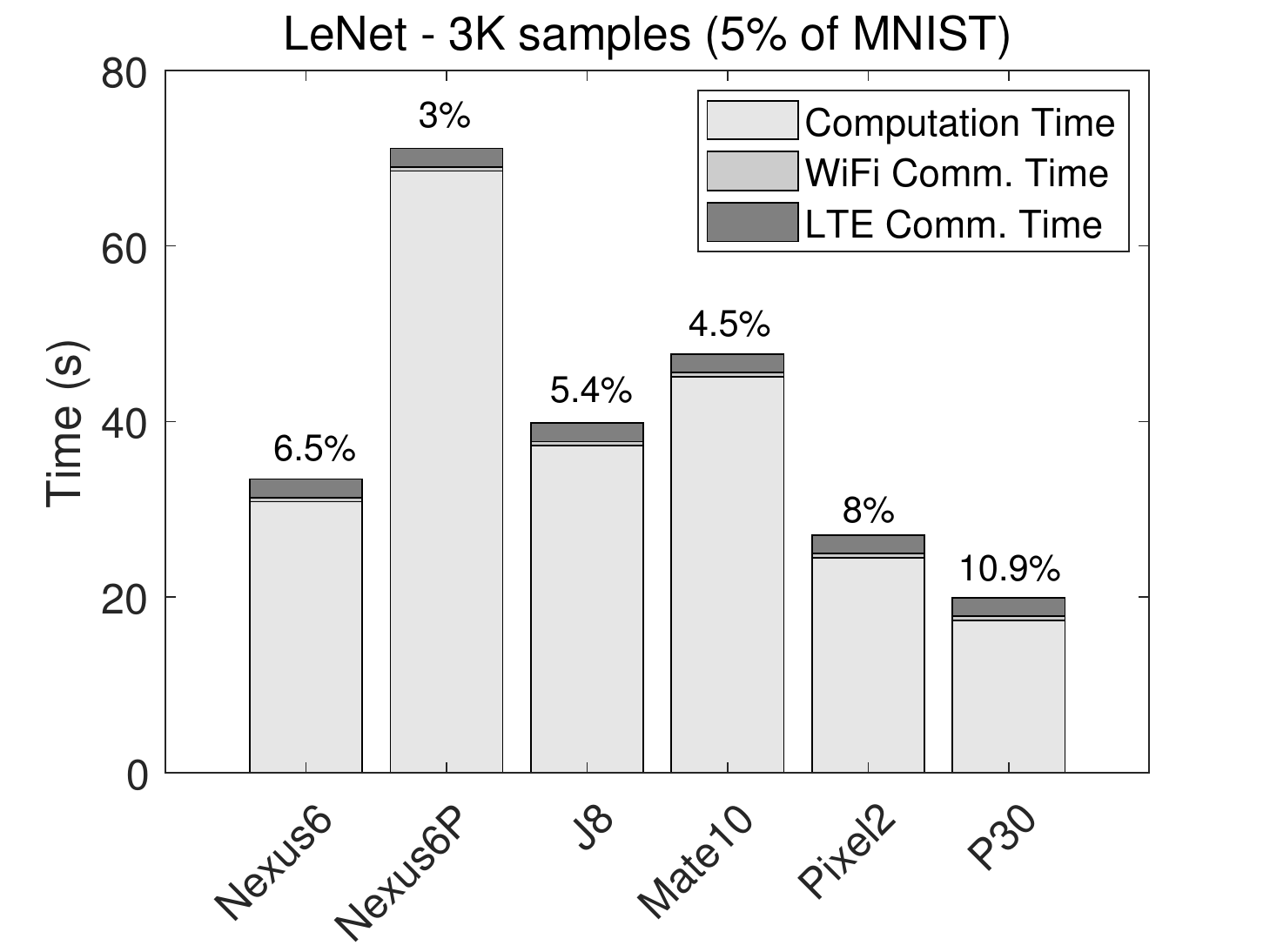}
\end{subfigure}
\hspace*{0.1in}
\begin{subfigure}[b]{0.23\textwidth}
        \includegraphics[width=1.1\textwidth]{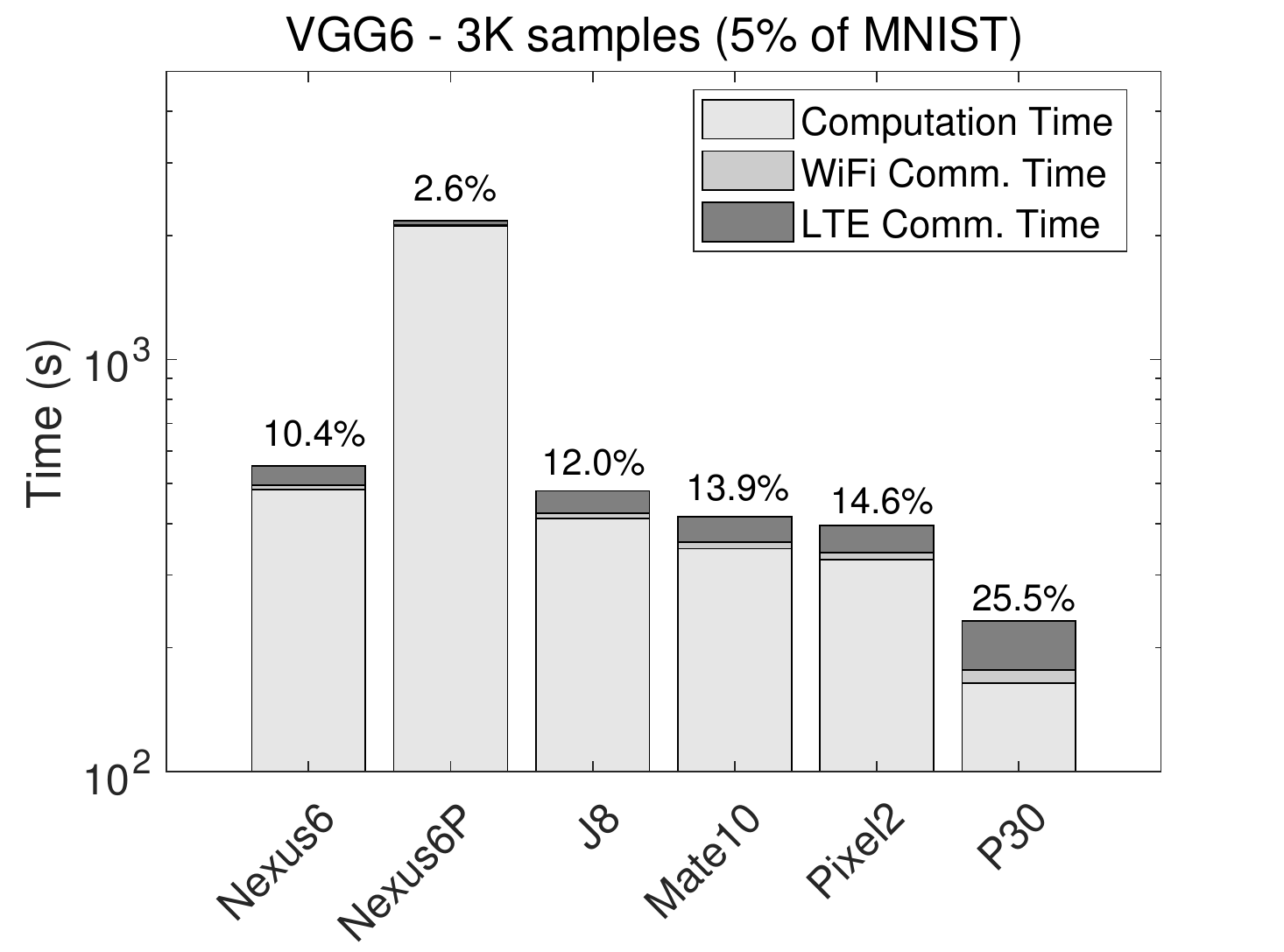}
\end{subfigure}
\vspace{-0.05in}
{\small \hspace{0.1in} (a) \hspace{1.55in} (b)}
\vspace{-0.1in}
\caption{Computation vs. communication time of training on MNIST samples per epoch (s) (a) LeNet (b) VGG6}
\label{comm_vs_compute_fig}
\end{figure}

To measure the communication time, we establish an AWS server for communicating the model between the cloud (Washington D.C.) and local devices (Norfolk, VA). The server pushes (pulls) the model to (from) the devices in each epoch and iterates through 5\% of the dataset with 3K samples. We measure the transmission time of the LeNet (2.5MB) and VGG6 (65.4MB) model over the 1 Gbps wireless link and T-mobile 4G LTE (-94 dBm), to emulate different networking environments. The WiFi uplink/downlink speed achieves around 80-90 Mbps on our campus network and LTE reaches about 60 Mbps and 11 Mbps for the uplink and downlink respectively. The makespan for each device is shown in Fig. \ref{comm_vs_compute_fig} with the percentage of communication overhead on top. Based on all the experiments above, we summarize the observations below.

\subsection{Key Observations}

\noindent\textbf{Observation 1.} Computation time is mainly governed by the processing power/performance of the CPUs/SoC (with some variations from the OEM implementations), as well as the computation intensity of the neural network model.

\noindent\textbf{Observation 2.} The continuous neural computation leads to \emph{thermal throttling}, where the governor quickly reacts to reduce the \texttt{cpufreq}, or even shuts down some cores, thereby causing a performance hit with large variance in the subsequent batch iterations (especially running heavy-weight networks like VGG6 on Nexus6/6P). Such phenomenon adds to the diversity of computation time.

\noindent\textbf{Observation 3.} In contrast to the hypothesis in~\cite{fedavg} that communication overhead dominates in FL, our experiments indicate that communication only takes a small portion of the training time (below 10\% on average as seen in Fig. \ref{comm_vs_compute_fig}). This confirms that with today's networking speed and the upcoming 5G, the bottleneck of FL on battery-powered mobile devices is expected to remain on the computational side. Part of the reason is because the consumer mobile devices cannot host heavy-weight neural architectures (such as increasing VGG6 to 16 layers), which easily overwhelm the memory limit of 512 MB per application set by Android~\cite{largeheap}. Even if more memory is permitted to execute large models, the bottleneck would still remain on the computational side because of the parabolical increase of computation time, compared to the relatively linear increase in communication time.

\noindent\textbf{Observation 4.} To process the same amount of data, the mobile devices exhibit substantial heterogeneity in their completion time. For example, the straggler takes more than $3\times$ compared to the mean completion time from other devices, and this deviation is expected to get larger with higher workloads such as more complex models or data iterations.

\subsection{CPU Frequency vs. Temperature}

We observe two major factors: hardware architecture and thermal throttling, that impact the computation speed for such sustained workloads of running backpropagation. Of course, speed is dominated by the hardware architecture such as the CPUs/SoC, and should be in general consistent and predictable with the hardware configurations. However, unlike desktops, thermally constrained mobile devices result higher diversity in runtime as per to the different workloads and computation intensity.

\begin{figure}[ht!]
\centering
\begin{subfigure}[b]{0.23\textwidth}
                \includegraphics[width=1.1\textwidth]{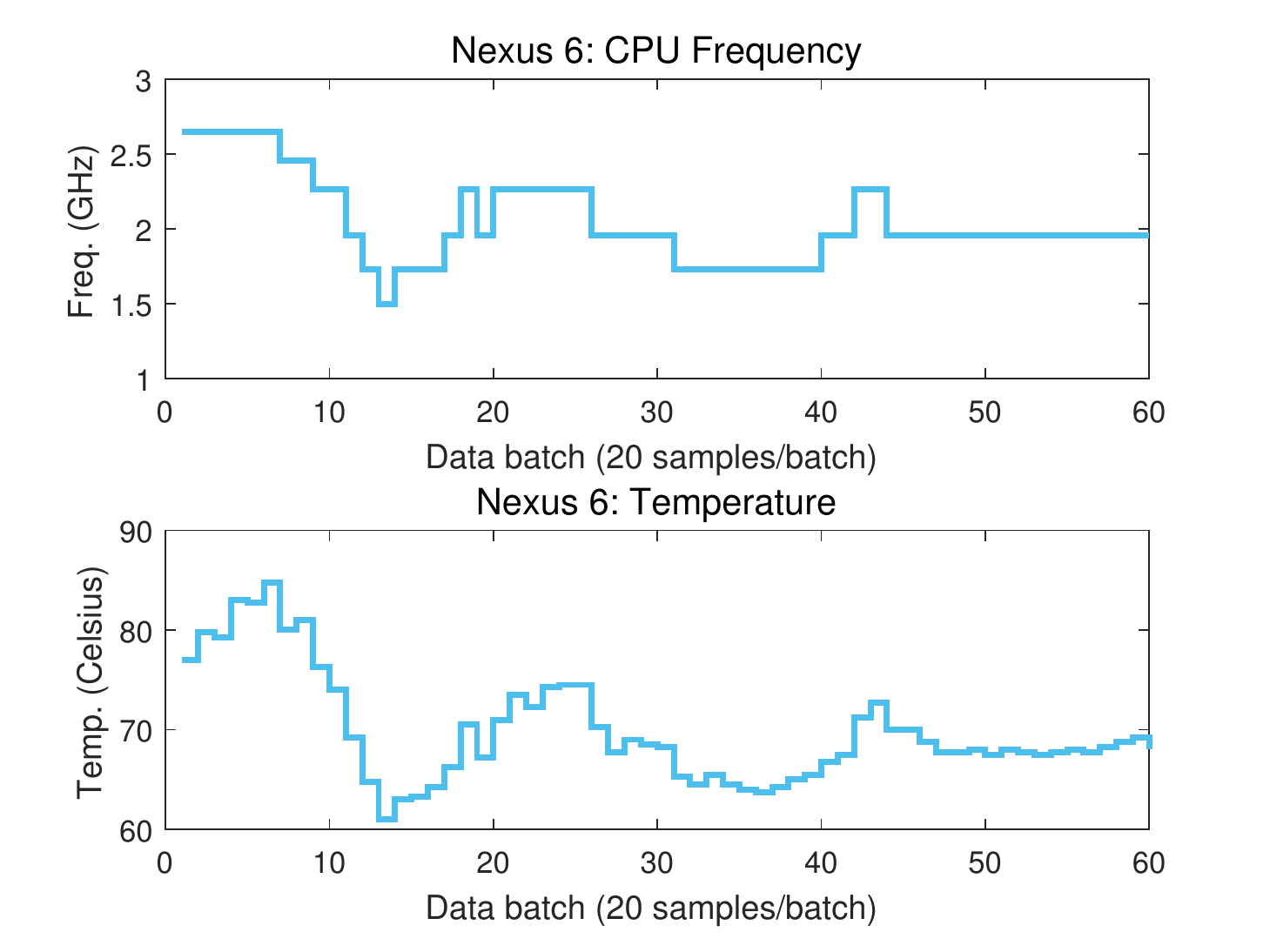}
\end{subfigure}
\begin{subfigure}[b]{0.23\textwidth}
        \includegraphics[width=1.1\textwidth]{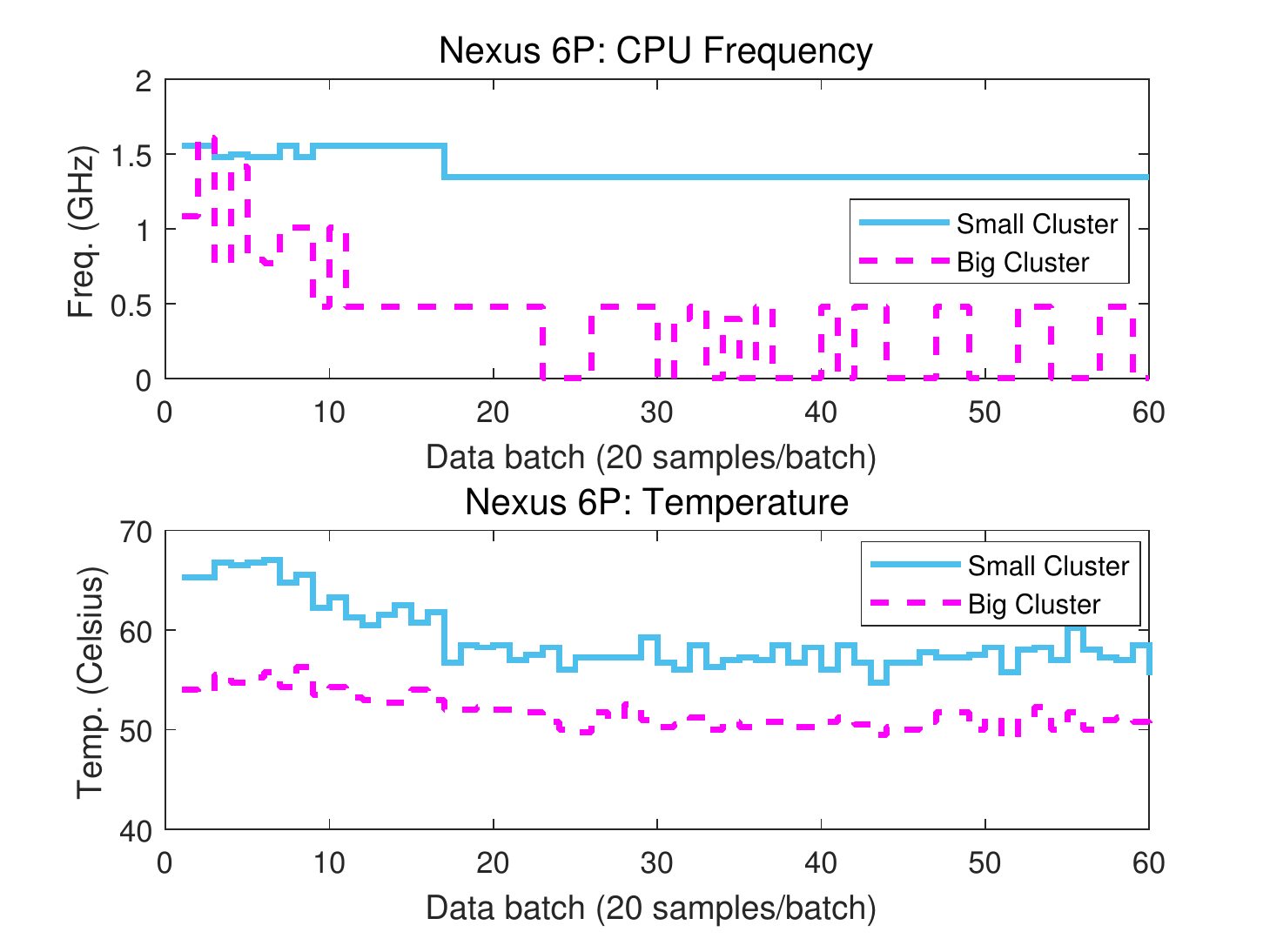}
\end{subfigure}
{\small \hspace{0.1in} (a) \hspace{1.55in} (b)}
\begin{subfigure}[b]{0.23\textwidth}
                \includegraphics[width=1.1\textwidth]{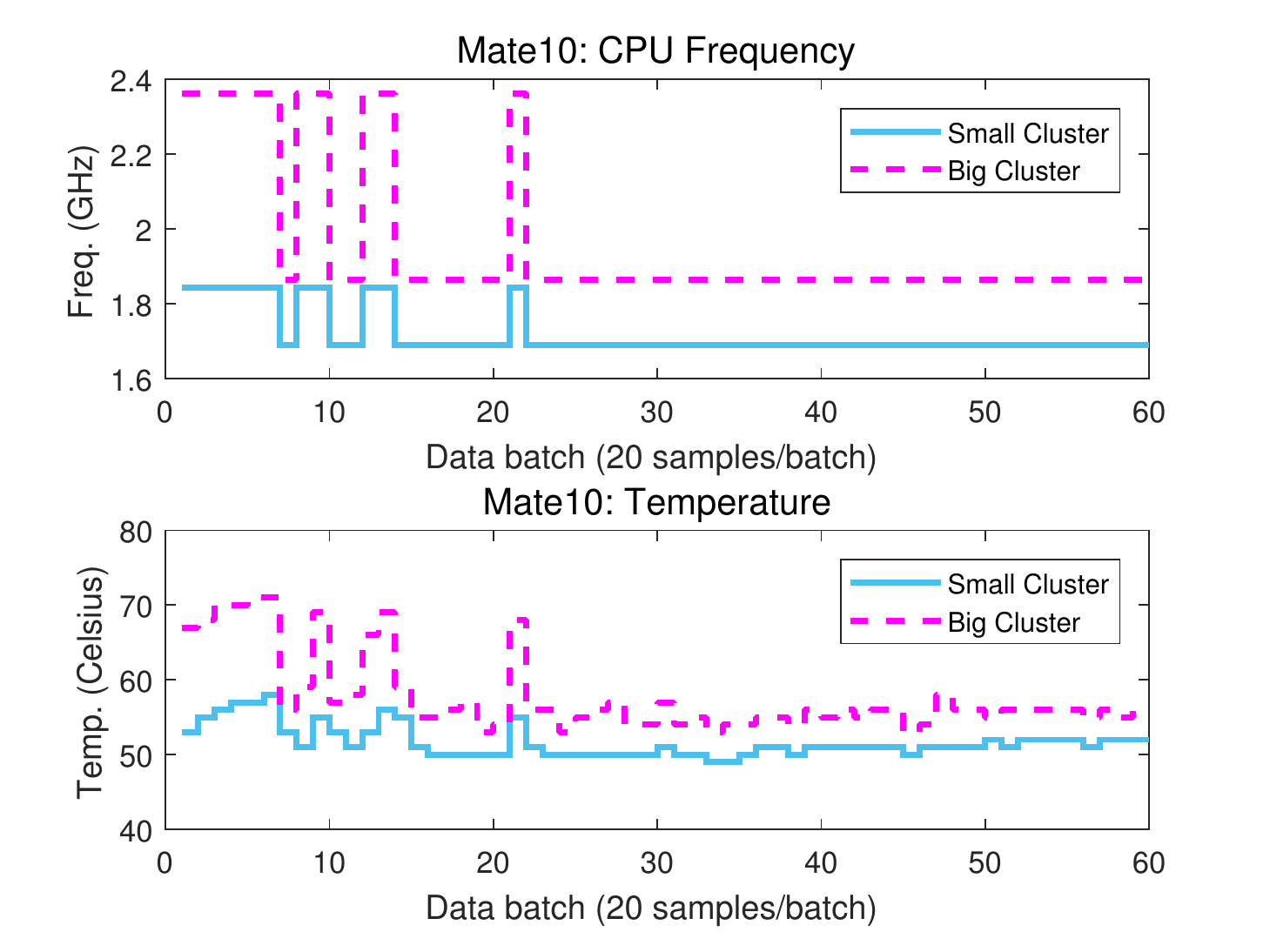}
\end{subfigure}
\begin{subfigure}[b]{0.23\textwidth}
        \includegraphics[width=1.1\textwidth]{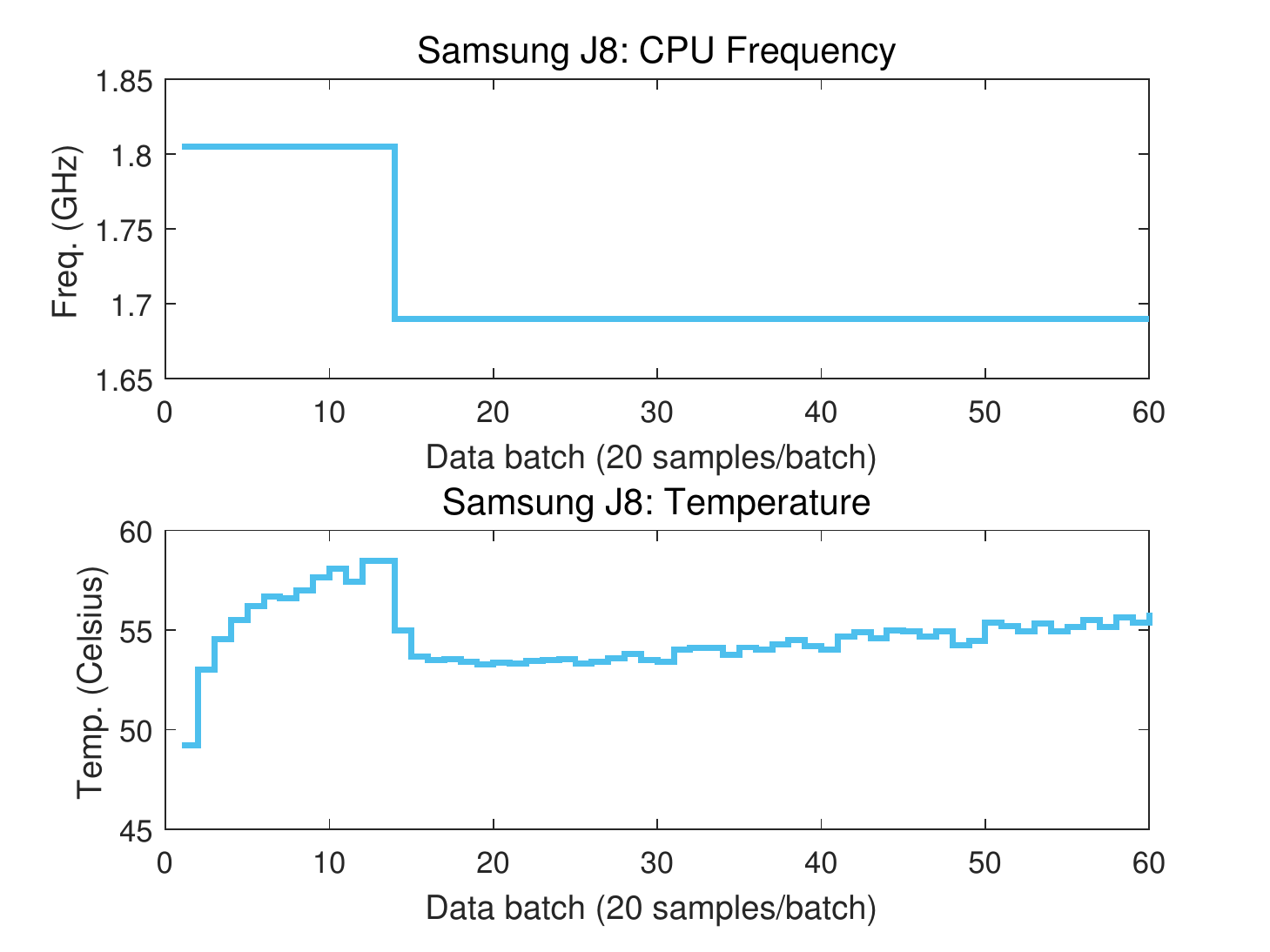}
\end{subfigure}
{\small \hspace{0.1in} (c) \hspace{1.55in} (d)}
\vspace{-0.05in}
\caption{CPU clock speed vs. temperature. (a) Nexus6 (b) Nexus6P (c) Mate10 (d) SamsungJ8.}
\label{cpu_temperature_fig}
\end{figure}

We show the trace of CPU frequency and temperature of four devices in Fig. \ref{cpu_temperature_fig}. Before analyzing the results, we briefly describe their CPU microarchitectures: 1) Nexus 6 has a single quad-core CPU cluster running at 2.7GHz; 2) Nexus 6P has octa-core CPU clusters. The four big cores are running at 2.0 GHz and the four little cores are running at 1.55 GHz; 3) Mate10 also has octa-core CPU clusters. The four big cores are running at 2.4 GHz and the four little cores are running at 1.8 GHz; 4) SamsungJ8 features octa-core all running at 1.8GHz in a single cluster. All of them are the maximum frequencies of the CPU. For clarity, we average the frequency and temperature for all the homogeneous cores in Fig. \ref{cpu_temperature_fig}.

We can see that different devices exhibit distinct behaviors of how the governor reacts to temperature surge. Nexus 6 allows the temperature to stay above $60^{\circ}$C and even surpass the $70^{\circ}$C level, in exchange for running the CPU only 20-30\% below the max. frequency. In contrast, Nexus6P is more conservative due to the controversial Snapdragon 810 SoC~\cite{heat}. It actively reduces the frequency of the big cores to below 50\%, and even switches off all the cores in the big cluster, in order to maintain the temperature around $50^{\circ}$C. The big cores go offline and migrate the tasks to the little cores after a moderate temperature surge, that occurs fairly often during the testing. The big cores never stay around their maximum frequency at 2.0 GHz, thus making Nexus 6P much slower than Nexus 6 even with more CPU cores. The recent generations of Mate10 and Samsung J8 exhibits more stability. The governors manage to maintain the temperature around $50^{\circ}$C with only 20-30\% discounted clockspeed.

The throttling process is controlled by the vendor-specific driver for frequency scaling. Obviously, Nexus 6 has a more relaxed throttling temperature, which allows the CPUs to stay at high frequency and persistently online than Nexus 6P. Such possible overclocking has made it outclass the newer versions of Mate10 on some low intensity tasks (about $3\times$ speedup calculated in from Fig. 1(a)), though Nexus 6 back in 2014 were not designed for intensive workload like neural computations. These experimental studies suggest that we should factor in both the device-specific characteristics and the workload intensity in estimating the computation time.

Since FL takes more than 10 epoches to converge, as shown in Fig. \ref{comm_vs_compute_fig}, only processing 5\% MNIST for 10 epoches leads to 0.7 to 2.6 hours time difference to wait for the stragglers. If each model update entails more local iterations, this delay is expected to increase exponentially (with more thermal throttling on the stragglers). An optimized scheduling mechanism should be built to account for these factors. In cloud environments, stragglers may be caused by resource contention, unbalanced workload or displacement of workers on different parameter servers~\cite{optimus-eurosys}, which are typically handled by \emph{load balancing}. In mobile environment, they are caused by the fundamental disparity among users' devices: can we do the opposite and leverage \emph{load unbalancing} to offset the computation time of those stragglers? If so, what about the side effects and how to mitigate? Since each epoch requires a full pass of the local data, among a variety of tunable knobs, workload is directly proportional to the amount of training data. Nevertheless, distributed learning often assumes a balanced data partition among the workers~\cite{fedavg}. Would data imbalance (either in the case of IID or non-IID data distribution) lead to significant accuracy loss? We further study additional impact from the data distributions before formulating our problem.

\section{Impact of Data Distributions} \label{sec:distribution}

\subsection{Impact of Data Imbalance to IID Data}   \label{sec:impact_imbalance}
\begin{figure}[!t]
\vspace*{-0.09in}
\centering
\hspace*{-0.3in}
\begin{subfigure}[b]{0.25\textwidth}
                \includegraphics[width=1.01\textwidth]{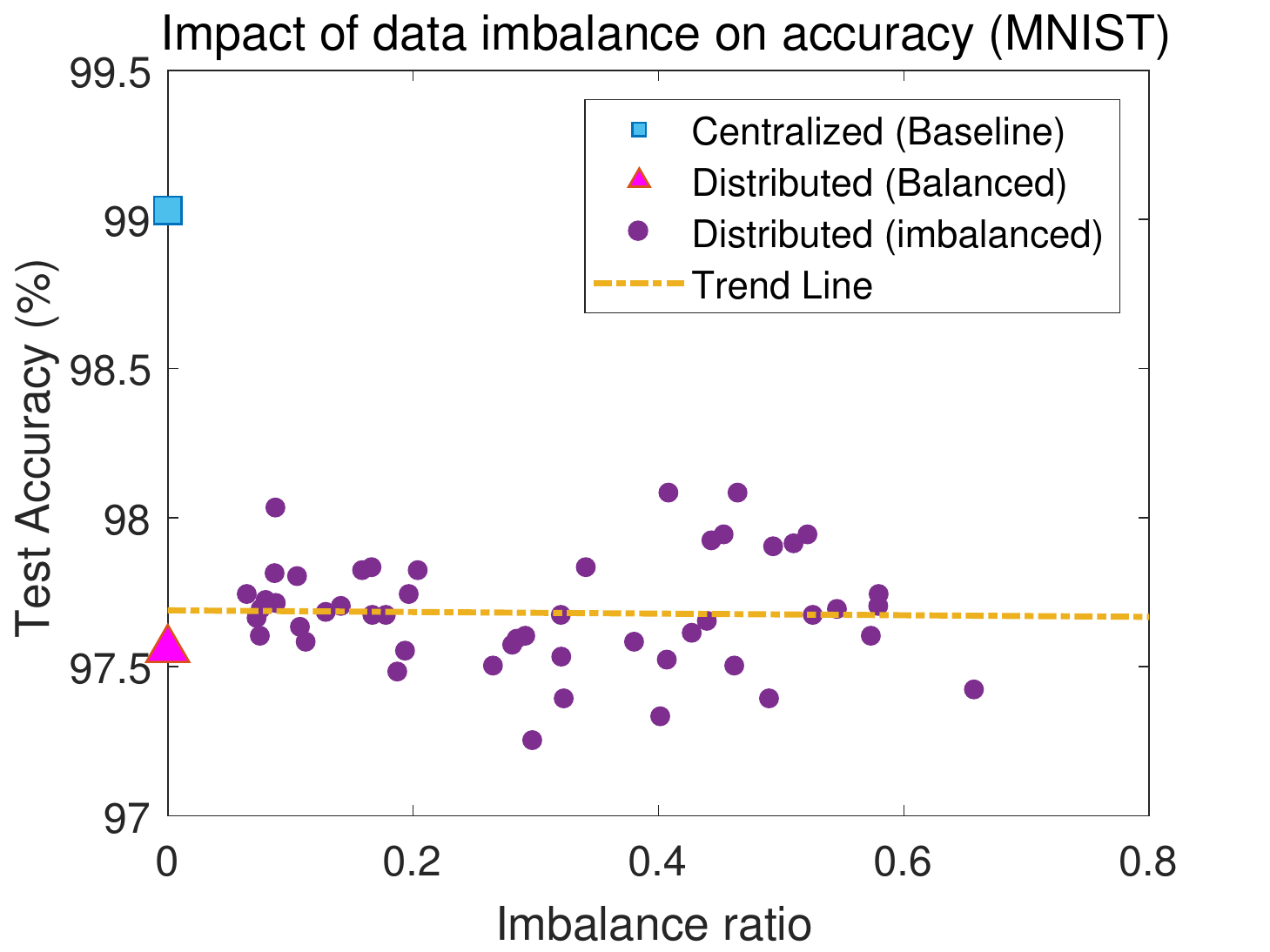}
                \vspace{-0.2in}
                \caption{}
\end{subfigure}
\begin{subfigure}[b]{0.25\textwidth}
                \includegraphics[width=1.01\textwidth]{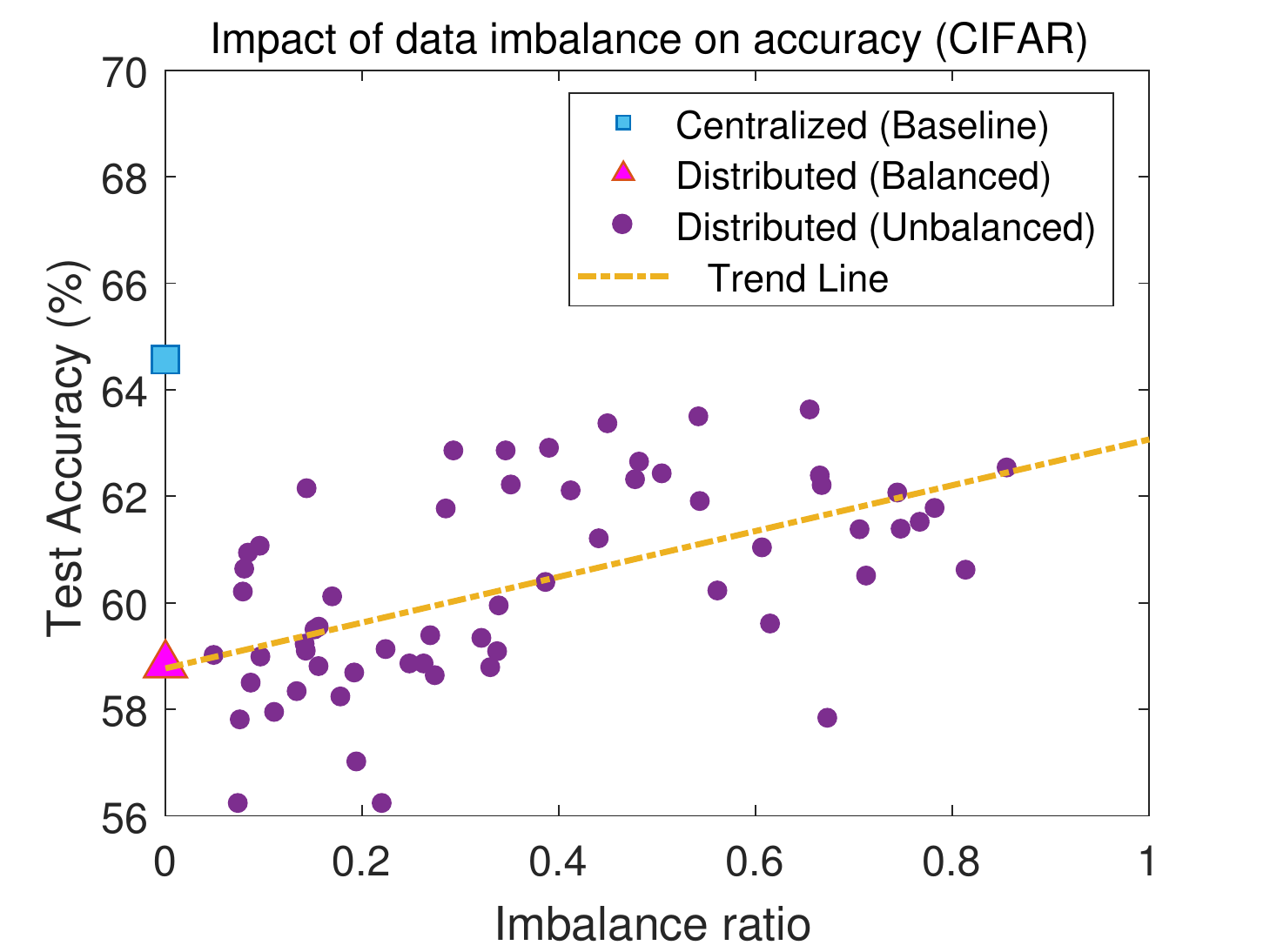}
                \vspace{-0.2in}
                \caption{}
\end{subfigure}
\hspace*{0.05in}
\hspace*{-0.3in}
\vspace*{-0.1in}
\caption{Impact of data imbalance (still IID) on FL accuracy (a) MNIST (b) CIFAR10}
\label{imbalance_fig}
\vspace*{-0.2in}
\end{figure}

We partition the datasets of MNIST and CIFAR10 among $20$ users. E.g., for MNIST, the training set of 60K images results an average of 3K images per user. Then we utilize a Gaussian distribution to sample around the mean and adjust the standard deviation to induce data imbalance among users. The ratio between different classes is uniform so no class dominates the local set. We utilize an index of \emph{imbalance ratio} between the standard deviation and the mean as the x-axis (larger ratio means more extreme), and benchmark the accuracy against the centralized and distributed learning with balanced data in Fig.\ref{imbalance_fig}. The results indicate that as long as the data remains IID, imbalance does not lead to accuracy loss. It is reasonable since the local gradients still resemble each other when data is IID. The accuracy even trends up a little for CIFAR10. This provides the basis for optimization discussed in the next section.

\subsection{Impact of non-IID Data}   \label{sec:impact_non_iid}
\begin{figure}[!t]
\vspace*{-0.09in}
\centering
\hspace*{-0.1in}
\begin{subfigure}[b]{0.25\textwidth}
                \includegraphics[width=1.05\textwidth]{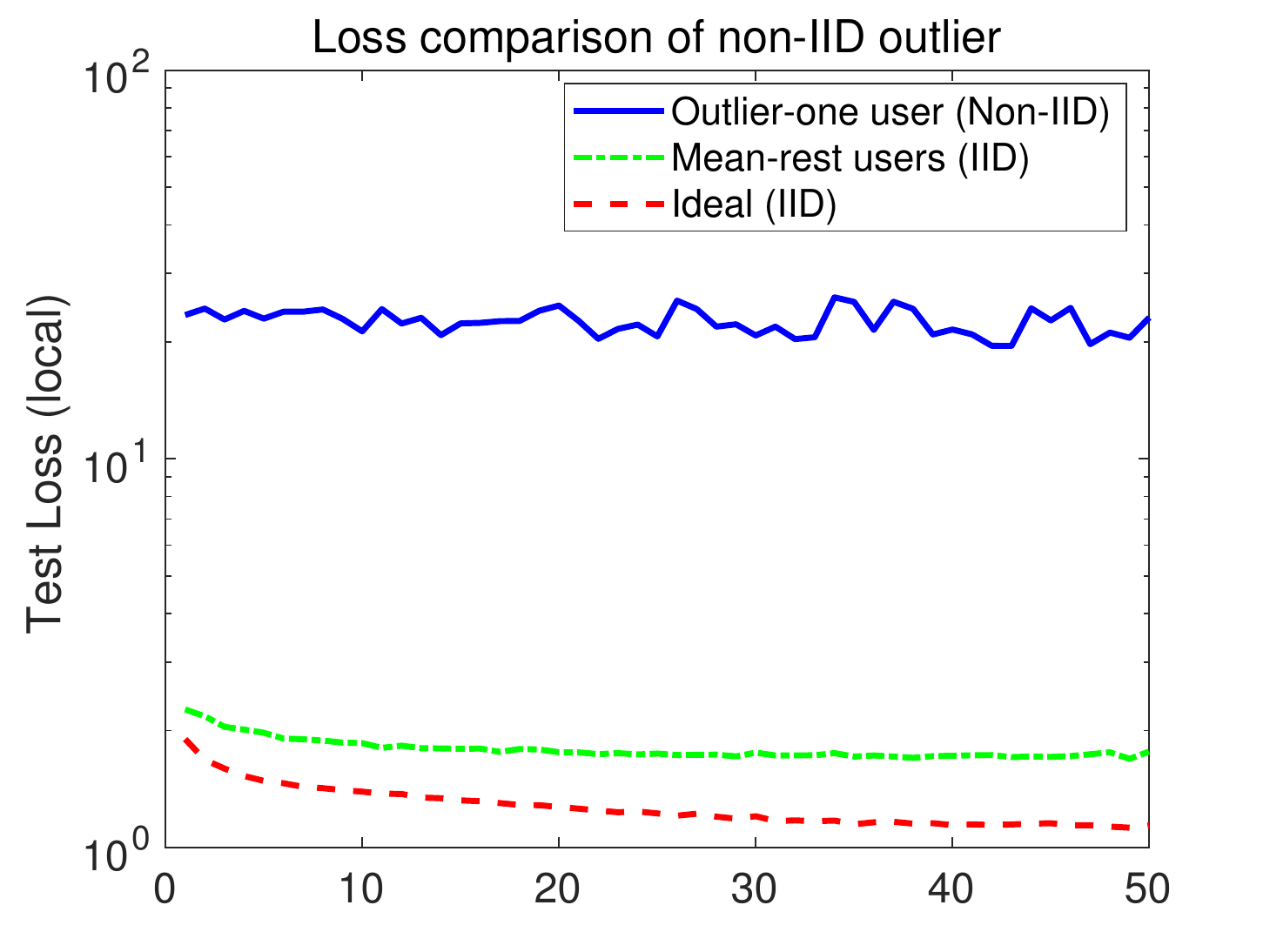}
                \vspace{-0.2in}
                \caption{}
\end{subfigure}
\hspace*{-0.1in}
\begin{subfigure}[b]{0.25\textwidth}
                \includegraphics[width=1.05\textwidth]{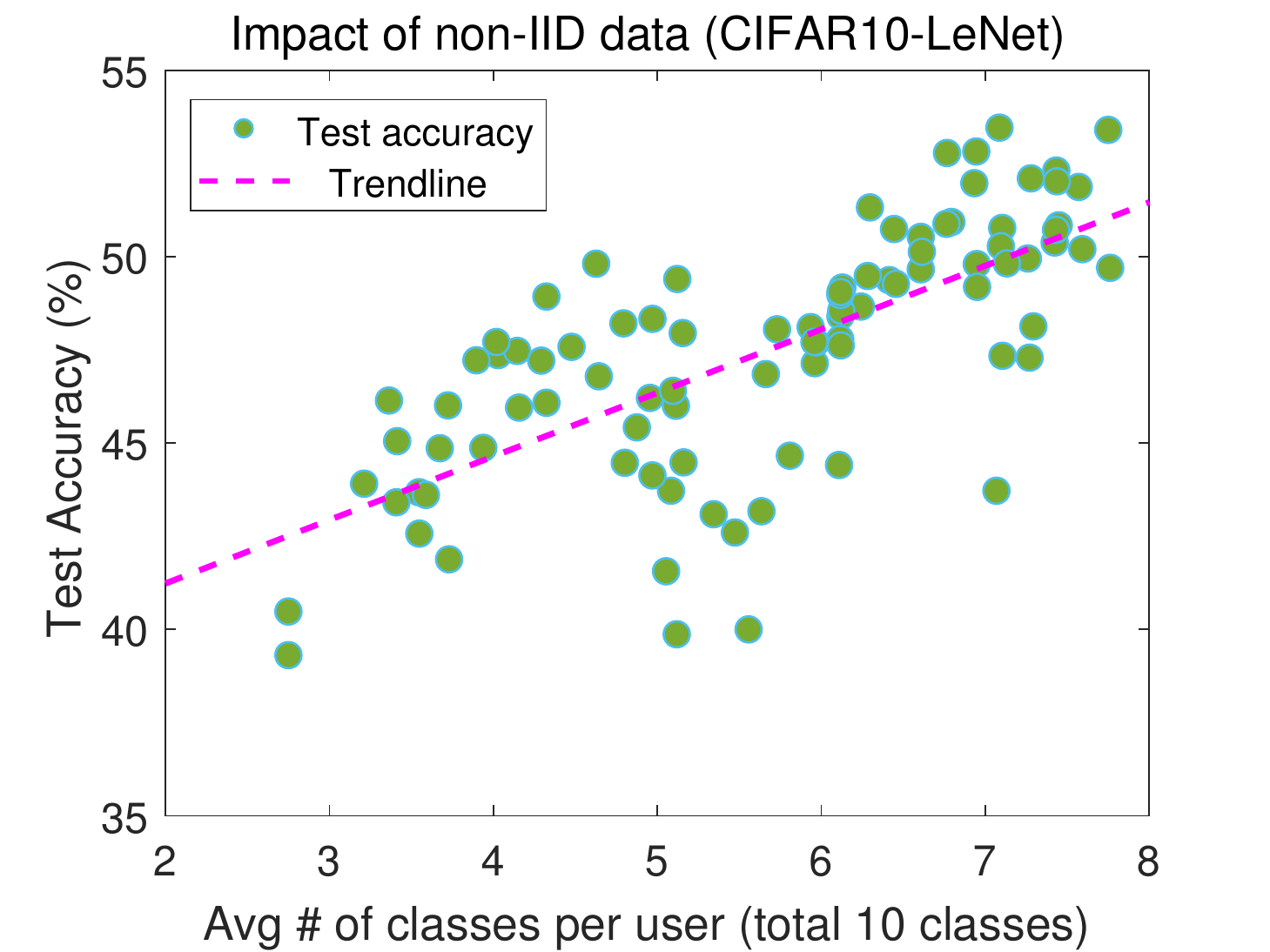}
                \vspace{-0.2in}
                \caption{}
\end{subfigure}
\hspace*{0.05in}
\vspace*{-0.1in}
\caption{Impact of non-IID data on local convergence and model accuracy (CIFAR10) (a) comparison of local loss (b) relation between the degree of non-IID class distribution and accuracy. }
\label{non_iid_fig_1}
\vspace*{-0.11in}
\end{figure}

Non-IIDness is shown to have negative impact on collaborative convergence~\cite{fedavg,arm-noniid} and data imbalance could exacerbate this issue. Instead of investigating data imbalance and non-IIDness together, we investigate how non-IID data alone is enough to impact accuracy and convergence. We seek answers to the fundamental question: \emph{How can we identify and deal with the users having non-IID distributions?} Weight divergence, $\|\bm{w}_i - \sum_{i=1}^{N} \bm{w}_i/N \|_2^2$ is used in~\cite{arm-noniid} to compare the norm difference between the local weights $w_i$ and the global average. Local loss is an equivalent indicator with less complexity, since it does not require pairwise weight computations. Fig. \ref{non_iid_fig_1}(a) compares the local loss of the non-IID outlier with the average loss from the rest users and the ideal case when the data is IID. The outlier user with only one class can be easily identified from the rest IID ones, which has over an order of magnitude loss value and is unable to converge compared to the rest.

While identifying outliers is simple, the existing studies~\cite{cmfl-icdcs,arm-noniid,fed-distillation} have yet to reach a consensus on how to deal with them. With the outlier user only having a subset of classes, our intuition is that accuracy is directly associated with the distribution of classes among the users~\cite{cmfl-icdcs}. To see how the number of classes impacts accuracy, we conduct the second experiment by iterating the number of classes per user from $2$-$8$ (out of $10$ classes) plus a standard deviation of samples among the existing classes as the $x$-axis. It is observed in Fig. \ref{non_iid_fig_1}(b) that higher disparity of class distributions among the users indeed leads to more accuracy degradation with a substantial loss of 10-15\% on CIFAR10.

\begin{figure}[!t]
\vspace*{-0.09in}
\centering
\hspace*{-0.1in}
\begin{subfigure}[b]{0.25\textwidth}
                \includegraphics[width=1.05\textwidth]{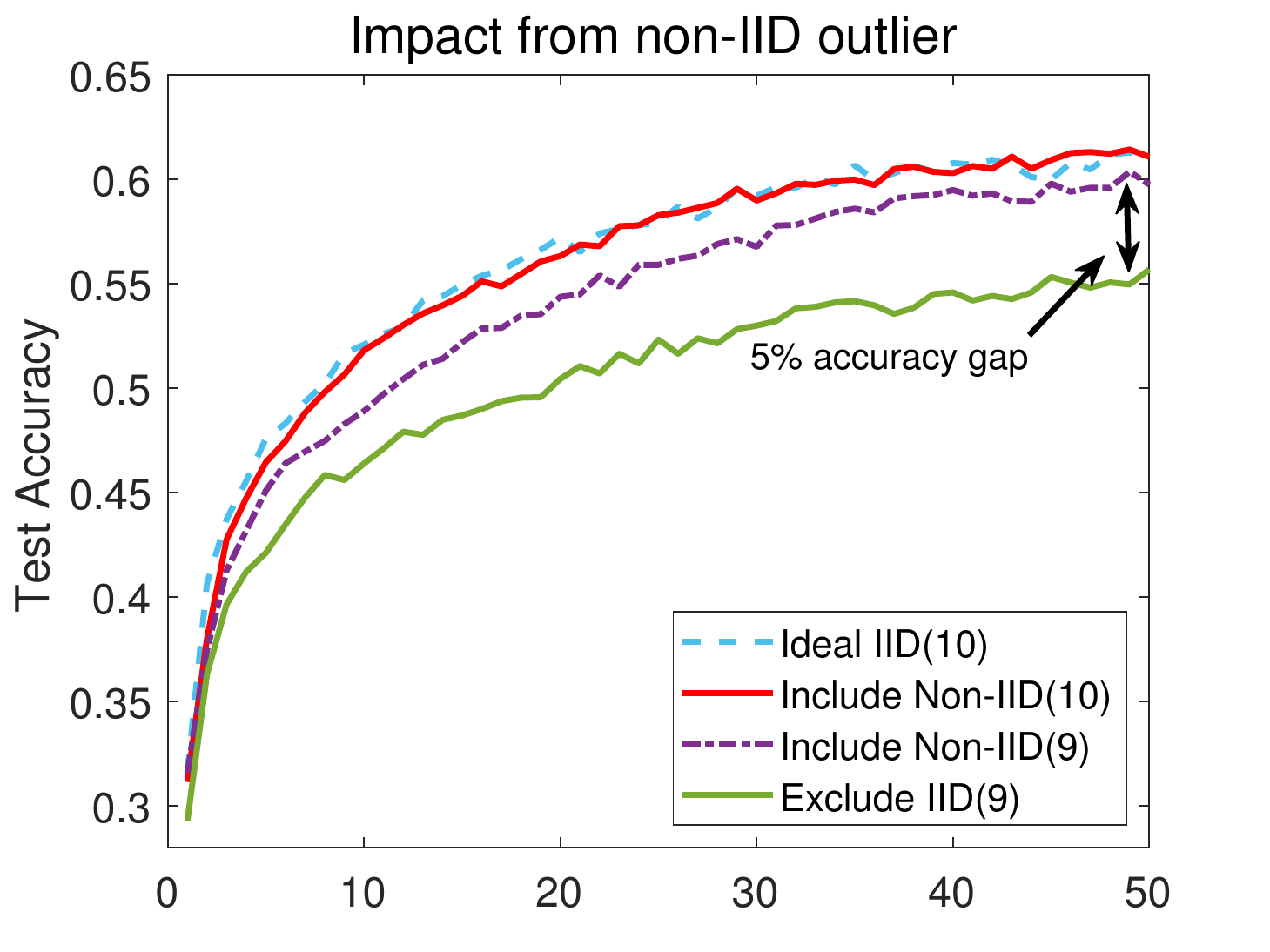}
                \vspace{-0.2in}
                \caption{}
\end{subfigure}
\hspace*{-0.1in}
\begin{subfigure}[b]{0.25\textwidth}
                \includegraphics[width=1.05\textwidth]{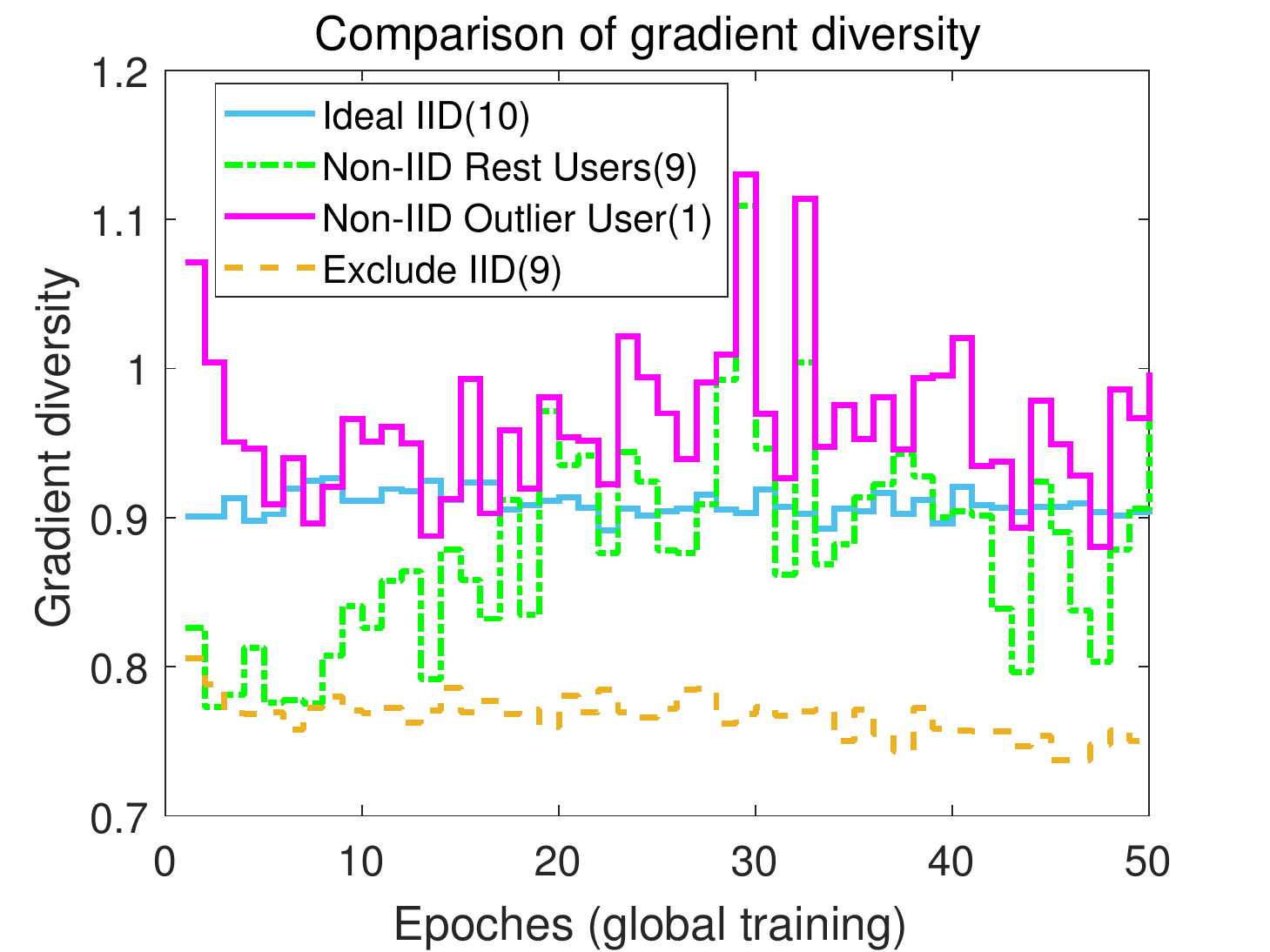}
                \vspace{-0.2in}
                \caption{}
\end{subfigure}
\hspace*{0.05in}
\vspace*{-0.1in}
\caption{Impact of non-IID data on model accuracy in CIFAR10 (a) relation between the degree of non-IID class distribution and accuracy (b) influence from individual outliers. }
\label{non_iid_fig_2}
\vspace*{-0.11in}
\end{figure}

Since the presence of non-IID outliers is inevitable in practices, we are facing two options: 1) simply exclude them from the population based on loss divergence~\cite{cmfl-icdcs}; 2) keep them in training. We take a closer look of these options and argue that the decision should be actually conditioning on the class distributions, rather than only based on the local loss or weight divergence. We demonstrate through a simple case to distribute CIFAR10 dataset among 4 users in different ways, and introduce a fifth user to act as the non-IID outlier.
\begin{itemize}
  \item \emph{Ideal IID(10)}: the ideal baseline when all $4$ users have identical distribution and all 10 classes are evenly distributed among them.
  \item \emph{Include Non-IID(10)}: the population has 10 classes. A fifth user with only one class is included, but her class has already presented in the population. The population becomes non-IID because of the fifth user.
  \item \emph{Include Non-IID(9)}: the population has 9 classes. A fifth user with that missing class (one-class outlier) is included. The population is also non-IID.
  \item \emph{Exclude IID(9)}: the population has 9 classes. Exclude the fifth user despite she possesses class samples from the missing class so the population remains IID~\cite{cmfl-icdcs}.
\end{itemize}
From the results in Fig. \ref{non_iid_fig_2}(a), we have two key observations: 1) if the class of the outlier user is also found in the population, inclusion of the outlier has minor influence on accuracy (but slows down convergence); 2) if such class is missing from the population, including the outlier user results 1-2\% accuracy loss compared to the ideal IID case, but achieving a significant improvement of 5\% accuracy compared to when the outlier user is completely removed from training. Based on these observations, we hypothesize that the selection of participants should not only rely on the local losses, but also whether they contain classes that are not yet included in the training set.

\textbf{Summary.} The findings are inline with two fundamental principles of machine learning and the generality is effective for other experimental setup and datasets as well. FL adopts the principles of \emph{data parallelism}~\cite{dean}. As long as the training data is partitioned in an IID fashion, it does not negatively impact the test accuracy. In fact, data parallelism will improve the accuracy during a fixed training time because of fast convergence. It is applicable to any neural network architecture and is model-agnostic. For non-IID data, our findings are based on: 1) class imbalance/non-IID data has negative impact on accuracy as validated in~\cite{fedavg,arm-noniid}; 2) models generalize poorly on unseen data samples, since machine learning is good at interpolation on the data it has been trained on, but bad at extrapolation to the  out-of-distribution data. We discover more subtleties that simple exclusion of the ``outliers'' depending on the degree of class imbalance may undermine model generalization. We need to look into whether the users possess unseen samples that can help generalize the model.

\subsection{Gradient Diversity}
We further investigate this issue through the lens of gradient dissimilarity. Backpropagation relies on gradient descent to minimize the loss objective. High similarity between concurrently processed gradients may cause degraded saturation during gradient descent. A notion called \emph{gradient diversity} is introduced in~\cite{gradient-diversity} for quantifying the difference between gradients from batched data samples. Here, we extend the concept to FL and evaluate the gradient difference between a user and the global model (without the user's gradients). We represent the $n$-layer network as a function $f_i(\bm{w})$, $i\in\{1,\cdots,n\}$. The gradient diversity is,
\begin{equation}
\small
D_{\Delta}(\bm{w}) = \frac{\sum\limits_{i=1}^{n} \| \nabla  f_i(\bm{w}) \|_2^2}{ \| \sum\limits_{i=1}^{n} \nabla  f_i(\bm{w}) \|_2^2 +  \sum\limits_{i=1}^{n} \sum\limits_{i=j} \langle \nabla f_i(\bm{w}), \nabla f_j^G(\bm{w})\rangle }   \label{diversity_eq}
\end{equation}
where the nominator is the sum of $l_2$ norm of the gradients from all the layers, and the second term of the denominator is the inner products of gradients between the local $f$ and global model $f^G$ for each user. If the local gradient has a positive correlation to the global one, $D_{\Delta}(\bm{w}) <1$ and vice versa. If the gradients are orthogonal, the inner product is zero and $D_{\Delta}(\bm{w}) \rightarrow 1$. A larger diversity value means more diverse of the user's data compared to the global model (rest users).

Fig. \ref{non_iid_fig_2}(b) shows the trace of gradient diversity for the 3 cases above (the results are averaged over the number of users shown). Clearly, the outlier user has the highest diversity value with an average above 0.95 compared to the IID baseline at 0.9 (sometimes above 1). In other words, the inner product between the outlier and global updates is approaching zero, which indicates their gradient directions are almost orthogonal. In contrast, the diversity contribution from the rest users with 9 classes are much weaker (Non-IID Rest Users) and excluding the outlier user yields the worst diversity. These results coincide with Fig. \ref{non_iid_fig_2}(a) and suggest gradient diversity as a good explanatory metric: although the outliers may induce local divergence and slightly higher variance in training, they might be conducive to the learning process depending on the gradient directions. The above discussion leaves the door open for further optimizations while jointly considering factors from computation and data distribution. We follow these guidelines to optimize the training time when data distributions are IID and non-IID in the next two sections.

\section{Optimization with IID Data Distribution} \label{sec:design_iid}

When the resource permits, we should strive to pursue IID settings first. The efforts in~\cite{fed-distillation,arm-noniid} attempt to restore the data back into IID. On the other hand, a partial reason of non-IIDness is due to the imperfect data collection process. For example, the collection period is inadequate or a necessary data cleaning/augmentation is missing. With the abundance of mobile data, the scheduler can ask the users to carefully select the data from a sufficiently longer period of time. Meanwhile, the application could also incentivize users to perform those activities that are needed in order to remain IID defined by the task objectives. To this end, we start with the case that the local dataset contains data from all the classes (i.e., IID) and optimize the training time per epoch in this section.

\subsection{System Model} \label{sec:system_model}
The success of machine learning algorithms relies on a large and broad dataset. The goal is to minimize the expected generalization error between training and testing by fitting the distributions of $D$ data. A theoretical bound between the amount of data needed for achieving certain error rates is available in~\cite{haussler}. The FL training task requires a total amount of data $D$, where $D$ can be either obtained empirically or estimated using~\cite{haussler}. We use \emph{shards} to represent the minimum granularity of samples (e.g. 100 samples/shard). The parameter server has sufficient bandwidth and simultaneous transmissions do not cause network congestion or performance saturation~\cite{sysml}. Our framework mainly tackles heterogeneity from computation and data distribution, and is amenable to decentralized topologies without a parameter server~\cite{decentralize_liu}. The users will agree on a protocol to execute training from the demanded amount of data requested by the scheduler. The training data can be a subset of the local data collected during a long time period. For comparison and reproducibility, we follow the same approach as~\cite{fedavg,arm-noniid} to partition public datasets on different devices. We delegate the role of management to the server to gather users' meta data such as smartphone model and information about non-IID class distribution. For simplicity, we assume the server is honest and does not attempt to infer user privacy from the collaborative model or class information as we can always resort to security protocols to protect the intermediate gradients, model and differentially-private class information~\cite{bonawitz}. We formalize the optimization problem for IID data distributions next.

\subsection{Problem Formulation (IID)}
FL follows the stochastic gradient descent to randomly select a number of users for training in each epoch~\cite{fedavg}. Our objective is to optimize the execution time of the selected $n=|\mathcal{N}|$ users. As shown in Section \ref{sec:impact_imbalance}, we can leverage unbalanced local data with minimum accuracy loss as long as the data is IID among the users. This gives enough latitude for task assignments. For all the permutations $\phi$ that partition the total data $D$, the \emph{computation time} $T_i^c(D_i)$ for user $i$ is a function of her data size $D_i$. Depending on the networking environments, the uplink and downlink network latency for user $i$ is a linear function of model size $M$, $T_i^u(M)+T_i^d(M)$. Our goal is to find an optimal assignment of training data so that the maximum processing time is minimized per epoch. The problem is formalized below.
\begin{equation}
\small
\mathbf{P1:}\hspace{0.2in} \min \limits_{D_i \in \phi} \hspace{0.02in} \max\limits_{i \in \mathcal{N}} \big( T_i^c(D_{i})+T_i^u(M)+T_i^d(M) \big) \label{objective_OP}
\vspace{-0.03in}
\end{equation}
\textbf{s.t.}
\vspace{-0.15in}
\begin{eqnarray}
\small
&\sum_{i \in \mathcal{N}} D_{i} = D,  \label{constraint1}\\
&\phi = \{D_1, D_2,\cdots,D_n\}. \label{constraint2}\\
&\sum_{i \in \mathcal{N}} x_i = n \label{constraint3}
\vspace{-0.08in}
\end{eqnarray}
The objective in Eq. \eqref{objective_OP} is to minimize the makespan given all possible data partitions and the assignment of $D_i$ data to user $i$. Eq. \eqref{constraint1} states that the sum of local data should be equal to $D$. Eq. \eqref{constraint2} denotes the permutations of all data partition. For completeness, Eq. \eqref{constraint3} requires all the users to participate in training. The decision variable $x_i=1$ if a user participates; otherwise, it is zero.

$\mathbf{P1}$ can be viewed as a combination of a \emph{partitioning problem} and a variant of the \emph{linear bottleneck assignment problem} (LBAP)~\cite{lbap}. The classic assignment problem finds an optimal assignment of workers to tasks with minimum sum of cost. LBAP is its min-max version. It assigns tasks to parallel workers and ensures the latest worker uses minimum time. We adopt the same analogy here to ensure each training epoch is finished in minimum time. The problem is different from both the classic assignment problem and LBAP. The number of \emph{potential tasks} is not equal to the number of workers (mobile devices), but rather, a much wider potential range to choose from due to the combinatorial partitions of the dataset. The final choice would be determined by the set of constraints that optimizes Eq. \eqref{objective_OP}. A naive solution is to list all the partitions of $D$ in brute force, construct cost values per user for all the potential permutations, solve an LBAP and find the assignment with the minimum makespan. For a total number of $s$ shards, the possible permutations are in the order of $s^n$, which makes it intractable even for small $n$.

\subsection{Joint Partitioning and Assignment}
Though the naive method turns out to be futile in polynomial time, the following property of mobile devices helps simplify the problem.

\noindent \textbf{Property 1.} For data $D_i$, $T_i^c(D_{i})+T_i^u(M)+T_i^d(M)$ is a non-decreasing function.

Then it is not necessary to test a large number of potential partitions, if a partition of smaller size has already satisfied Eq. \eqref{constraint1} with less computation time. For example, consider possible permutations of $\sum_{i=1}^{3} D_i=13$ among three users. If the first or the second user is the straggler in partition $(4,4,5)$, then partitions such as $(5,5,3)$, $(6,6,1)$ definitely leadnA to more running time. This allows us to potentially skip a large number of sub-optimal solutions.

The classic LBAP has a polynomial-time thresholding algorithm in $\mathcal{O}(n^{\frac{5}{2}}\log n)$~\cite{lbap}. This algorithm checks whether a \emph{perfect matching} exists in each iteration using the Hopcroft-Karp algorithm, that takes $\mathcal{O}(n^\frac{5}{2})$. Here, when $D$ is divided into $s$ shards, perfect matching between user and data shard is no longer needed as introduced in the following property.

\noindent \textbf{Property 2.} A bipartite graph $\mathcal{G}=(\mathcal{U},\mathcal{V};\mathcal{E})$ can be constructed with $|\mathcal{U}|= n$, $|\mathcal{V}|= \phi$ and edges $(i,j) \in \mathcal{E}$. Each vertex in $\mathcal{U}$ should have degree of $1$ and vertices in $\mathcal{V}$ can have degree of $0$ (as long as the sum of vertices having degree $1$ equals $D$).

\begin{figure}[t!]
\centering
\includegraphics[width=3.6in]{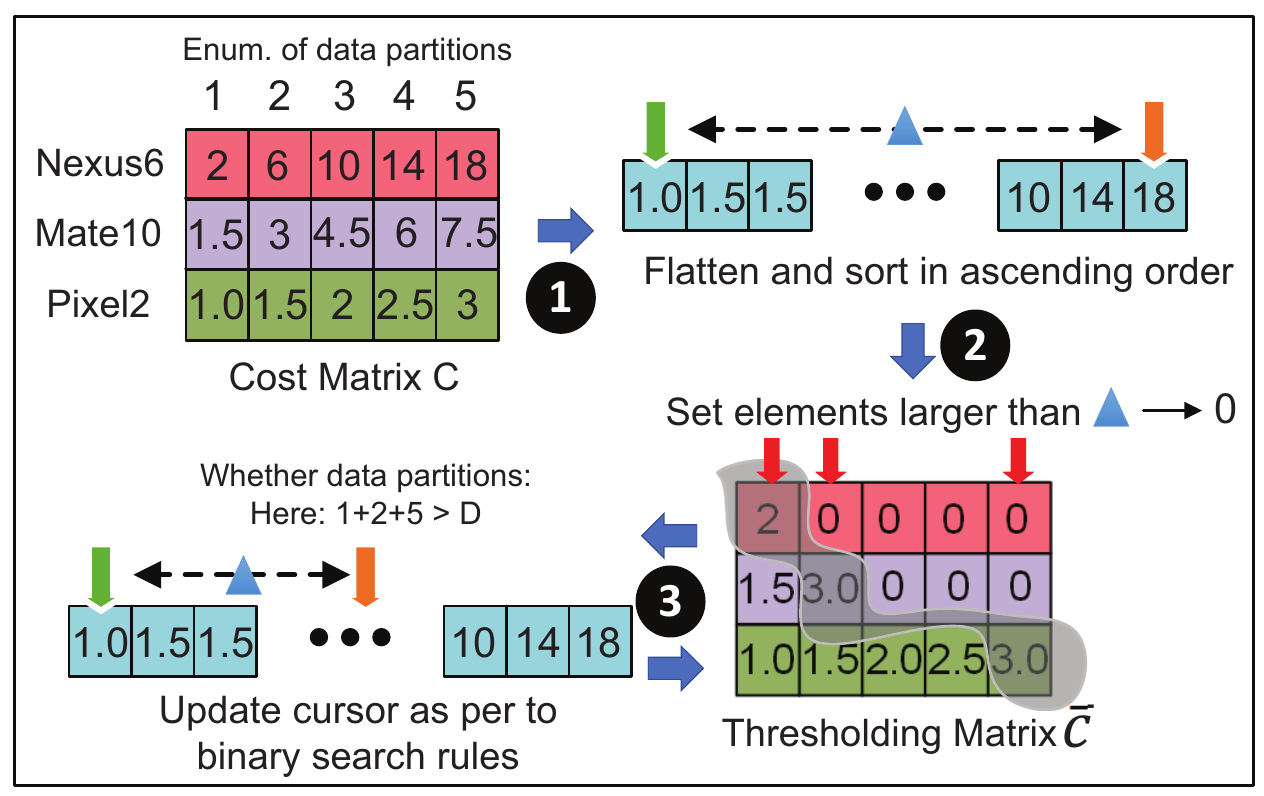}
\caption{Illustration of the procedures in Fed-LBAP: \ding{182} cost matrix $C$ with columns representing the enumeration of data partitions; \ding{183} flatten $C$ and sort elements in ascending order; \ding{184} set elements larger than the threshold in $\overline{C}$ to zero and performs binary search to find the optimal threshold.}
\label{BLAP_fig}
\end{figure}

\noindent \textbf{Fed-LBAP Algorithm.} Based on Properties 1 and 2, we can further reduce the time complexity by extending~\cite{lbap}. We propose a joint partitioning and assignment algorithm to solve the problem in polynomial time. The procedure is explained in Fig. \ref{BLAP_fig}. For the $n$ users, we define a cost matrix $C=\{c_{ij}\}$ of dimension $n \times s$ (i.e., the matrix represents the cost to assign $j$ shards to user $i$). A \emph{thresholding matrix} $\overline{C}$ with the same setting is also initiated. We sort all the elements from the cost matrix in ascending order and perform binary search by utilizing a threshold $c^\ast$: if $c_{ij}>c^\ast$, $\overline{c_{ij}}=0$; otherwise, $\overline{c_{ij}}=1$. The sum of all cost values found in each iteration is compared to $D$. If larger, find a new median for the left half; otherwise, find a new median for the right half until the optimal median value is reached. In short, our algorithm first performs sorting of all the cost values and conducts a binary search for the minimal threshold $c^\ast$ such that \emph{Property 2} and Eq. \eqref{constraint1} hold. The procedures are summarized in Algorithm \ref{fed_learning_blap}.


The time complexity is analyzed below. In the worse case, binary search takes $\mathcal{O}(\log ns)$ iterations. We need to check whether $\overline{c_{ij}}=0$ during the iterations. This takes $\mathcal{O}(s)$ time for one user and is repeated for $n$ times. The time complexity is $\mathcal{O}(ns \log ns)$ with $s\geq n$. To be consistent with~\cite{lbap}, when $s=n$, our algorithm is $\mathcal{O}(n^2 \log n)$.


\noindent \textbf{Analytical Solution (Linear Case).} When the training time $T_i^c(\cdot)$ has a linear relationship with the number of data shards, the problem has a (relaxed) analytical solution. Denote the function by $T_i$ in short and data size $D_i$ of a user $i$, we have
\begin{equation}
\small
D_i=T_i/a_i-b_i/a_i, \label{linear}
\end{equation}
where $a_i$ and $b_i$ are device and model-specific parameters found by the profiler discussed in Sec. \ref{sec:profiling}.

\noindent \textbf{Property 3.} If the integer requirements of data shards are relaxed, for a solution to be optimal, the training time is equivalent on all the mobile devices.

\begin{proof}
We prove this property by contradiction. Assume the optimal solution $T^\ast$ is reached when all users have the same training time of $T$, except a user $j$ that takes $T+\triangle T$. On the other hand, we can always reduce $D_j$ by $\triangle D$ and proportionally increase the rest users by $ \triangle D \cdot r_i$, $\forall i \in \mathcal{N} \backslash j$, such that $T_i=T_j=T' < T+\triangle T$, where $\sum_{i \in \mathcal{N} \backslash j} r_i = 1.$ This results a contradiction with $T^\ast=T+\triangle T$, thus property 3 holds.
\end{proof}

Based on Property 3, we replace $T_i$ in \eqref{linear} with $T^\ast$ and take summation over all the users on both sides, we obtain the optimal solution and data partitions\footnote{The data partitions are derived by plugging the optimal time into Eq. \eqref{linear} and treating $T_i=T^\ast$.},
\begin{equation}
\small
T^\ast = \frac{D + \sum_{i \in \mathcal{N}} \frac{b_i}{a_i}}{\sum_{i \in \mathcal{N}} \frac{1}{a_i} } , D_i =  \frac{D + \sum_{i \in \mathcal{N}} \frac{b_i}{a_i}}{\frac{1}{a_i}\sum_{i \in \mathcal{N}} \frac{1}{a_i} } - \frac{b_i}{a_i} \label{optimal_linear}
\vspace{-0.08in}
\end{equation}

We reuse notation $T^\ast$ as the optimal solution derived by the relaxed solution in $\mathbb{R}$. By rounding off the number of data shards to integers, we can obtain an $(1+\epsilon)$ approximation, and integral gap is bounded by $\epsilon=\max(a_1,a_2,\cdots,a_n)$. The analytical solution reduces the time complexity to $\mathcal{O}(n)$ for computing the assignment and $\mathcal{O}(1)$ for calculating the training time. Note that it only holds for the linear case. For some mobile devices that exhibit superlinear or even weak quadratic running time (due to thermal throttling), we refer to the general mechanism of Algorithm \ref{fed_learning_blap}.

%
%
%


\setlength{\textfloatsep}{0pt}
\begin{algorithm}[t!]
\caption{Fed-LBAP (for IID data)}
\label{fed_learning_blap}
\small
\textbf{Input:} Total data size $D$, cost matrix $C=\{C_{ij}\}$, number of users $n$.\\
\textbf{Output:} The assignments of tasks $\{A_j\}$ for each user $j$.

$\overline{C}\leftarrow$ $C$ sorted in the ascending order.

$\text{min}\leftarrow 0$, $\text{max}\leftarrow |\overline{C}|$,  $\text{median}\leftarrow \lfloor\frac{\min+\max}{2}\rfloor$; $D'\leftarrow 0$

\While{$\min<\max$}
{
$C^{\ast}\leftarrow \overline{C}(\text{median})$

\For{$j=1\;\text{to}\;m$}
{
$A_j \leftarrow \arg\max_j \{C_{ij}| C_{ij}\leq C^{\ast}\}$

$D'\leftarrow D'+A_i$
}

\If{$\forall i, A_i=0$ or $D'<D$}
{
$\text{min}\leftarrow\text{median}$
}
\Else
{
$\text{max}\leftarrow\text{median}$
}
}
\end{algorithm}

%
%
%
%

\section{Optimization with Non-IID Data Distribution}  \label{sec:non_iid}
Non-IIDness is inherent in mobile applications due to the diverse behaviors and interests from users. This section studies the situations when we are unable to restore the distribution back to IID.

\subsection{Problem Formulation (Non-IID)}
To connect non-IIDness with the scheduling decisions, we introduce an accuracy cost $\alpha^{w_i}$ with a base parameter $\alpha$ to the power of a weight $w_i$ for user $i$. $\alpha$ balances the makespan and the potential convergence time. For the same weight, a large $\alpha$ weighs more on those non-IID outliers and possibly excludes them from selection. A small $\alpha$ weighs less on the data distributions and focuses more on the makespan. Its value is determined empirically in Section \ref{alpha:sec} and the construction of the weight value $w_i$ is discussed in the next subsection. We formulate the optimization problem first. The new objective is to find a schedule with the minimum average cost.
\begin{equation}
\small
\mathbf{P2:}\hspace{0.2in} \min \sum_{i \in \mathcal{N}} \big(T_i^c (D_i) + \big( T_i^u(M)+T_i^d(M)+ \alpha^{w_i} \big) y_i\big) \label{obj_2}
\vspace{-0.03in}
\end{equation}
\hspace{0.2in}\textbf{s.t.}
\vspace{-0.15in}
\begin{eqnarray}
\small
& \sum_{i \in \mathcal{N}} D_{i} = D,  \label{constraint2_1}\\
& D_i \leq U_i, i \in \mathcal{N} \label{constraint2_2} \\
&y_i = \mathbbm{1} \bigl(D_i > 0) \label{constraint2_3}
\vspace{-0.08in}
\end{eqnarray}
We re-use most of the notations from $\mathbf{P1}$ and assume an initial equal partition of $D$ among the users, but to be adjusted afterwards. The new objective is to determine the data shards $D_i$ to be assigned to user $i$ such that the sum of computation/communication and cost of accuracy (scaled by $\alpha$) is minimized. We can consider the accuracy cost as a fixed cost when a user is involved, which gradually changes defined by Eq. \eqref{acc_cost} later. Constraint \eqref{constraint2_1} ensures that all the data partitions sum up to $D$ in total. Constraint \eqref{constraint2_2} states that the size of data does not exceed user $i$'s capacity $U_i$, which can be quantified by storage or battery. Constraint \eqref{constraint2_3} makes $y_i$ equal to $1$ if user $i$ is selected; otherwise, $y_i$ is $0$. In addition to the accuracy cost, the difference between $\mathbf{P1}$ and $\mathbf{P2}$ is that $\mathbf{P2}$ allows the users to be deselected because of high overall cost.


\subsection{Accuracy Cost}
Given the disparity of class distributions, user selection is vital to the computation time and accuracy. One may use the previous LBAP algorithm to weigh more on those devices with higher processing power. But if those users are non-IID outliers, they may adversely prolong global convergence, though each epoch is time-optimized. On the other hand, the study in Sec. \ref{sec:impact_non_iid} suggests further look into those outliers: if a class is not yet included in the population, inclusion is beneficial to convergence and model generalization; while at the same time, it is necessary to screen non-contributing outliers out of the population. The design behind this new \emph{accuracy cost} is to encourage/penalize the assignments that empirically lead to accuracy gain/loss. As mentioned, we utilize a parameter $\alpha$, increased to the power of $w_j$, so users' accuracy cost is sufficiently distinctive regarding their class distributions. Denote the class set of each user as $\mathcal{C}_i$ and all classes as $\mathcal{C}$. In most cases, the weight should be inversely proportional to the number of classes of a user,
\begin{equation}
\small
w_j = |\mathcal{C} - \mathcal{C}_i|. \label{weight_eq}
\end{equation}
We also define the lowest weight $w_l (w_l \leq w_i, \forall i \in \mathcal{N})$ ($w_l=|\mathcal{C}| - \max_{i\in\mathcal{N}} |\mathcal{C}_i| $ in experiment, where the second term is the maximum number of classes a user has). Eq. \eqref{weight_eq} can be illustrated by an example. Using MNIST as an example, an outlier user with only class $\{7\}$ has cost $\alpha^9$ and another user with classes $\{2,5,6,8,9\}$ has cost $\alpha^5$, and $\alpha^9 > \alpha^5$ when $\alpha$ is larger than 1. It captures the general case that the cost grows with a reducing number of classes. However, when the intersection between class set of user $i$, $\mathcal{C}_i$ and population $\mathcal{P}_i$ is empty, $\mathcal{C}_i \cap \mathcal{P}_i = \emptyset$, $i$'s classes are not present in the population ($\mathcal{P}_i = \bigcup_{j\in \mathcal{N}, j\neq i} \mathcal{C}_j$). The weight should take a small value, so we set $w_i = w_l$ to encourage these contributing outliers.

In practice, we should implement the above design more carefully. Consider a special case with two users having only class $\{7\}$, which also happens to be the only users with this class. Yet, viewing from either one of them, they would think that the population has already included $\{7\}$ so their weights are set to $\alpha^9$ together, causing the algorithm to exclude both of them from the population. Then the population cannot learn from this class and the accuracy is greatly undermined. Obviously, coordination is needed but including both of them is unnecessary. After the first user has been included, the second one becomes an outlier since the class has been covered by the first user already. Our experiment suggests that these contributing outliers having the same classes are mutually exclusive: introducing the second user would have negative impact on training. Hence, we keep the number small by assigning a higher cost to the subsequent users. More formally, when $\mathcal{C}_i = \mathcal{C}_j$, we set $w_i=w_l$ and pursue a higher cost for $w_j = |\mathcal{C} - \mathcal{C}_j|$. The strategy is summarized as,
\begin{equation}
\small
\label{acc_cost}
\hspace{-0.05in}
w_i =\left\{
\begin{split}
& |\mathcal{C}| - \max_{i\in\mathcal{N}} |\mathcal{C}_i| , \;\;\;\;\;\;\;\;\;\;\;\;\;\; \mathcal{C}_i \cap \mathcal{P}_i = \emptyset \\
& |\mathcal{C} - \mathcal{C}_i|, \;\;\;\;\;\;\;\;\;\;\;\;\;\;\;\;\;\;\;\;\;\;\; otherwise
\end{split}
\right.
\end{equation}
If $\mathcal{C}_i \subseteq \mathcal{P}_i$ and $\exists j \in \mathcal{N} \setminus i$, such that $\mathcal{C}_j = \mathcal{C}_i$, overwrite $w_i = |\mathcal{C}| - \max_{i\in\mathcal{N}} |\mathcal{C}_i|$ in Eq. \eqref{acc_cost} and $w_j$ remains $|\mathcal{C} - \mathcal{C}_j|$.



%
%
%
%

\subsection{Min Average Cost Algorithm}
After the problem is fully formulated, we can see that the previous min-max problem is converted into a min average cost problem, which is in close analogy to the \emph{bin packing problem with item fragmentation}~\cite{bpf1}. The problem finds an assignment of items to a fixed number of bins by splitting them into fragments. For each fragmentation, there is an associated unit cost. In our scenario, the items correspond to the learning tasks splittable into data shards and the users represent the bins. Unlike the original bin packing, the objective no longer minimizes the number of users (with unit cost); instead, it is characterized by the functions of computation time and accuracy cost. The fragmentation cost is also different from the unit cost in~\cite{bpf1}. It actually depends on which destined user the fragments are assigned to. If the user has been already involved in training (bin/user is open), the cost depends on the increment of computation time from the new fragments, plus the initial cost of accuracy as described next.

We propose the \emph{Min Average Cost Algorithm} to tackle the problem. The main idea is to iteratively assign the data shards to the user with the minimum average cost in a greedy fashion. Consider the dataset of $D$ data shards and $n$ users. The initial cost is $T_i(d)+ \alpha^{w_i}$, if a user $i$ is open for training with $d$ data (omit the communication cost here for clarity). Starting from $i$ with the lowest initial cost, we assign $d_1=d$ to $i$. Denote the set of users that are already involved in training as $\mathcal{O} \subseteq \mathcal{N}$. For $d_2=d$, we compare the cost by either assigning it to $i$ with cost $T_i(2d) + \alpha^{w_i}$, or to $j$ with cost $T_j(d) + \alpha^{w_j}$ ($j \in \mathcal{N}\setminus\mathcal{O}$), and select the one with less cost. For all the users $i \in \mathcal{O}$ and a potential user $j \in \mathcal{N}\setminus\mathcal{O}$, we assign $d$ according to,
\begin{equation}
\small
i^{\ast} = \mathop{\arg\min}_{i\in \mathcal{O}, j \in \mathcal{N}\setminus\mathcal{O}} \bigl\{ T_i\bigl((l_i+1)\cdot d\bigr)+\alpha^{w_i}, T_{j}(d) + \alpha^{w_j} \bigr\}, \label{min_cost}
\end{equation}
where $l_i$ is the current number of data shards of user $i$. If all the users are involved ($i, j \in \mathcal{O}$), we compare $T_i\bigl((l_i+1)\cdot d\bigr)+\alpha^{w_i}$ with $T_j\bigl((l_j+1)\cdot d\bigr)+\alpha^{w_j}$ and select the one with less cost. If $i$ reaches the capacity that $l_i \cdot d \geq U_i$, it is excluded from further selections (bin is closed); otherwise, it remains open. The algorithm repeats until $D$ is exhausted and runs in $\mathcal{O}(m n)$ time, where $m$ is much larger than $n$. The procedure is summarized in Algorithm \ref{minavg}.

\setlength{\textfloatsep}{0pt}
\begin{algorithm}[t!]
\caption{MinCost Algorithm (for Non-IID Data)}
\label{minavg}
\small
\textbf{Input:} Number of data shards $D$ and size $d$ per shard, $\mathcal{N}$ users, cost profiles $T(\cdot)$, class coverage $\mathcal{C}_i$ and population viewing from $i$, $\mathcal{P}_i = \bigcup_{j\in \mathcal{N}, j \neq i} \mathcal{C}_j$, user coverage $\mathcal{O}$, parameters $\alpha$, number of data shards $l_i$ for user $i$.

\textbf{Output:} Data assignment for each user $l_i$.

Initialize $\mathcal{O} \leftarrow \emptyset$, $d=1$.

$\forall i \in \mathcal{N}$ \If{$\mathcal{C}_i \cap \mathcal{P}_i = \emptyset$ }
{
$  w_i \leftarrow |\mathcal{C}| - \max_{i\in\mathcal{N}} |\mathcal{C}_i|.$
}
\Else
{
    $w_i = |\mathcal{C} - \mathcal{C}_i|$.

    \If{$\mathcal{C}_i \subseteq \mathcal{P}_i$ and $\exists j \in \mathcal{N} \setminus i$, $\mathcal{C}_j = \mathcal{C}_i$}
    {
        $w_j = |\mathcal{C} - \mathcal{C}_j|$.
    }
}
\While{$d<D$}
{
\If{$\mathcal{N}\setminus\mathcal{O} \neq \emptyset$}
{
$i \leftarrow \mathop{\arg\min}\limits_{i\in \mathcal{O}, j \in \mathcal{N}\setminus\mathcal{O}} \bigl\{ T_i\bigl((l_i+1)\cdot d\bigr)+\alpha^{w_i}, T_{j}(d) + \alpha^{w_j} \bigr\}.$
}
\Else
{
$i \leftarrow \mathop{\arg\min}\limits_{i,j\in \mathcal{O}} \bigl\{ T_i\bigl((l_i+1)\cdot d\bigr)+ \alpha^{w_i}, T_j\bigl((l_j+1)\cdot d\bigr)+ \alpha^{w_j} \bigr\}.$
}
$l_i \leftarrow l_i + 1.$\\
\If{$l_i \geq U_i$}
{
$w_i \leftarrow \infty$
}
$\mathcal{O} \leftarrow \mathcal{O} + i$, $\mathcal{N} \leftarrow \mathcal{N} - i$, $d \leftarrow d + 1$.

}
\end{algorithm}

\begin{figure}[!ht]
\centering
\vspace{-0.08in}
\includegraphics[width=3.65in]{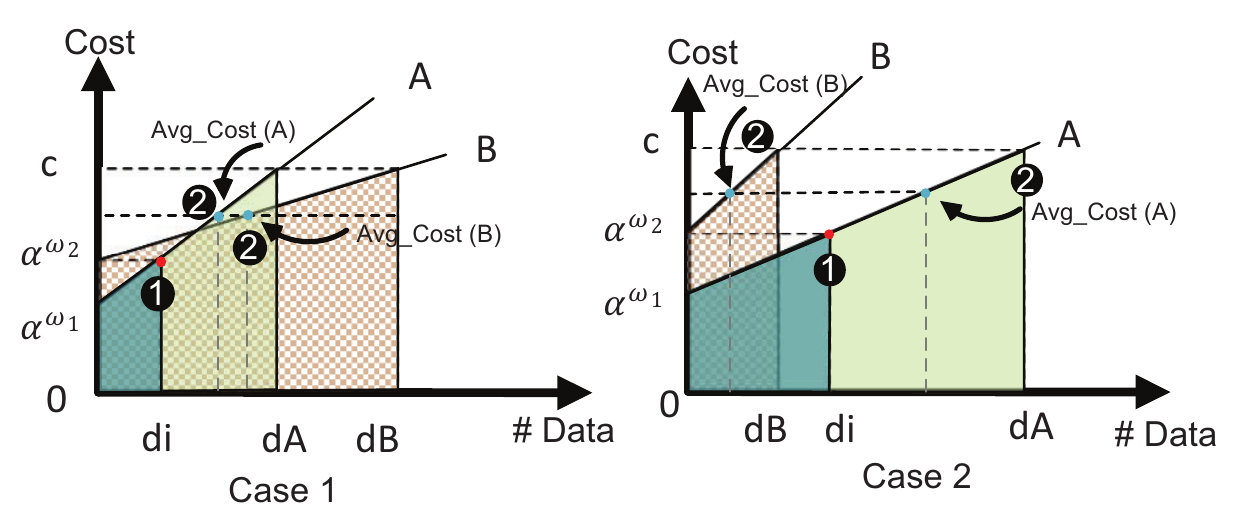}
\vspace{-0.1in}
\caption{Visualize solutions of the Mincost algorithm when cost function is linear.}
\label{bin_analytical}
\vspace{-0.05in}
\end{figure}

\textbf{Analytical Solution.} Similar to the IID case, when the cost function is linear, the solution process can be visualized analytically. We illustrate with the simplest case of two users and their cost functions are, $y_1(x) = b_1 x + \alpha^{w_1}$, $y_2(x) = b_2 x + \alpha^{w_2}$, represented by lines $A$ and $B$ in Fig. \ref{bin_analytical}. $x$ is the number of data shards. $y$ is the cost and $b_1, b_2$ are the slopes representing the computational capacity of the devices. Recall that $\alpha^{w_1}, \alpha^{w_2}$ are defined as the accuracy cost of the users in Eq. \eqref{acc_cost} based on their class distributions. Since it does not change with the number of data so $\alpha^{w_1}$ and $\alpha^{w_2}$ can be treated as constants here (the intercept on the y-axis in Fig. \ref{bin_analytical}). Shown in Fig. \ref{bin_analytical}, $A$ either has lower initial cost and climbs faster than B, or lower cost than $B$ throughout. In both scenarios, the strategy is the same: \ding{182} data is assigned to the one with lower initial cost until the current cost equals the initial cost of the other user (at $d_i$ in Fig. \ref{bin_analytical}); \ding{183} alternate between the two users with the data assignment in proportion to the ratio of slopes. If $b_1 > b_2$, assign $\lfloor \frac{b_1}{b_2} \rfloor$ data to $B$ for every one unit of data to $A$, and vice versa. The average cost indicated by the median line is equal for $A$ and $B$, i.e., the assignment does not stop if the current cost to both $A$ and $B$ is not equal. The process is illustrated in Fig. \ref{bin_analytical}, which can be generalized to multiple users. We omit it due to space limit.

%


\section{Profiling Device Heterogeneity}   \label{sec:profiling}
The optimization algorithm relies on the estimation of computation time using a function $T_i^c(D_i)$ given the neural network model $M$, when the training data is $D_i$ for user $i$. In this section, we develop a method that the server can leverage to build profiles for the participants. In practice, profiling can be done either online through a bootstrapping phase or offline measured on users' devices. The objective is to estimate the training time given the \emph{model parameters} and \emph{data size}, where both of them hold linear relationships with the training time in VGG-type networks. Thus, we take a two-step approach to first profile the computation time regarding model parameters while fixing the data size. Since the convolutional layers have higher computation intensity, their parameters are separated from the dense layers. We profile a number of $k$ different model architectures and their training time of $d$ data, denoted by, $\mathbf{y}^{(d)}=[y_1,y_2,\cdots,y_k]^{(d)}$. $\mathbf{x}_i^{(d)}=[x_{i,1},x_{i,2}]^{(d)}$ are the number of parameters for convolution and dense layers of different models. We employ a \emph{multiple linear regression model},
\begin{equation}
\small
y_i = \alpha_0 + \sum_{j=1}^{2}\alpha_j x_{i,j} + e_i, \label{mlr1}
\end{equation}
where $e_i$ is a noise vector to compensate measurement error. The parameters are found by solving the least square problem, $\hat{\beta}=\mathbf{y}\cdot \mathbf{X}^{-1}$, which is computed by $\hat{\alpha}=\underset{\alpha}{\mathrm{argmin}} \norm{\mathbf{y} - \beta \mathbf{X}}_2^2$. The output of the first step is $\{\alpha_0,\alpha_1,\alpha_2\}^{(d)}$ for different $d \leq D$. With an unknown model architecture, the first step provides $d$ estimates $[\hat{y_1},\hat{y_2},\cdots,\hat{{y_d}}]$ of computation time. The second step extends the estimates from the first step for unknown data sizes by applying (linear) regression again to fit the estimations. We evaluate this method next and derive $\{\alpha_0,\alpha_1,\alpha_2\}$ for different mobile devices.

\vspace*{-0.15in}
\section{Evaluation} \label{sec:eval}

In this section, we evaluate the proposed algorithms on a testbed of various combinations of mobile devices using two public datasets. The main goals is to investigate the effectiveness of: 1) the profiling method; 2) the proposed algorithms for both IID and non-IID cases in terms of computation, accuracy and convergence.



\textbf{Mobile Development.} The mobile framework is developed in DL4J~\cite{dl4j}, a java-based deep learning framework that can be seamlessly integrated with Android. Training is conducted using multi-core CPUs enabled by OpenBLAS in Android 8.0.1. We use \texttt{AsyncTask} to launch the training process by the foreground thread with the default \texttt{interactive} governor and scheduler. To avoid memory error, we enlarge the heap size to $512$ MB by setting \texttt{largeHeap} and use a batch size of $20$ samples. This allows us to train VGG-like deep structures.

\textbf{Experiment Setting.} We use the collection of devices to construct five combinations of mobile testbeds as shown in Table \ref{testbed_combination}. The experiment is conducted on two commonly used datasets: MNIST~\cite{mnist} and CIFAR10~\cite{cifar} with 60K and 50K training samples. We fully charge all the devices, pre-load both datasets into the mobile flash storage and read them in mini-batches of 20 samples. Users perform one epoch of local training in each round and the global gradient averaging iterates 50 and 100 epoches for MNIST and CIFAR10 respectively.

To emulate the dynamics of mobile data, we generate random distributions among the users: 1) For IID data, each user retains all the classes and the ratio between samples from different classes is equivalent; 2) For non-IID data, each user has a random subset of classes and each class may also have different number of samples. We set the maximum number of classes in the subset to $7$ (out of the total 10 classes), i.e., on average, a user would have about $3.5$ classes. The purpose is to see whether our algorithm can handle various random cases of non-IID distributions. Two fundamental networks of LeNet~\cite{lenet} and VGG6~\cite{vgg} are evaluated and their efficiency has been proved to handle learning problems at sufficient scales. To meet the input dimensions, we tailor the original 16 layers of VGG16 by stacking five $3\times3$ convolutional layers with one densely connected layer. We set parameter $\alpha$ in Mincost to 1.8 and 2.45 for LeNet and VGG6 empirically as discussed in Section \ref{alpha:sec}. The uplink and downlink latencies are added to computation time.

\textbf{Benchmarks.} The proposed algorithms are compared with several benchmarks: 1) Proportional: a simple heuristic that assigns training data proportional to the processing power of mobile devices statically measured by their max CPU frequencies; 2) Random: random data partitions among the users; 3) FedAvg~\cite{fedavg}: assign equal shares of data to users. Since the model architecture is fixed, we mainly compare the computational time and treat the communication time as a constant. To facilitate the evaluation, we also adopt pytorch with GTX1080/K40 GPUs to evaluate different benchmarks.

\begin{table}
\small
\begin{center}
\begin{tabular}{c c c c c}
\hline
model & SoC & CPU  & big.LITTLE\\
\hline
Nexus 6 & Snapdragon 805& 4$\times$2.7GHz  & \xmark \\
\hline
Nexus 6P & Snapdragon 810 & \makecell{4$\times$1.55 GHz\\4$\times$2.0 GHz}  & \cmark \\
\hline
Samsung J8 & Snapdragon 450& 8$\times$1.8GHz  & \xmark \\
\hline
Mate 10 & Kirin 970 & \makecell{4$\times$2.36GHz\\ 4$\times$1.8GHz}  & \cmark  \\
\hline
Pixel2 & Snapdragon 835 & \makecell{4$\times$2.35 GHz\\ 4$\times$1.9 GHz} & \cmark  \\
\hline
P30 Pro & Kirin 980 & \makecell{2$\times$2.6 GHz\\ 2$\times$1.92 GHz \\ 4$\times$1.8 GHz} & \cmark  \\
\hline
\end{tabular}
\caption{Hardware configurations of benchmarking testbed.}
\label{table:testbed}
\end{center}
\end{table}

\begin{table}
\small
\centering
  \begin{tabular}{c c c c c c c c}
    \hline
    {}  & N6  & N6P  & J8  & Mate10 & Pixel2 &P30 &Total   \\
    \hline
    T1  & 1 & - & - & 1 & 1 &- &3\\

    T2  & 2 & 2 & - & 1 & 1 &- &6   \\

    T3  & 4 & 2 & - & 2 & 2 &- &10   \\

    T4  & 6 & 2 & 1 & 2 & 2 &1 &14   \\

    T5  & 8 & 3 & 2 & 2 & 3 &2 &20   \\
    \hline
  \end{tabular}
\caption{Experimental mobile testbeds}
\label{testbed_combination}
\vspace{-0.05in}
\end{table}

\begin{table}[!ht]
	\begin{minipage}{0.5\linewidth}
		\includegraphics[width=1.0\textwidth]{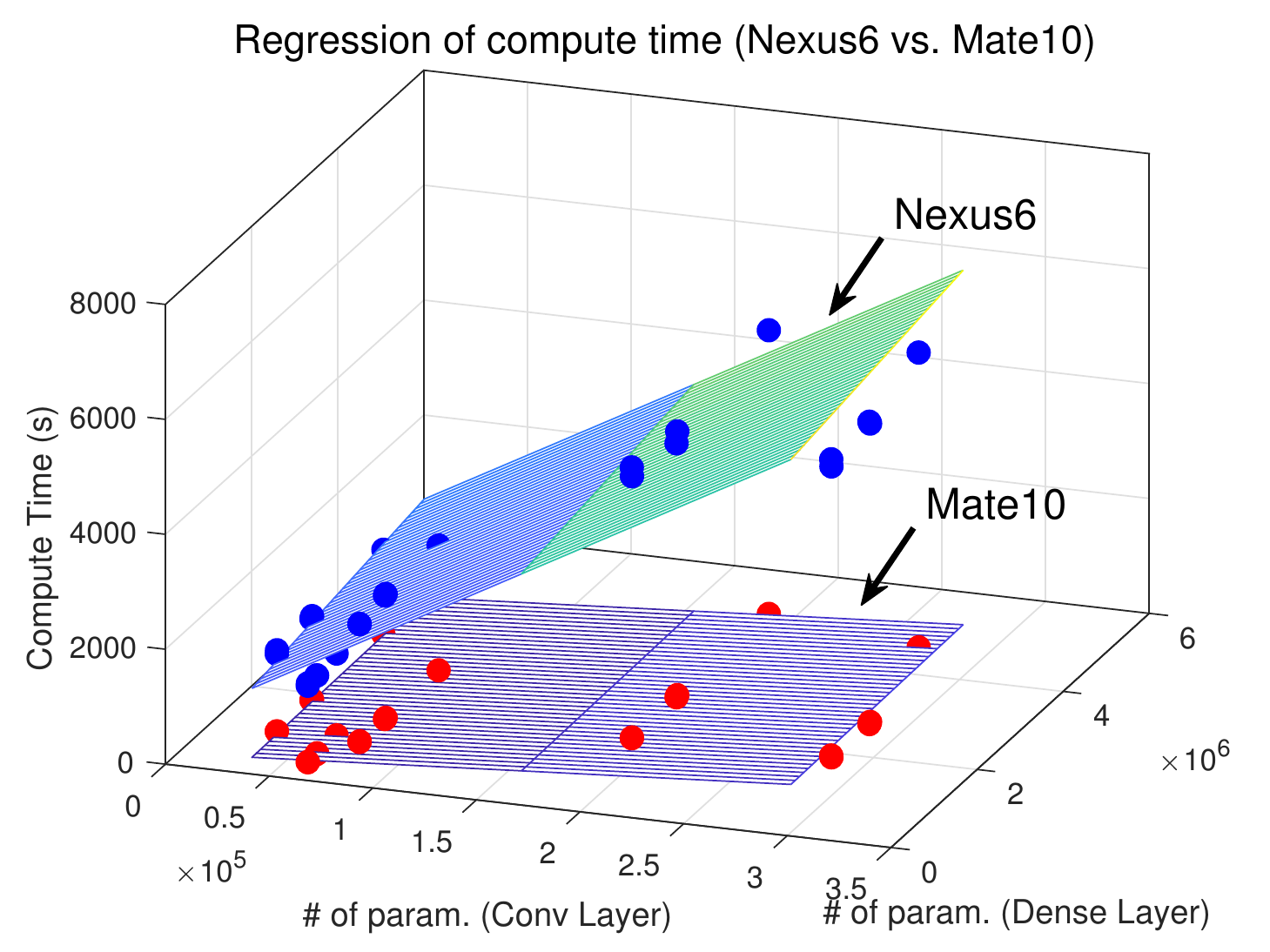}\hspace{0.1in}
		\captionof{figure}{Step 1, profile training time with model parameters.}
		\label{profiling_fig_1}
	\end{minipage}
    \begin{minipage}{0.5\linewidth}

		\centering
		\begin{tabular}{cc}
			\hline
			Device      & $[\alpha_0,\alpha_1,\alpha_2]$ \\
			\hline
			Nexus6       & [578, 0.02, 2e-5]    \\
			Nexus6P      & [647, 8e-3, 3e-4]    \\
			SJ8     & [183, 1e-2, 9e-5]     \\
			Mate10       & [47, 2e-3, 2e-5]     \\
			Pixel 2      & [68, 2e-3, 1e-5]  \\
            P30         & [42, 2e-3, 1e-5] \\
			\hline
		\end{tabular}
    \caption{Learned parameters (Eq. \ref{mlr1}): $\alpha_0$, intercept; $\alpha_1$, conv. layer; $\alpha_2$, dense layer.}\label{table:profile_param}
	\end{minipage}\hfill

\end{table}

\begin{figure}
\vspace*{-0.09in}
\centering
\hspace*{-0.22in}
\begin{subfigure}[b]{0.25\textwidth}
                \includegraphics[width=1.1\textwidth]{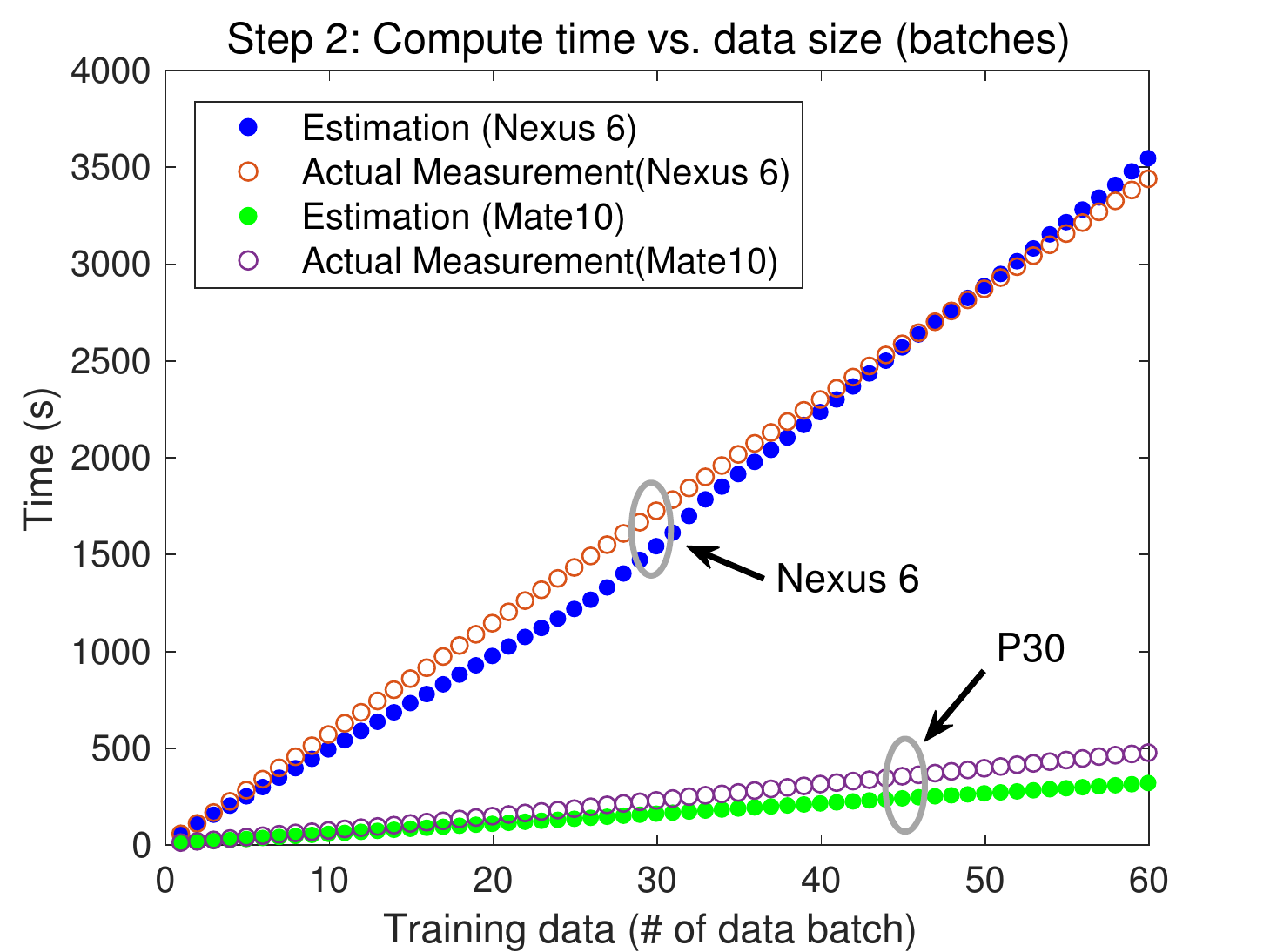}
                \vspace{-0.2in}
                \caption{}
\end{subfigure}
\hspace*{-0.02in}
\begin{subfigure}[b]{0.25\textwidth}
                \includegraphics[width=1.1\textwidth]{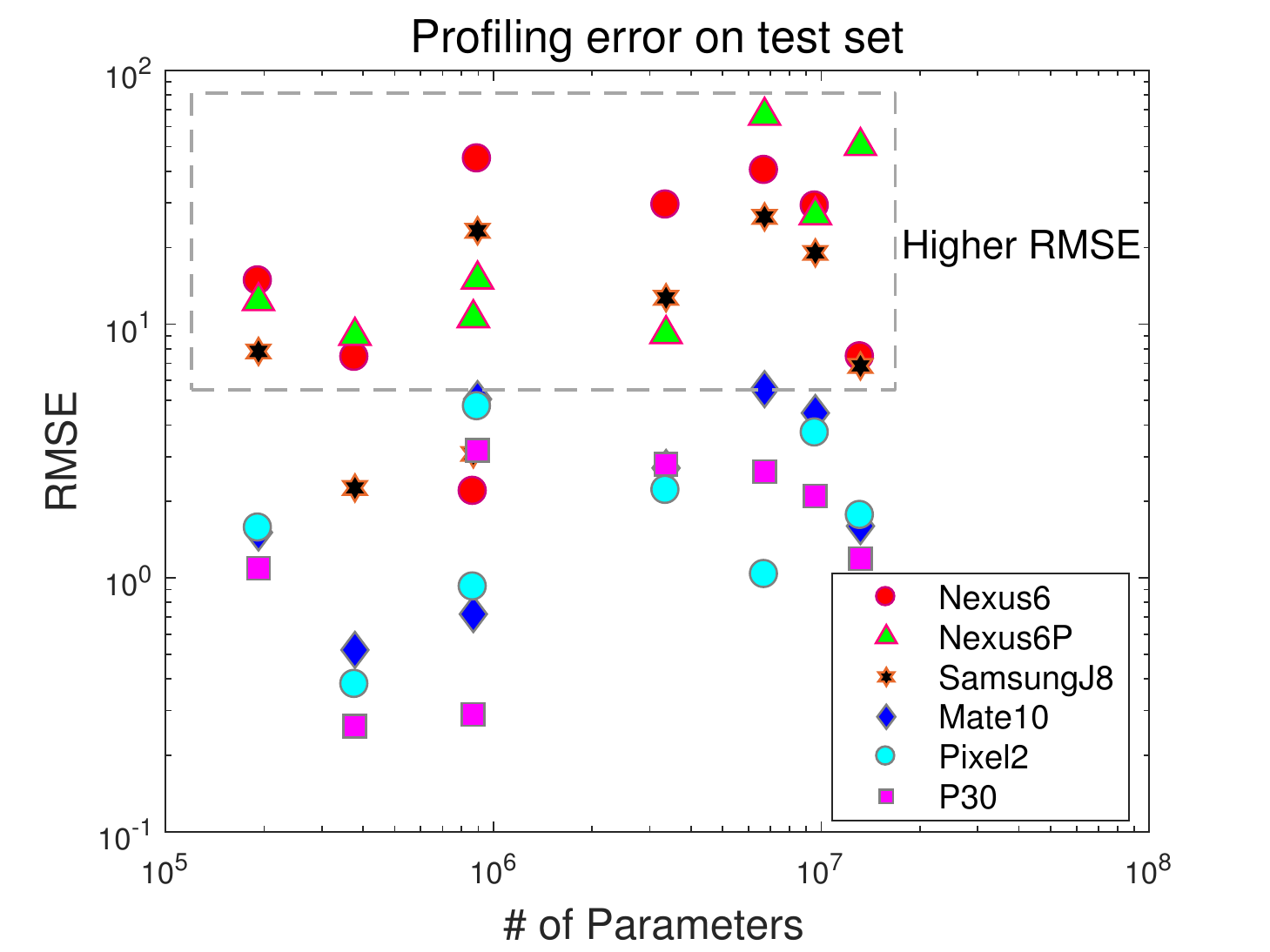}
                \vspace{-0.2in}
                \caption{}
\end{subfigure}
\hspace*{0.05in}
\vspace{-0.1in}
\caption{Step 2: predict training time vs. data size. a) Nexus 6 vs. P30; b) profiling error on test set. }
\label{profiling_fig_2}
\end{figure}

\begin{figure*}[!ht]
\vspace*{-0.09in}
\centering
\hspace*{-0.05in}
\begin{subfigure}[b]{0.24\textwidth}
                \includegraphics[width=1.03\textwidth]{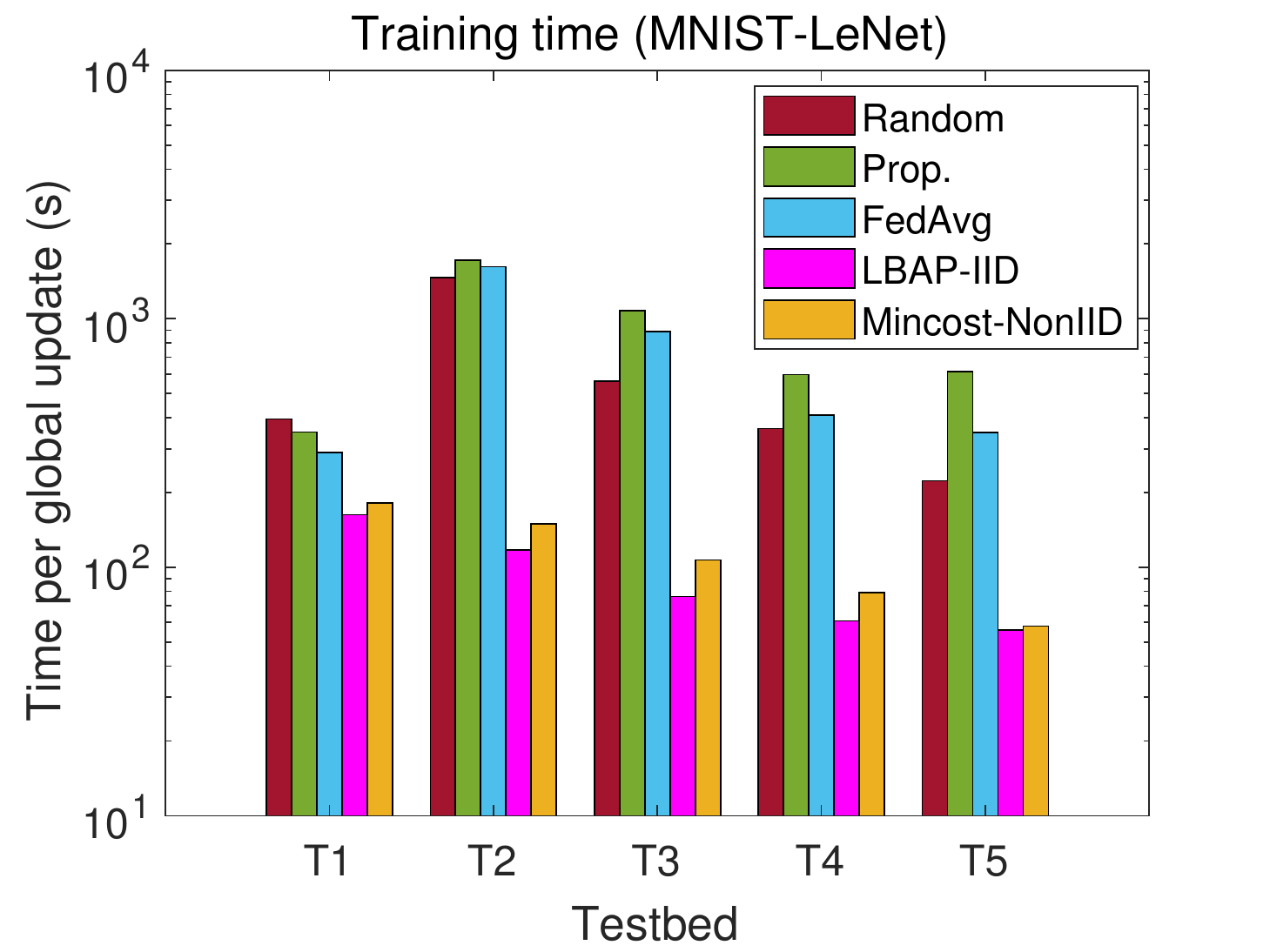}
                \vspace{-0.21in}
                \caption{}
\end{subfigure}
\begin{subfigure}[b]{0.24\textwidth}
                \includegraphics[width=1.03\textwidth]{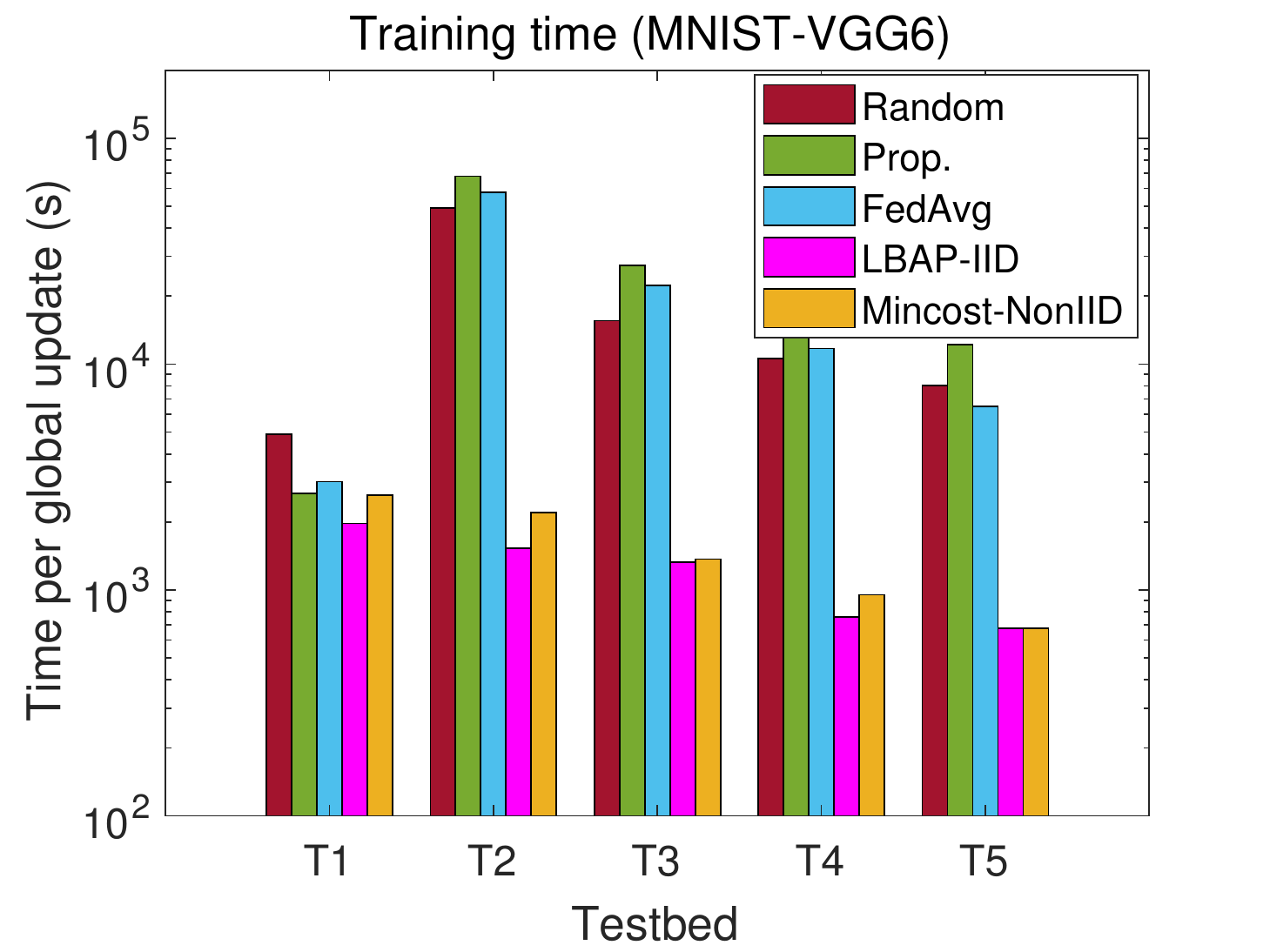}
                \vspace{-0.21in}
                \caption{}
\end{subfigure}
\begin{subfigure}[b]{0.24\textwidth}
                \includegraphics[width=1.03\textwidth]{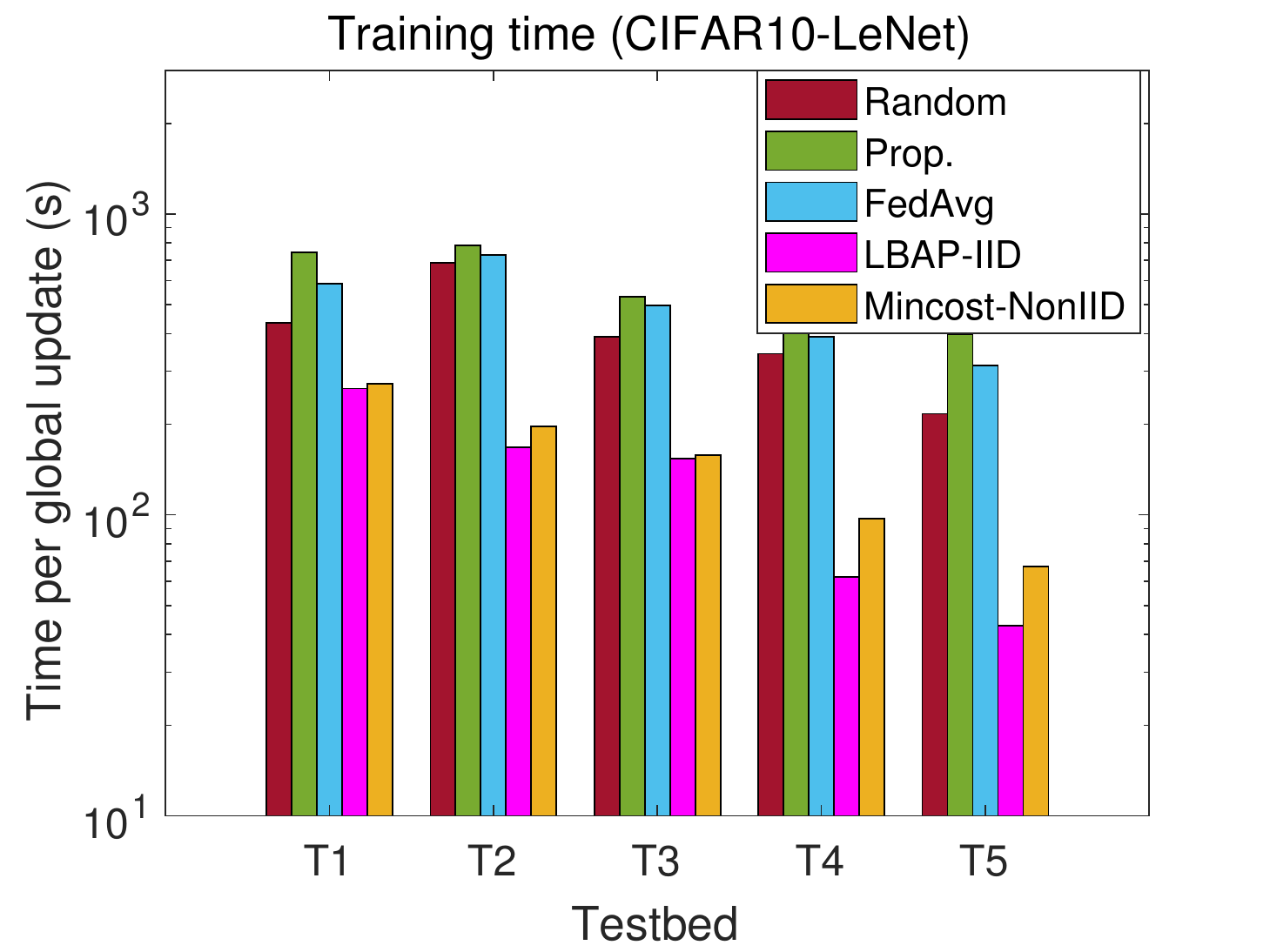}
                \vspace{-0.21in}
                \caption{}
\end{subfigure}
\begin{subfigure}[b]{0.24\textwidth}
                \includegraphics[width=1.03\textwidth]{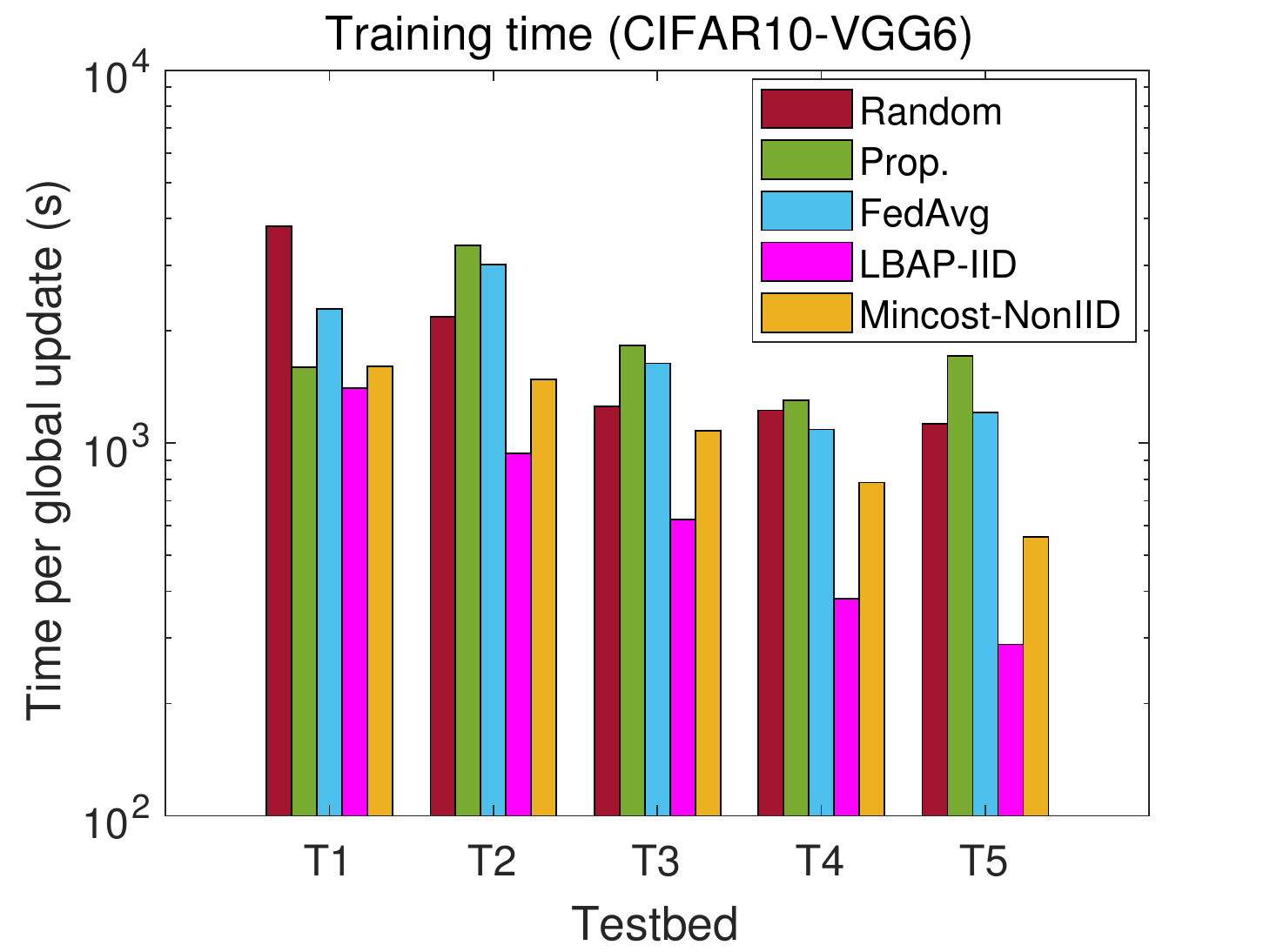}
                \vspace{-0.21in}
                \caption{}
\end{subfigure}%
\hspace*{-0.01in}
\vspace*{-0.13in}
\caption{Comparison of computation time when data is IID (time in log-scale) (a) training MNIST with LeNet; (b) training MNIST with VGG6; (c) training CIFAR10 with LeNet; (d) training CIFAR10 with VGG6. }
\label{iid_time_fig}
\vspace*{-0.1in}
\end{figure*}

\begin{figure}[!ht]
\centering
\vspace{-0.08in}
\hspace*{-0.35cm}
\includegraphics[width=3.8in]{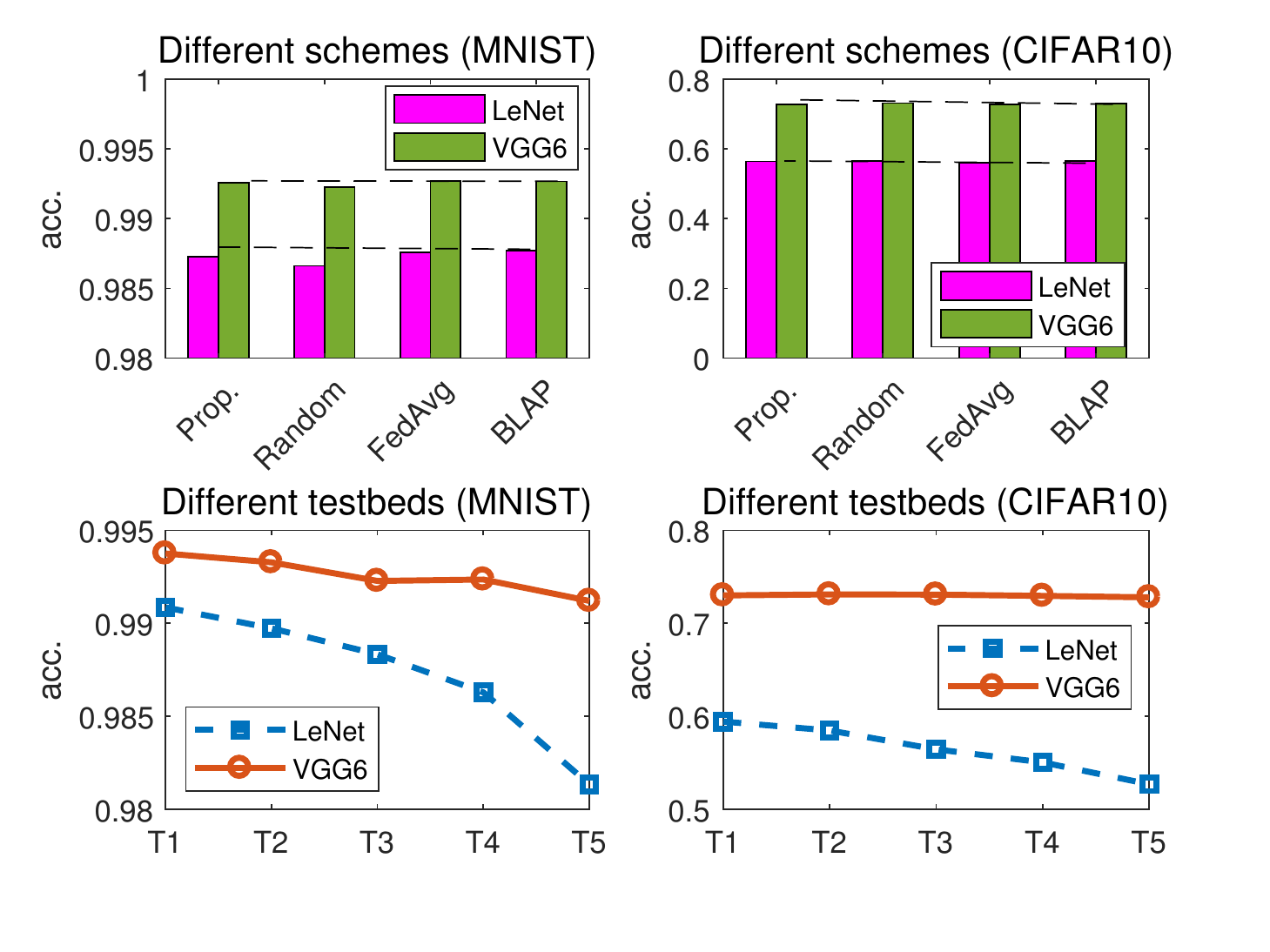}
\vspace{-0.35in}
\caption{Comparison of accuracy across different schemes and testbeds (IID).}
\label{iid_accuracy_fig}
\end{figure}

\begin{figure*}[!ht]
\vspace*{-0.09in}
\centering
\hspace*{-0.2in}
\begin{subfigure}[b]{0.24\textwidth}
                \includegraphics[width=1.02\textwidth]{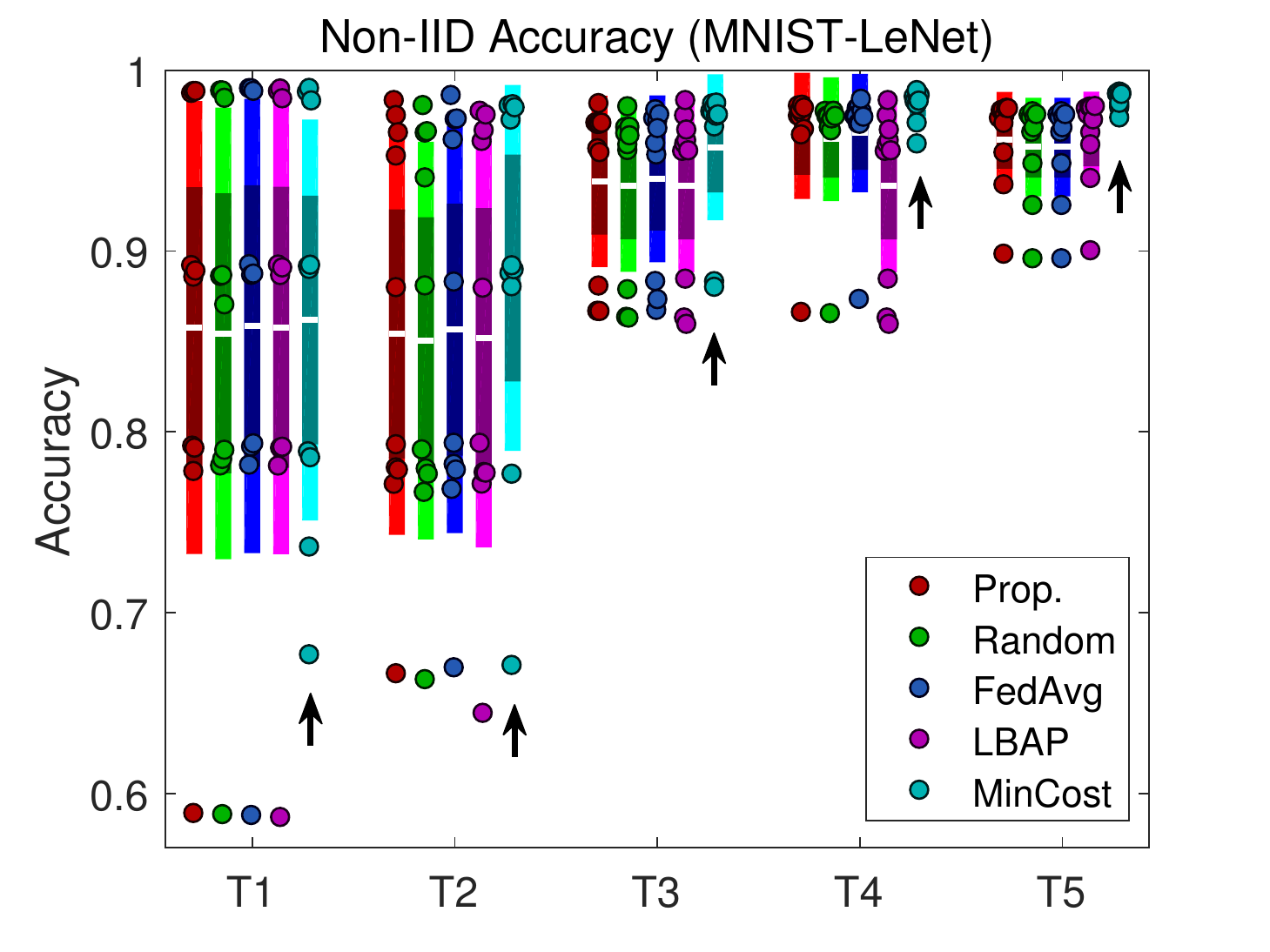}
                \vspace{-0.21in}
                \caption{}
\end{subfigure}
\begin{subfigure}[b]{0.24\textwidth}
                \includegraphics[width=1.02\textwidth]{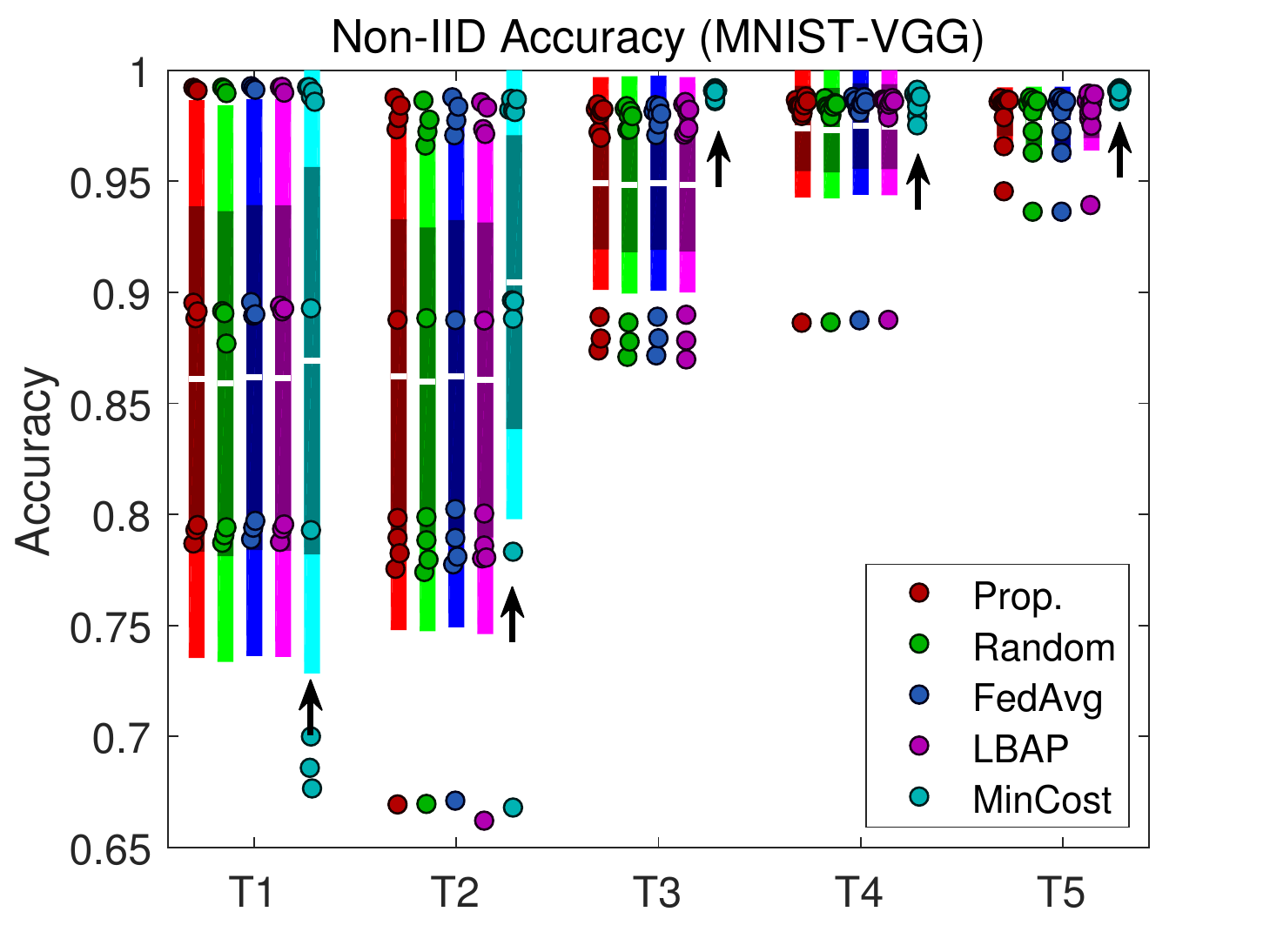}
                \vspace{-0.21in}
                \caption{}
\end{subfigure}
\begin{subfigure}[b]{0.24\textwidth}
                \includegraphics[width=1.02\textwidth]{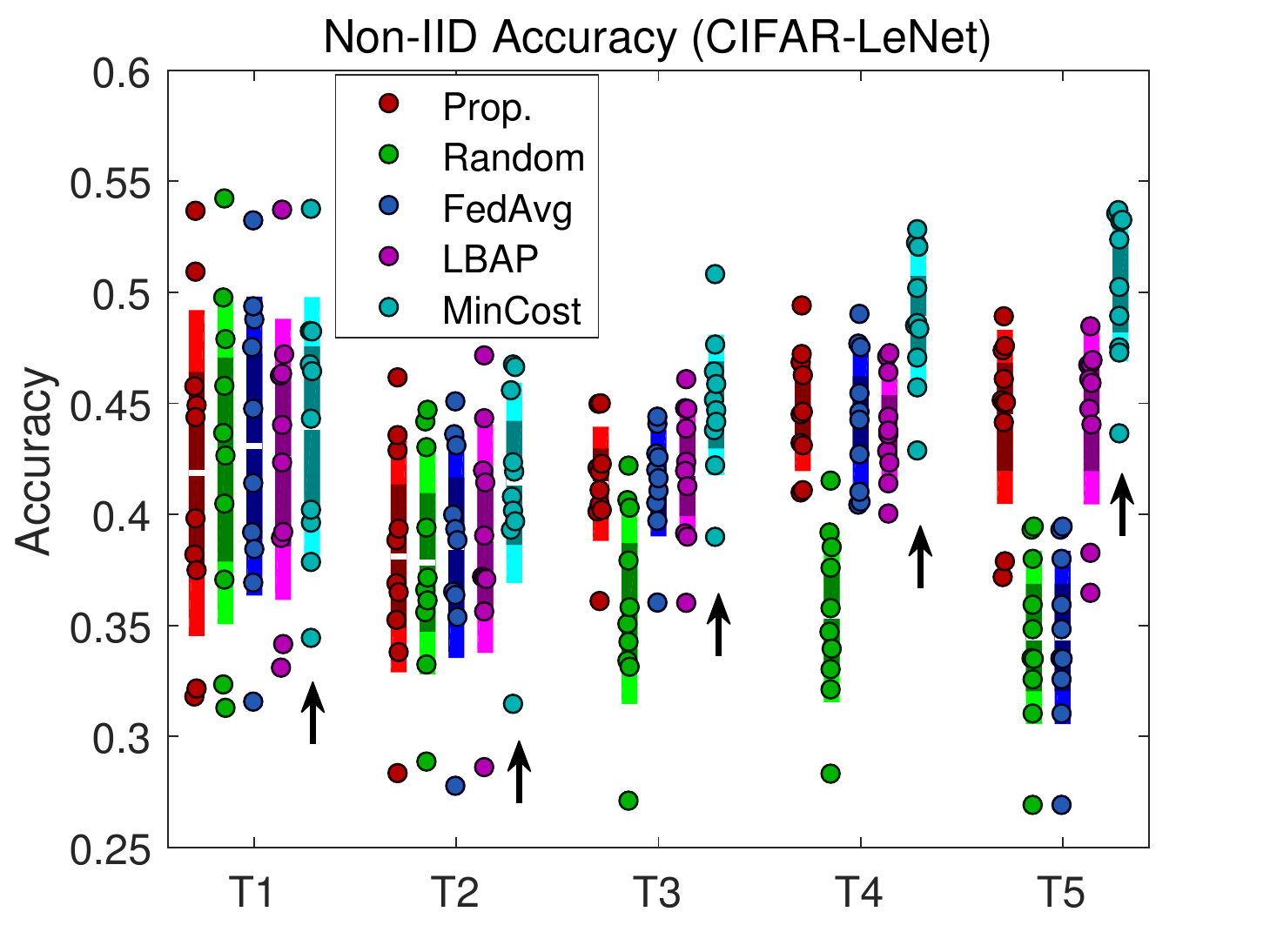}
                \vspace{-0.21in}
                \caption{}
\end{subfigure}
\begin{subfigure}[b]{0.24\textwidth}
                \includegraphics[width=1.02\textwidth]{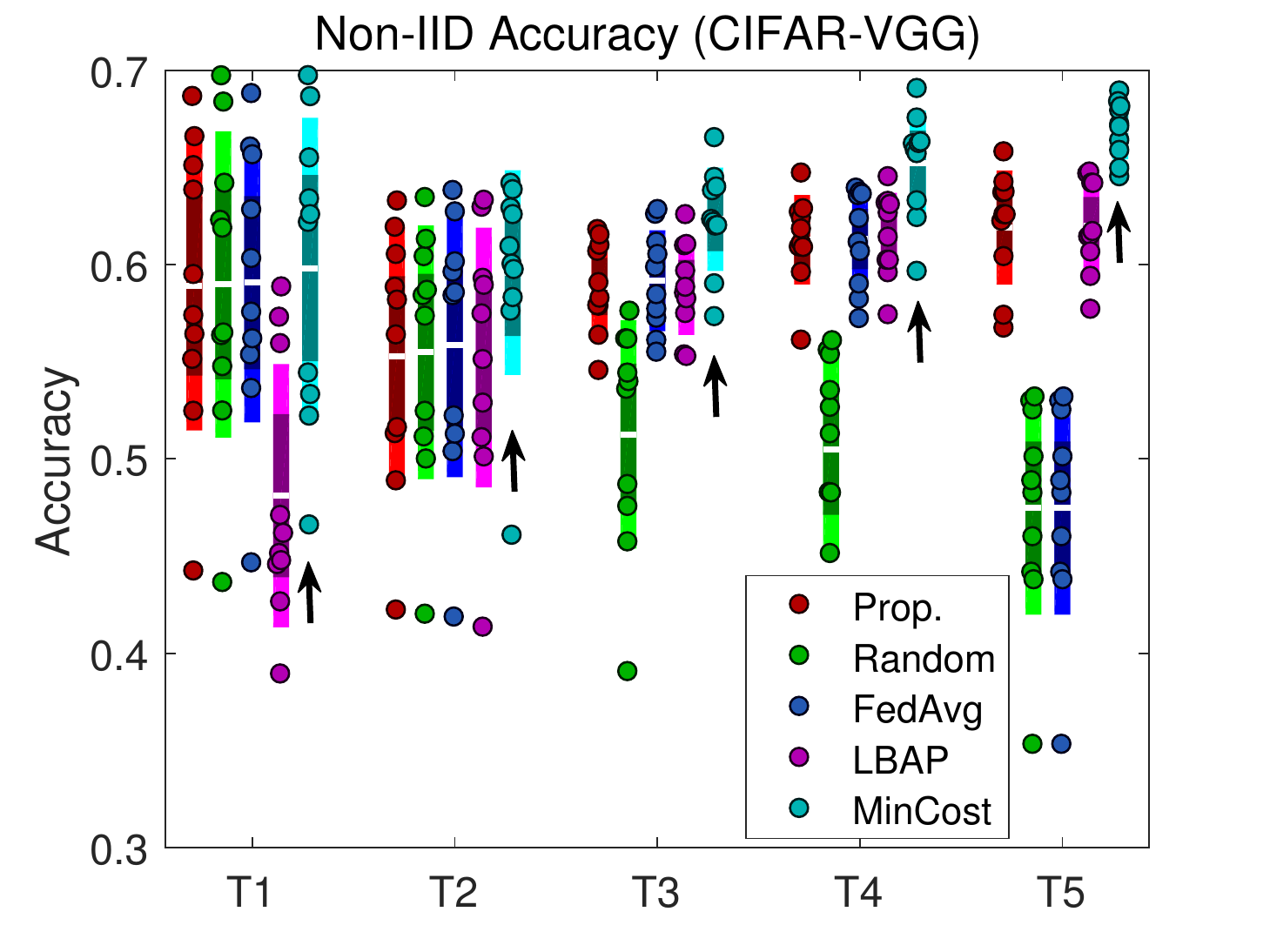}
                \vspace{-0.21in}
                \caption{}
\end{subfigure}%
\hspace*{-0.01in}
\vspace*{-0.13in}
\caption{Comparison of accuracy (Non-IID) (a) MNIST-LeNet; (b) MNIST-VGG6; (c) CIFAR10-LeNet; (d) CIFAR10-VGG6.}
\label{noniid_accuracy_fig}
\vspace*{-0.1in}
\end{figure*}

\begin{figure*}[!ht]
\vspace*{-0.09in}
\centering
\hspace*{-0.2in}
\begin{subfigure}[b]{0.24\textwidth}
                \includegraphics[width=1.02\textwidth]{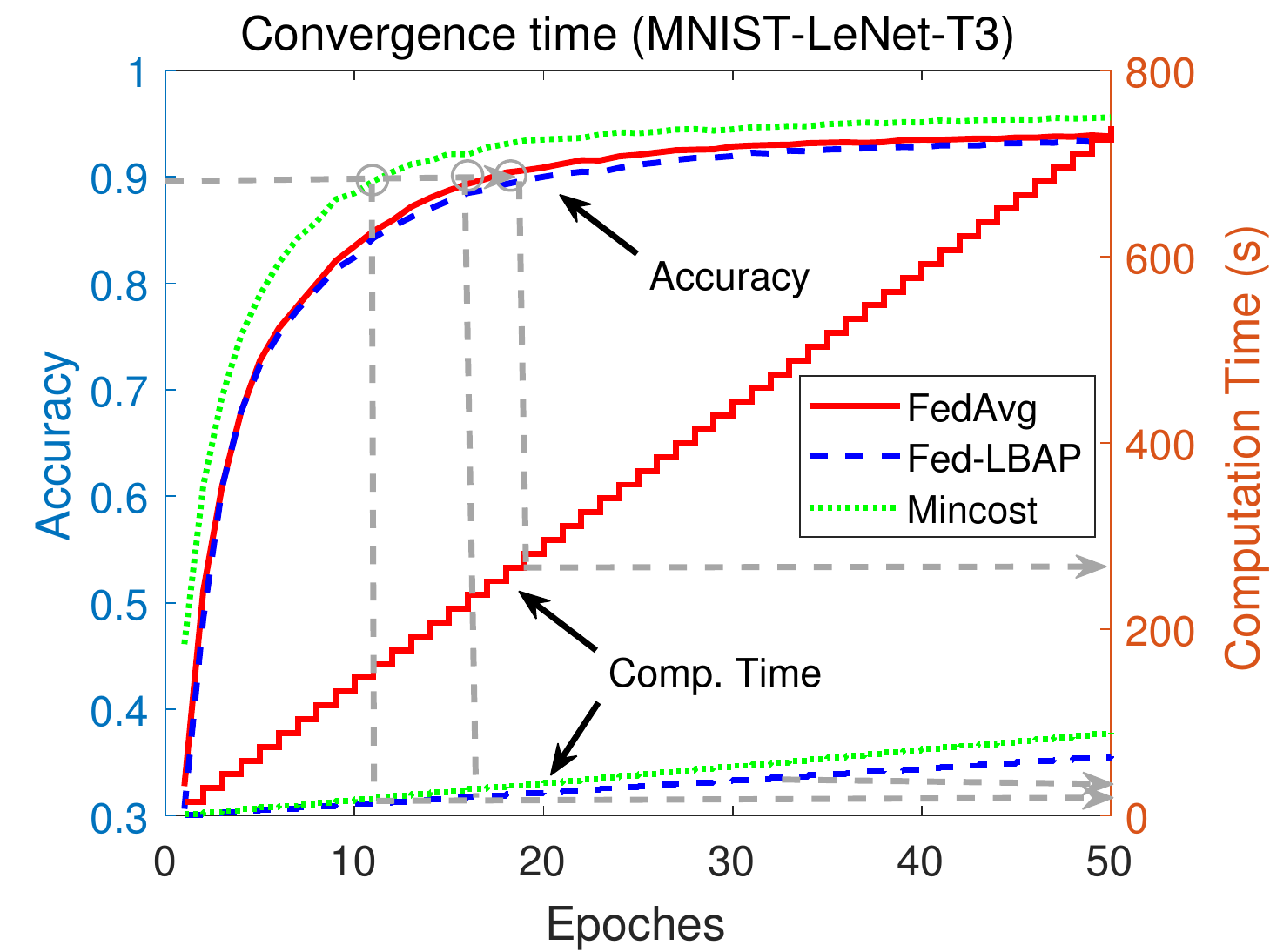}
                \vspace{-0.21in}
                \caption{}
\end{subfigure}
\begin{subfigure}[b]{0.24\textwidth}
                \includegraphics[width=1.02\textwidth]{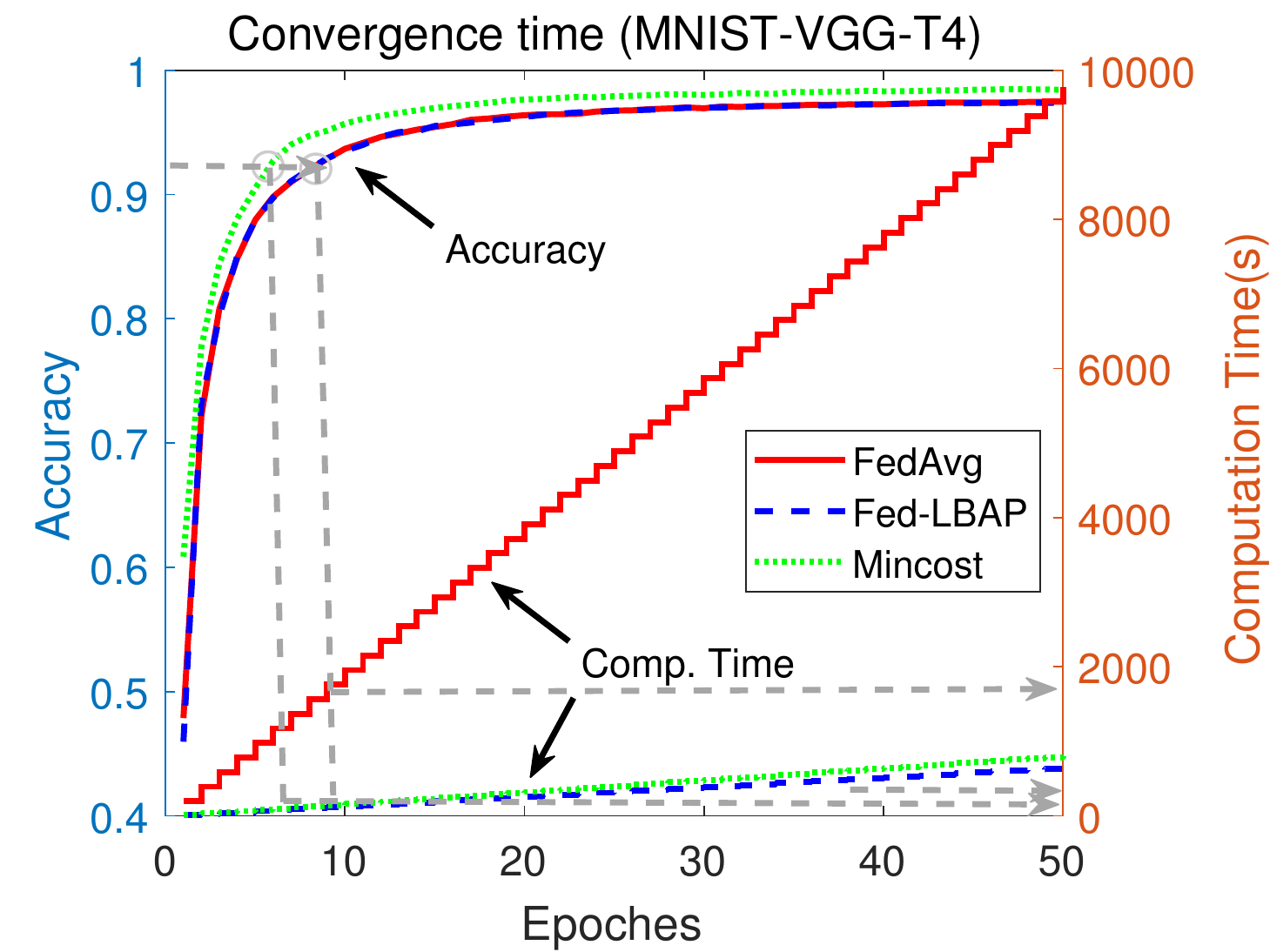}
                \vspace{-0.21in}
                \caption{}
\end{subfigure}
\begin{subfigure}[b]{0.24\textwidth}
                \includegraphics[width=1.02\textwidth]{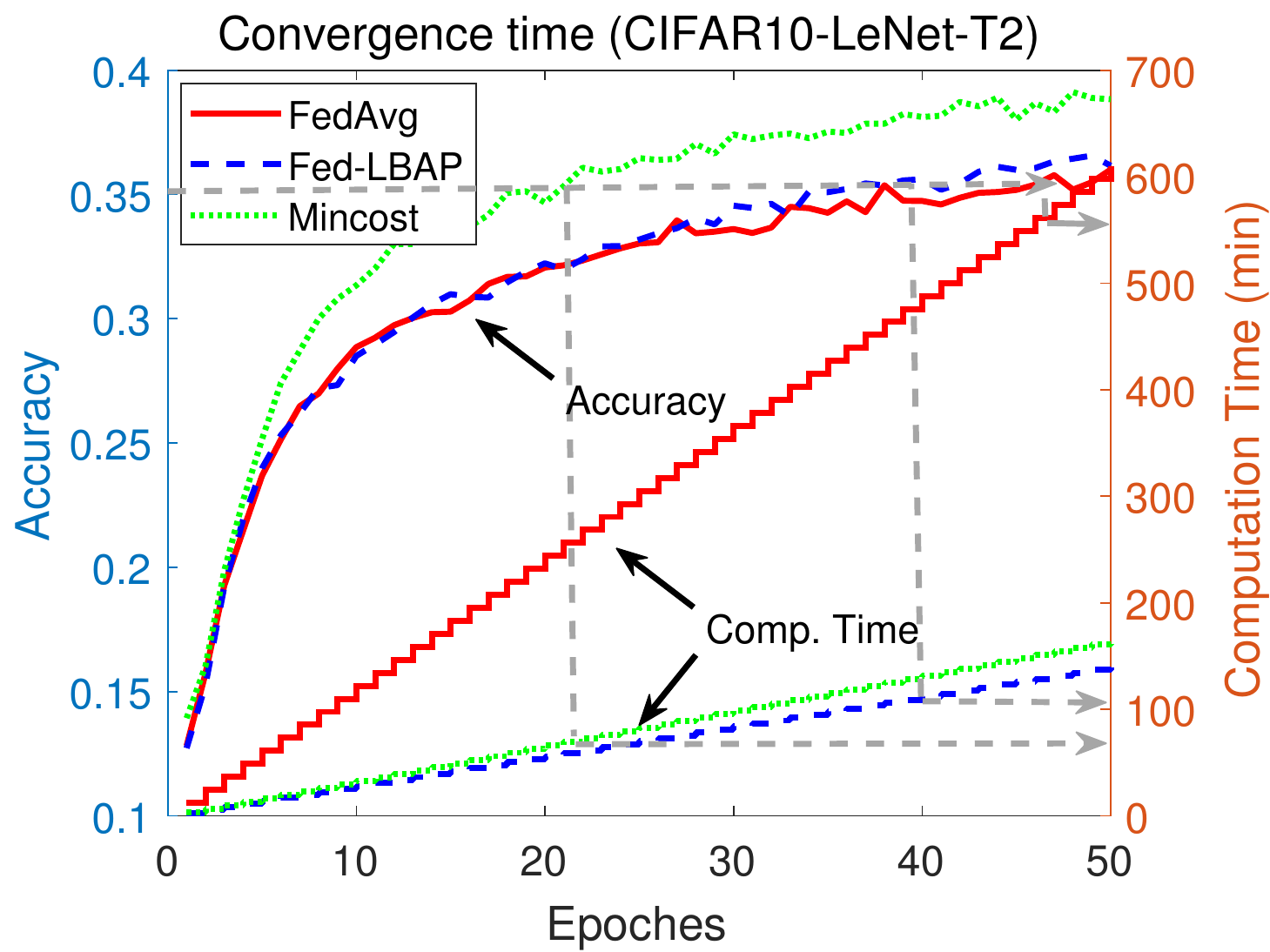}
                \vspace{-0.21in}
                \caption{}
\end{subfigure}
\begin{subfigure}[b]{0.24\textwidth}
                \includegraphics[width=1.02\textwidth]{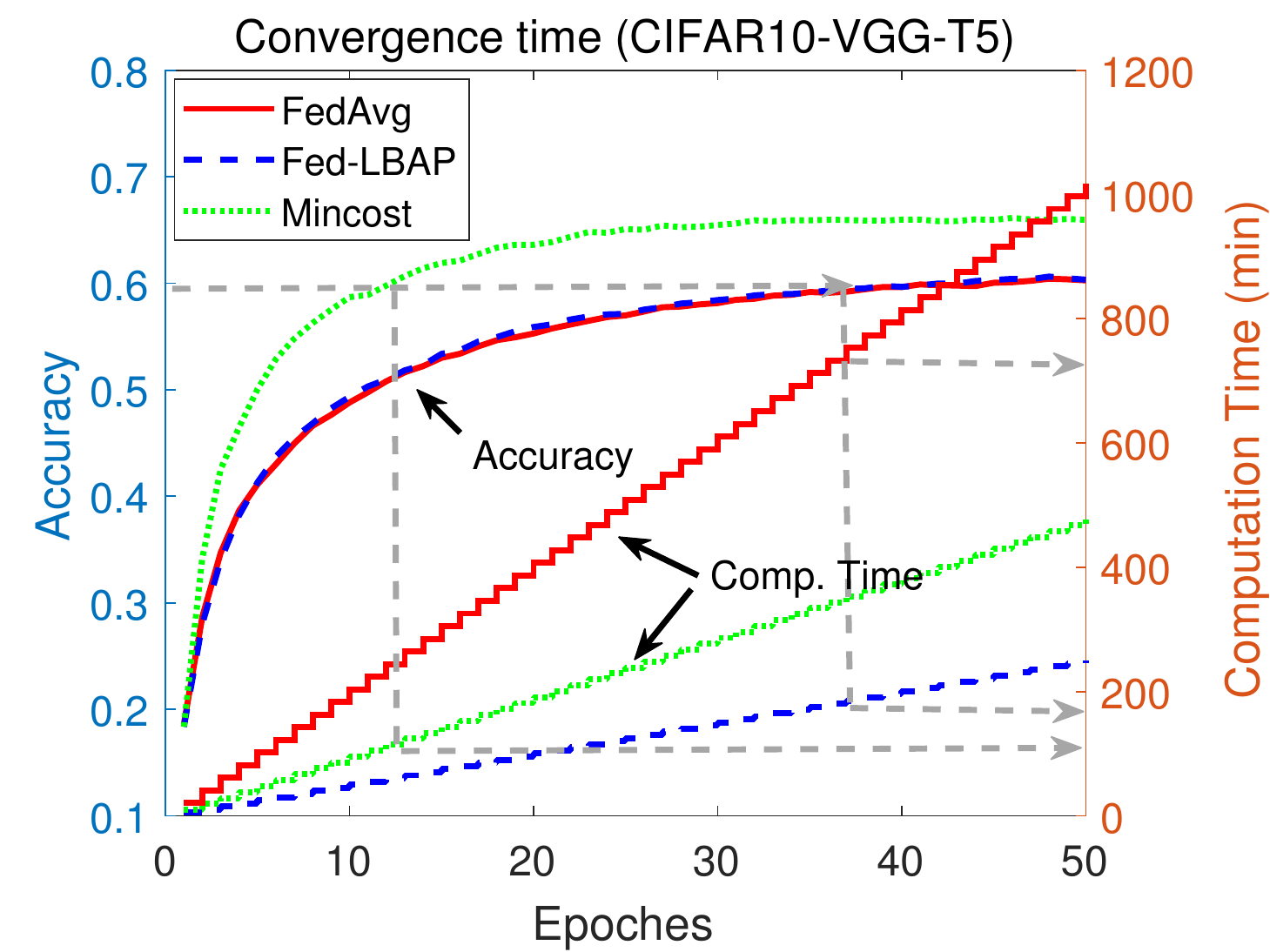}
                \vspace{-0.21in}
                \caption{}
\end{subfigure}%
\hspace*{-0.01in}
\vspace*{-0.13in}
\caption{Comparison of total convergence time (a) MNIST-LeNet-T3 (10 users) (b) MNIST-VGG6-T4 (14 users); (c) CIFAR10-LeNet-T2 (6 users); (d) CIFAR10-VGG6-T5 (20 users).}
\label{convergence_time_fig}
\vspace*{-0.15in}
\end{figure*}

\vspace{-0.03in}
\subsection{Profiling Performance}  \label{subsec:profiling}

As the basis to launch optimization, we evaluate the effectiveness of the profiling method in Sec. \ref{sec:profiling}. We learn the regressor on $25$ neural networks and test on $8$ networks with parameters ranging from 0.2M to 12M. Recall that in the first step, we model the relation between the parameters and training time. Some results are shown in Fig.\ref{profiling_fig_1}. We can see that the slope of the hyperplane of Nexus6 is steeper than Mate10, representing more computational time from the older smartphone generations. The upward trend is more pronounced in the convolutions and this validates the operation to separate convolution from the dense layers. Table \ref{table:profile_param} shows the learned parameters by the linear regression model. The values $\alpha_1, \alpha_2$ directly associate with the computational capabilities in conducting matrix multiplications for the convolution and dense layers. A larger value indicates less computational power and the ranking order is consistent with the rest experiments.

Fig.\ref{profiling_fig_2}(a) shows the second step of the estimation vs. true measurement on some examples, while a small gap is observed. Though generally small on our testing data, the gap could become larger if there is not enough training samples collected around the targeted architecture. Note that the estimation has a little bias towards a longer execution time after processing about 30 data batches on Nexus 6, and the estimation curve is not perfectly linear. This is because a majority of the data used to train the linear regressors exhibits thermal throttling on Nexus 6. Thus, our method generates estimations with a little bias assuming the Nexus 6 would underperform. Fig.\ref{profiling_fig_2}(b), we plot the root mean square error (RMSE) against the number of parameters for the 8 testing cases to see in which situation the regression model could potentially yield higher error. It is observed that the devices with less computational power are the ones having higher prediction error (about an order of magnitude), because of thermal throttling in the later iterations when the device heats up. It drives up the time curve into a slightly superlinear region, causing imperfect fitting from the linear regressors.

\vspace{-0.03in}
\subsection{Computation Time}

We first evaluate the computation time in each epoch by comparing the Fed-LBAP (IID) and MinCost algorithms (Non-IID) with the benchmarks shown in Fig. \ref{iid_time_fig}. We enumerate all the combinations between the testbed, datasets and models. Fed-LBAP achieves the lowest computational time with $2 \times$ to $100\times$ speedup compared to the benchmarks (time-optimal). Mincost adjusts the workload assignment based on the Non-IID distributions to trade epoch-wise computation time for faster convergence, which requires 5-10\% time in most cases.

By taking a closer look of the testbed devices, we can see that the mobile processing power (both individual and collective) and their workloads play key roles in the computational time, which sum up to nontrivial relations. First, unlike cloud settings in which computation time scales well with the number of workers, mobile stragglers easily slows down the entire training even if more users are involved: the time surge from T1 (3 users) to T2 (6 users) is due to the addition of Nexus6P, impacted by the severe thermal throttling. This drag is magnified with complex network architectures of higher computation intensity (VGG6 with more convolutional layers) and more training data (60K of MNIST vs. 50K CIFAR10). The shares from the 10K data addition exacerbate the computation time parabolically by 20 times (T2 between Figs. \ref{iid_time_fig}(b) and (d) running VGG6 on MNIST and CIFAR10), if the scheduling is done inappropriately. Bringing more devices could ease up the bottleneck as the time declines from T3 to T5 with more participants.

Using the vanilla schemes (Prop., Random and FedAvg), we hardly see any consistent parallelism when more users are involved, where the stragglers defeat the original purpose of distributed learning. In contrast, Fed-LBAP and Mincost are capable of utilizing the additive computational resources by appropriately assigning workloads to the more efficient users, so the training time accomplishes a downtrend with more users, even when the worst-case stragglers are present. This is because the proposed algorithms can purposely assign data in proportion to the device's capacity, once the thermal effects have been quantified. Although naive schemes may look for an optimal scheduling that is proportional to device CPU frequency or equally assign workloads as~\cite{fedavg}, the runtime behavior may be drastically different due to complex system dynamics, and our experiment shows that such schemes are on par with a purely random schedule. Note that we only perform one local epoch in each iteration. More local epoches can accelerate global convergence~\cite{fedavg}. Our strategy is invariant to the upper-layer learning algorithms and the time saving would be indeed much higher if more than one local epoches are performed.

\vspace{-0.03in}
\subsection{Accuracy}

It is essential to evaluate the test accuracy after workload re-assignment, especially our efforts to improve accuracy in case of non-IID data. Fig. \ref{iid_accuracy_fig} summarizes the test accuracy under different scheduling schemes and testbeds when data is IID. The two upper plots compare the average accuracy of Fed-LBAP with the benchmarks. Our findings in the large-scale experiment are consistent with the previous motivation discovery, which drives the design of Fed-LBAP. For IID data, even random assignments do not have accuracy loss. Since Fed-LBAP can be considered as one special permutation from the random partitions, the results indicate that \emph{we can always leverage load unbalancing to optimize computation time without worrying about accuracy loss, if user data is IID}. The two lower plots show that accuracy trends down (with LeNet) when more users are involved (from 3 to 20 users). The observation is inline with~\cite{fedavg} and suggests an inherent trade-off between parallelism and global convergence, which is furthered discussed in the next subsection.

Fig. \ref{noniid_accuracy_fig} compares Mincost with the benchmarks when data is Non-IID. We vary the random seeds to generate different class distributions, i.e., each user has a random subset from all the classes. Each point in the figure corresponds to an accuracy measurement. Mincost surpasses all the benchmarks by $0.02$ in MNIST and $0.04$ in CIFAR (2-7\% increase in accuracy), including the Fed-LBAP algorithm which we directly apply on the Non-IID data. This justifies the introduction of accuracy cost in Mincost $-$ though using Fed-LBAP for non-IID data is time-optimal, its accuracy is on the same level of the benchmarks. It is interesting to see that the accuracy actually climbs up with more users in Non-IID data, where the opposite is perceived in Fig. \ref{iid_accuracy_fig} with IID data. This is because more users increase the class coverage of the population. Mincost can utilize these dynamics from more dispersed class distributions, and select the participants wisely to either avoid or retain those $n$-class outliers (where $n=1-2$ in our experiment). Hence, the accuracy improvement is significant with more users, especially on more complex dataset as CIFAR10.

\vspace{-0.03in}
\subsection{Convergence Time}

The Mincost algorithm trades the computational time per epoch over the long-term convergence time. We evaluate the effectiveness of such trade-off by comparing to the time-optimal Fed-LBAP and the vanilla FedAvg. The goal is to compare the total time in order to achieve 95\% accuracy among the weakest of FedAvg, Fed-LBAP and Mincost, which is either FedAvg or Fed-LBAP according to the previous accuracy evaluation. We select a case in each dataset-network combination and plot accuracy and computational time in two y-axis with the training epoches on the x-axis of Fig. \ref{convergence_time_fig}. The results are averaged over 10 different random class distributions among the users. Network communication time to upload/download model is added to the computation time, so it represents the entire duration of each global update. The convergence time is obtained by connecting the accuracy goal on the left y-axis to the accumulated computation time on the right y-axis.

Mincost offers much faster convergence - about half number of epoches to achieve the same level of accuracy. However, for the entire duration, \emph{only if the time savings from faster convergence surpasses the extra time spent in each epoch, Mincost could outperform Fed-LBAP in Non-IID data}. Before we see the results, let us derive this relation analytically. The convergence curve of the $i$-th neural network can be modeled by an exponential function $p_i(x) = A_i - e^{-\beta_i x}$. $x$ is the number of epoches. $p_i(x)$ is the prediction accuracy after $x$ epoches. $A_i$ is the best accuracy the network could achieve empirically. $\beta_i$ reflects the convergence rate. Once the computational time is known per epoch, the total time can be represented by a linear function $k_i \cdot x$ corresponding to the y-axis on the R.H.S. in Fig. \ref{convergence_time_fig}. The condition for the compuational time of Mincost ($k_1 x_1$) to be less than Fed-LBAP ($k_2 x_2$) is,
\begin{eqnarray}
\vspace{-0.03in}
\small
k_1 x_1 &<& k_2 x_2 \nonumber \\
\frac{1}{\beta_1} \log(A_1-p) k_1 &<& \frac{1}{\beta_2} \log(A_2-p) k_2 \nonumber \\
p &<& \frac{e^{\beta_1 k_2}A_2 - e^{\beta_2 k_1}A_1}{e^{\beta_1 k_2} - e^{\beta_2 k_1}}   \label{relation_eq}
\vspace{-0.03in}
\end{eqnarray}
The relation indicates that Mincost outperforms Fed-LBAP when the goal of accuracy is upper-bounded by Eq. \eqref{relation_eq}. It echoes with our observations in Fig. \ref{convergence_time_fig} that training progresses much faster with Mincost at the beginning. It provides $1.5-1.8\times$ and a tremendous $7-200\times$ speed-up compared to Fed-LBAP and FedAvg respectively. The experiment ultimately justifies our design of accuracy cost on carefully selecting the participants and assigning workloads based the class distributions. Compared to our previous work~\cite{ipdps}, which strives to preserve accuracy without loss, we not only improve the testing accuracy, but also the overall computational time.

\subsection{Selection of Parameter $\alpha$}  \label{alpha:sec}

\begin{figure}[!t]
\vspace*{-0.09in}
\centering
\hspace*{-0.22in}
\begin{subfigure}[b]{0.25\textwidth}
                \includegraphics[width=1.1\textwidth]{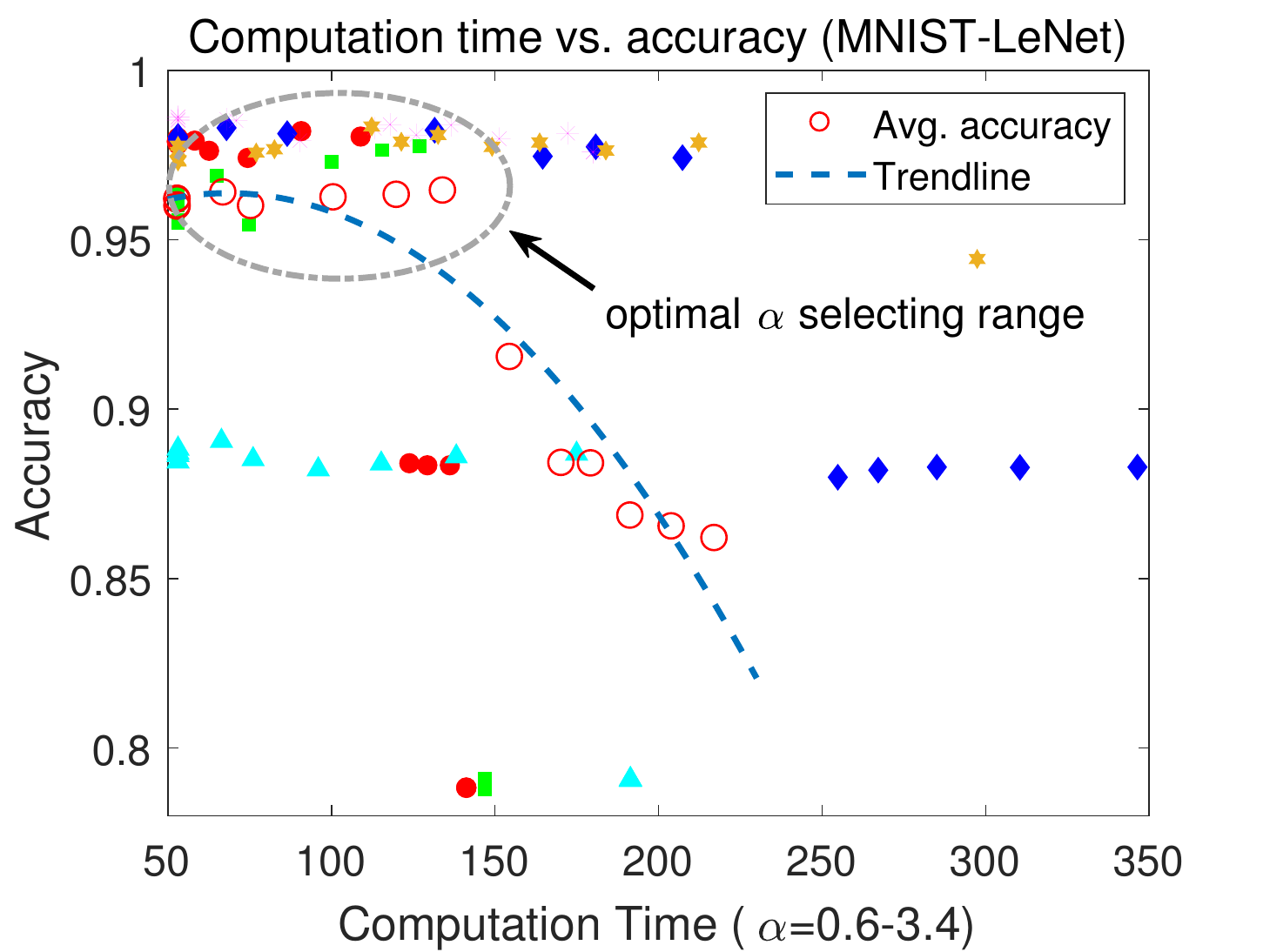}
                \vspace{-0.2in}
                \caption{}
\end{subfigure}
\hspace*{-0.02in}
\begin{subfigure}[b]{0.25\textwidth}
                \includegraphics[width=1.1\textwidth]{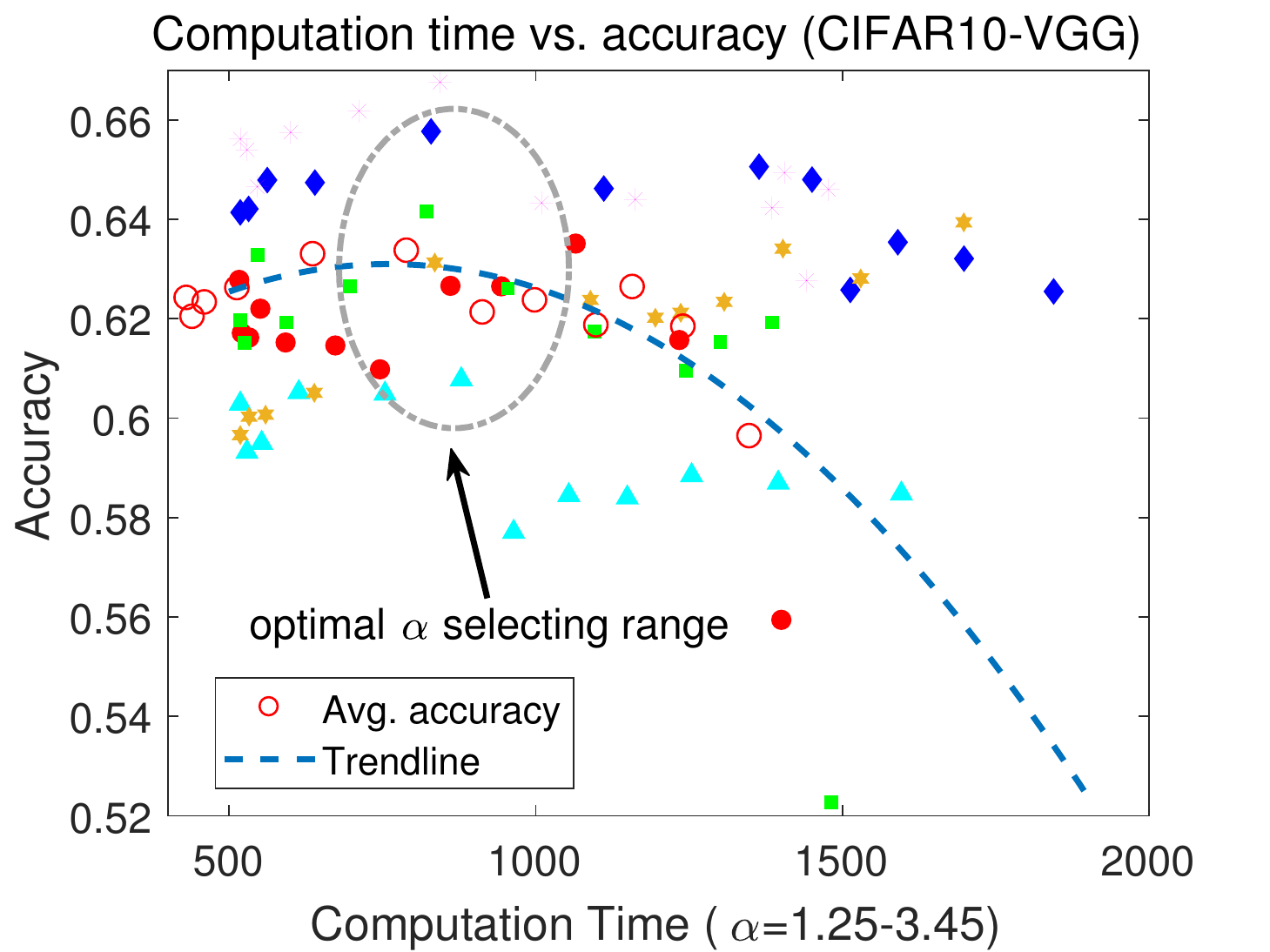}
                \vspace{-0.2in}
                \caption{}
\end{subfigure}
\hspace*{0.05in}
\vspace{-0.15in}
\caption{Relations between computation time and accuracy with different parameter $\alpha$. a) MNIST-LeNet b) CIFAR10-VGG6. (Colored dots represent different non-IID distributions generated from different random seeds). }
\label{alpha_relation_fig}
\end{figure}

There is still an important hyperparameter left to explain in Mincost - the scaling parameter $\alpha$ that balances the accuracy cost and computation time. We seek answers to the fundamental question: whether there exists an optimal $\alpha$ that can minimize the total computation time and how its value drives the process of decision making? Fig. \ref{alpha_relation_fig} partially answers the optimality question by enumerating its value from $0.6-3.4$ for MNIST-LeNet and $1.25-3.45$ for CIFAR10-VGG6. For clarity, we present 6 random non-IID distributions (represented by the dots in different colors) and their average accuracy vs. computation time. The computation time is monotonically increasing as $\alpha$ grows, because Mincost deviates from Fed-LBAP to favor those users with less accuracy cost (more num. of classes). They are most likely less time-optimal. An extreme case is when Nexus 6P device has all the classes, whereas the rest users only have a few classes. A large $\alpha$ assigns more data to the straggler and makes the time cost higher.

Despite its monotonicity with computation time, the implication on accuracy is not straightforward as the relationship is highly nonconvex. As seen in Fig. \ref{alpha_relation_fig}, the average accuracy trends up initially and drops promptly. We utilize a quadratic least square fitting to characterize both trendlines. However, the individual accuracy-vs-computation may not always hold such parabolic shape. The general downtrend is still intact as we increase $\alpha$ (and in turn computation time). This is because as $\alpha$ goes up, Mincost merely assigns workloads to those users with more classes and leaves the rest vacant. Of course, it undermines parallelism of federated learning and elongates computation time. There does exist a ``sweet spot'' for choosing $\alpha$ as the highlighted region, in which a mild increase of computation time can be compensated by the faster convergence as demonstrated previously.

\subsection{Performance Gap due to Profiling Error}

\begin{figure}[!ht]
\vspace*{-0.09in}
\centering
\hspace*{-0.22in}
\begin{subfigure}[b]{0.25\textwidth}
                \includegraphics[width=1.1\textwidth]{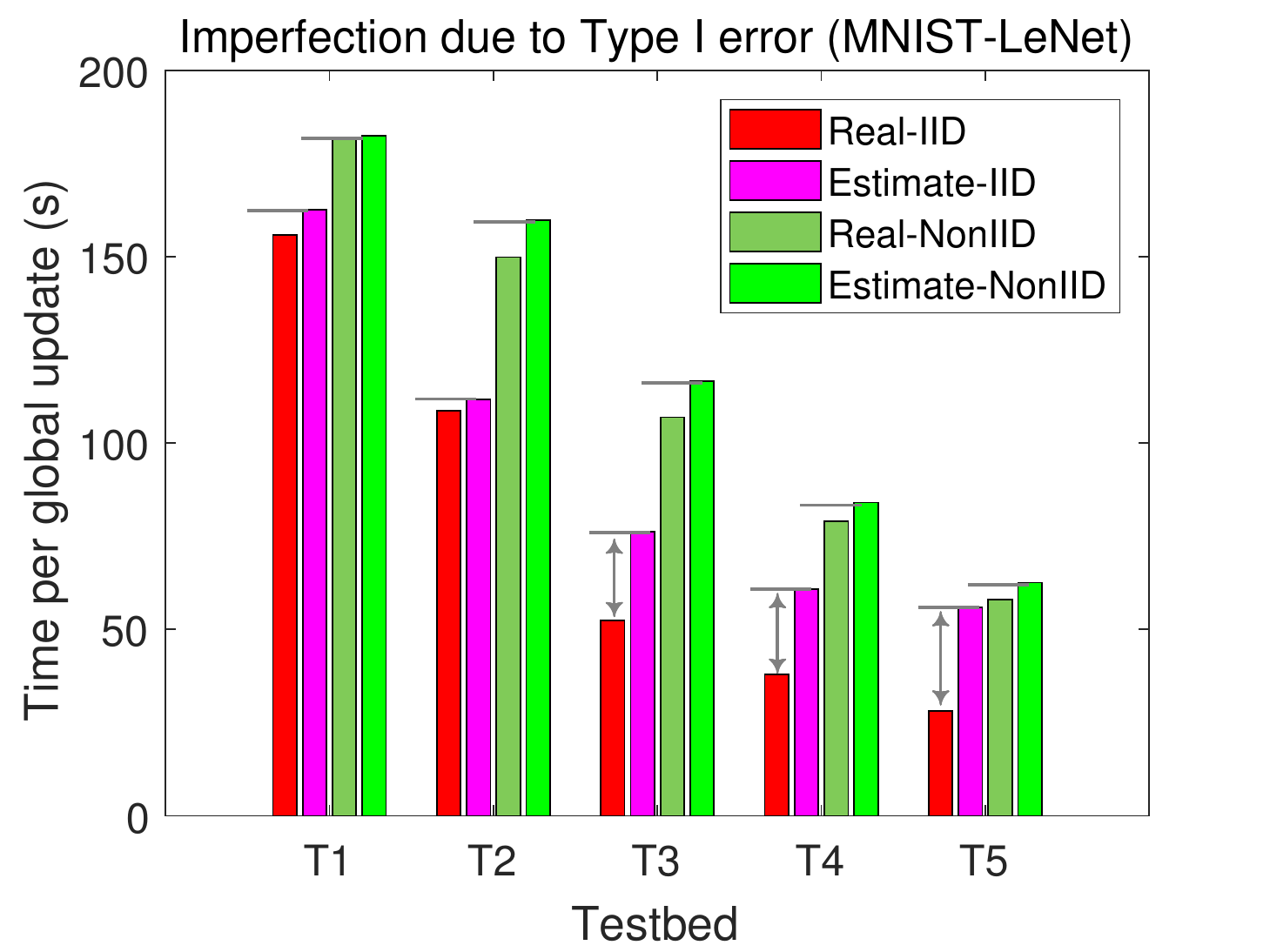}
                \vspace{-0.2in}
                \caption{}
\end{subfigure}
\hspace*{-0.02in}
\begin{subfigure}[b]{0.25\textwidth}
                \includegraphics[width=1.1\textwidth]{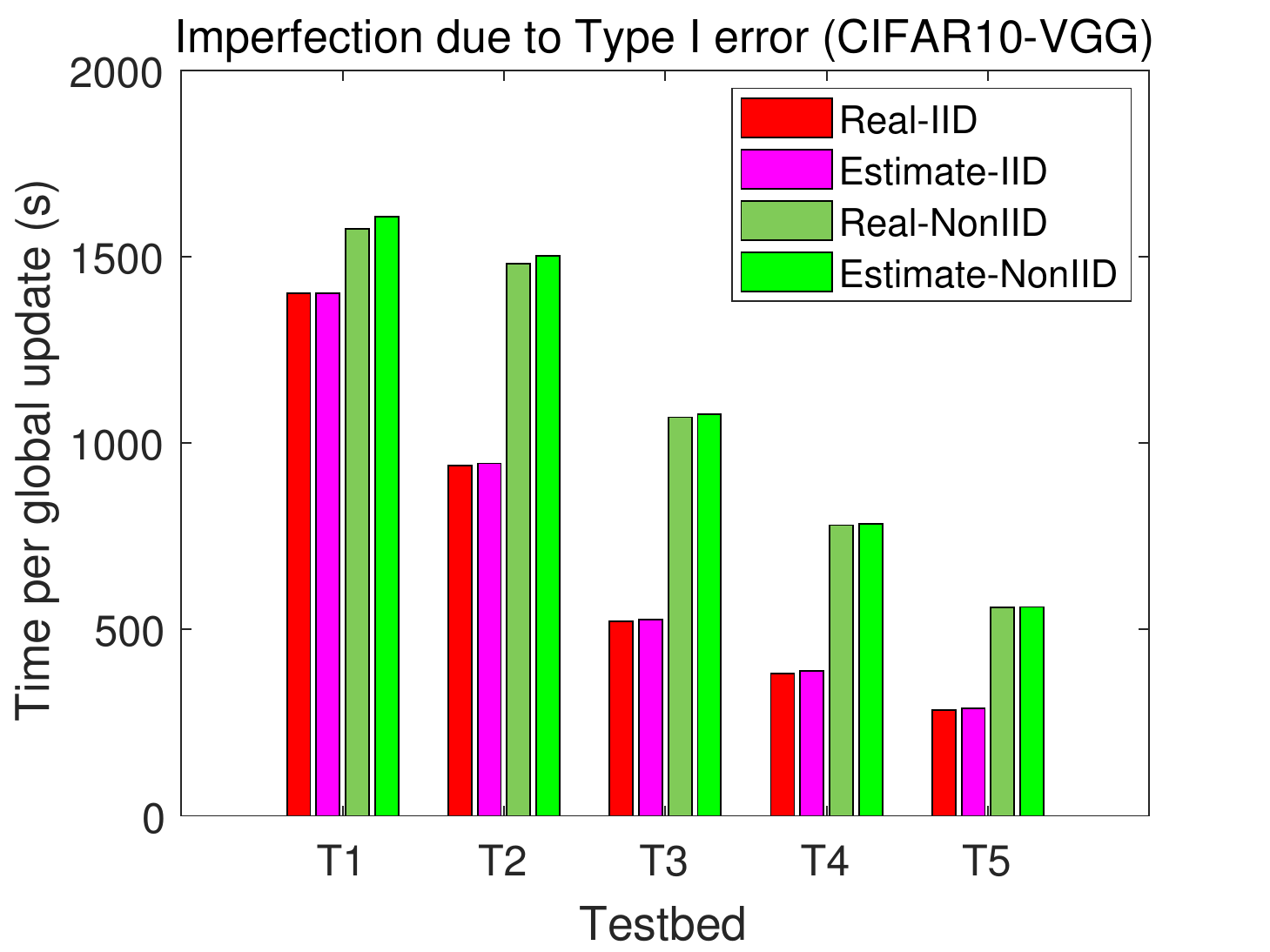}
                \vspace{-0.2in}
                \caption{}
\end{subfigure}
\hspace*{0.05in}
\vspace{-0.15in}
\caption{Performance gaps due to Type \Romannum{1} error a) MNIST-LeNet b) CIFAR10-VGG. }
\label{type_1_error_fig}
\end{figure}

Finally, we evaluate possible performance gaps due to two types of imperfections in profiling. Type \Romannum{1}: if the run-time trace is available from Step 1 in Fig. \ref{profiling_fig_1}, the second step with least square estimation incurs some minor gaps shown in Fig. \ref{profiling_fig_2}(a); Type \Romannum{2}: for untested architectures (run-time trace unavailable from Step 1), using the number of parameters for prediction brings more uncertainties. For Type I error, we measure the performance gaps between the real measurement, ``Real'', and the least square estimate, ``Estimate'' in Fig. \ref{type_1_error_fig}. The result shows the time difference induced by the inaccurate profile while running Fed-LBAP and Mincost. The estimation error is within 5\% for most cases and at most 25\% for some cases in MNIST. We scrutinize the data assignments and find that they are almost the same. Thus, there is no visible accuracy loss due to Type \Romannum{1} error.
%
%

\begin{figure}[!ht]
\vspace*{-0.09in}
\centering
\hspace*{-0.22in}
\begin{subfigure}[b]{0.25\textwidth}
                \includegraphics[width=1.1\textwidth]{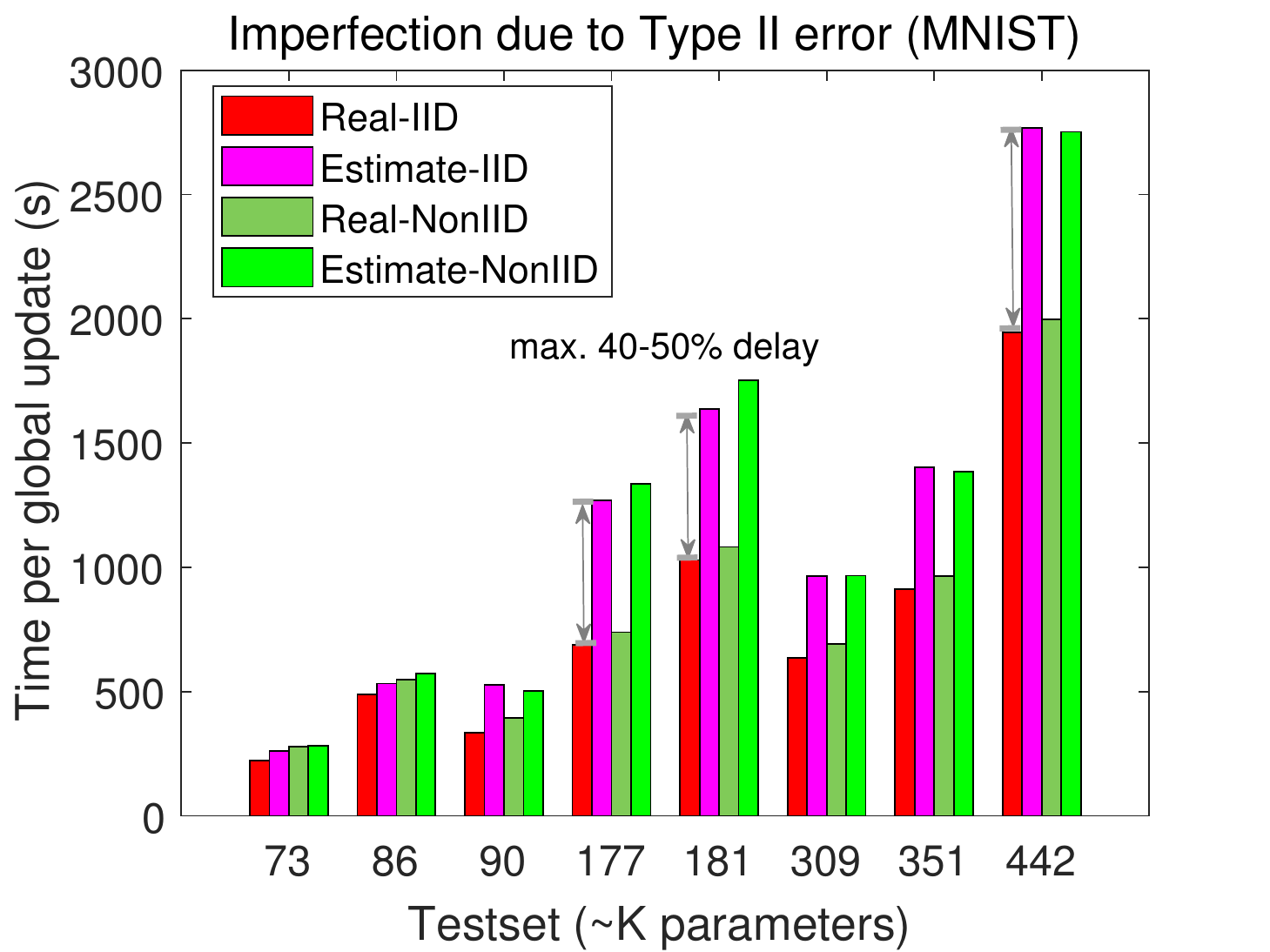}
                \vspace{-0.2in}
                \caption{}
\end{subfigure}
\hspace*{-0.02in}
\begin{subfigure}[b]{0.25\textwidth}
                \includegraphics[width=1.1\textwidth]{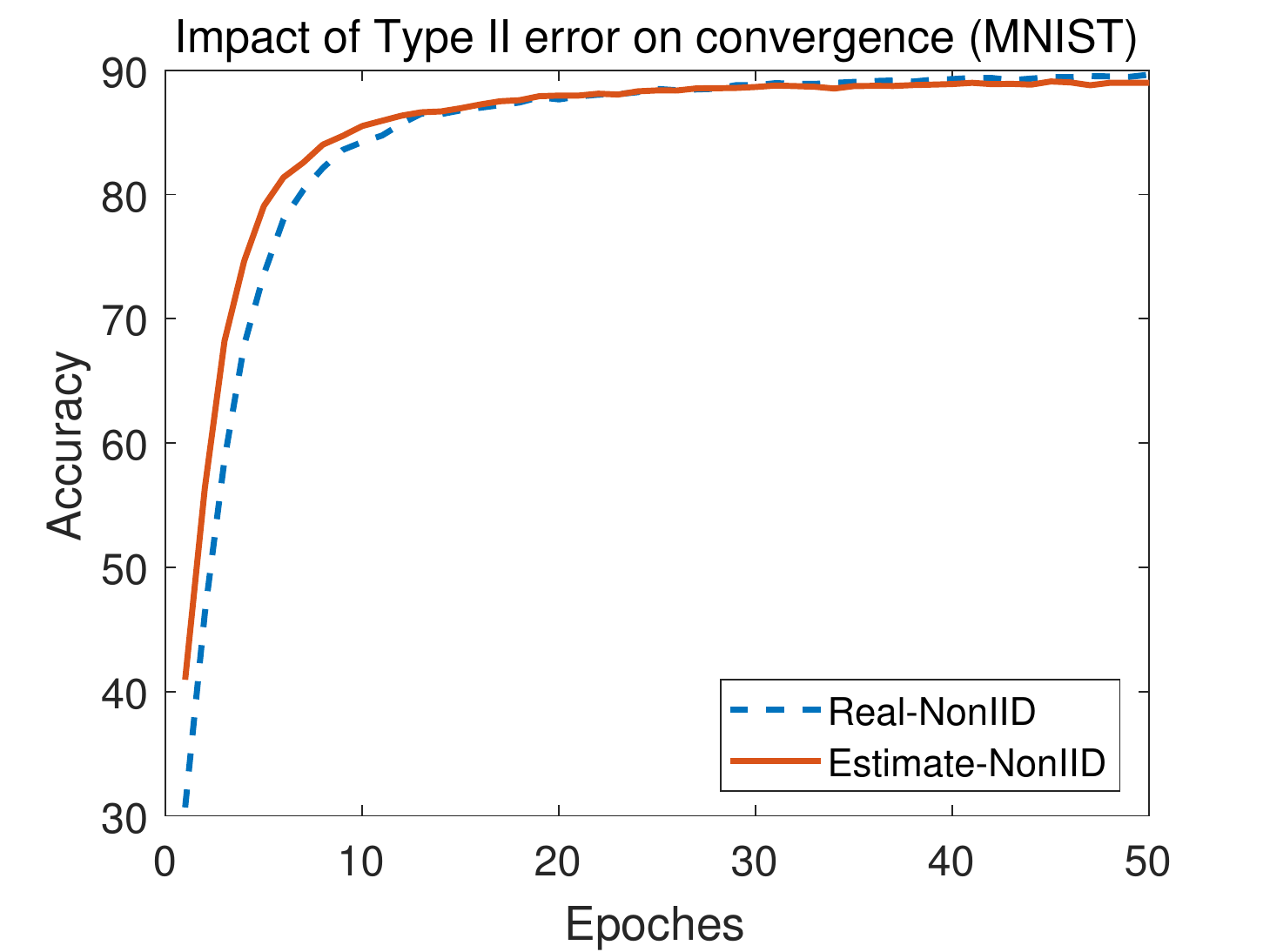}
                \vspace{-0.2in}
                \caption{}
\end{subfigure}
\hspace*{0.05in}
\vspace{-0.15in}
\caption{Performance gaps due to Type II error a) MNIST-LeNet b) CIFAR10-VGG. }
\label{type_2_error_fig}
\end{figure}

There are often cases that a new network architecture is implemented and the profiling data has not been collected yet. Our best shot is to use the known profiles to predict the computation time on a target mobile device given the number of model parameters (from convolutional and dense layers separately) and the amount of training data. Fig. \ref{profiling_fig_2}(b) indicates that the prediction RMSE is nonzero and the estimation error is accumulated in the later epoches. We construct a new testbed with 6 devices (one from each smartphone model) and utilize the 8 testing architectures. As seen in Fig. \ref{type_2_error_fig}(a), the error is much larger than Type \Romannum{1}. A common error we found is caused by overestimating the running time of some devices, which causes the algorithm to miscalculate (reduce the data assignment but increase for others. For example, we found that sometimes the Nexus6P training time is mistakenly doubled by the estimation and the algorithm assigns more data to the rest devices, which could have been run on Nexus6P without severe throttling. A partial reason is due to the limitations of the \emph{linear} models when some devices actually exhibit \emph{nonlinear} behaviors under different computational intensity.

We conduct more experiments to see the impact on convergence in Fig. \ref{type_2_error_fig}(b). The curve is the averaged accuracy over all 8 datasets by enumerating 10 different non-IID class distributions. Fortunately, unlike the timing gap, both curves converge to almost the same accuracy. However, the estimation has about 1-2 epoches latency, which amounts to 10-20 mins using only 6 devices. With more devices, such slowdown can be amortized effectively. As a result, the major instability mainly comes from modeling an untested architecture, when any under/overestimations would lead to sub-optimal assignment.


\section{Discussion} \label{sec:discussion}

The proposed framework utilizes linear regression to profile the training time on different devices. For federated tasks, the models are usually known beforehand, with necessary architectural adaptations to add/remove filters and adjust layers commensurate to the input dimension and memory limit. In this paper, we estimate the training time of specific model architectures related to the size of training data. Although our evaluations are based on the networks in the VGG family, the method is applicable to the more complex Convolutional Neural Networks (CNN) and Recurrent Neural Networks (RNN) such as the ResNet~\cite{resnet}, Inception-ResNet~\cite{resinception} and Long Short Term Memory (LSTM) networks~\cite{lstm}.

Modern neural architectures are designed with complex layer dependencies and irregular dataflow, that lead to more time in memory access~\cite{memory-sysml}. For example, the ResNet utilizes bypass links to facilitate gradient flow and the Inception modules~\cite{resinception} merge multiple branches. In principle, computation in neural networks is dominated by the number of multiply-add, e.g., from the \emph{convolutional} and \emph{fully connected layers}. It can be efficiently estimated using the number of Floating Point Operations (FLOPs)/model parameters. For the same type of networks, the linear relation generally holds because the number of FLOPs is proportional to the amount of training time when the computing resources are fully utilized on mobile. The time spent during memory access also remains proportionally similar. We have conducted more experiments to execute training of ResNet and LSTM in our testbed as discussed below. For LSTM, we use the dataset from the Physionet Challenge~\cite{physionet}, which consists of 4000 patients ICU visits of 86 feature vectors and a binary mortality label. We take 4\% of the data to train: i) a one-layer LSTM network with 300 units; ii) a two-layer LSTM network with 300 units in each layer.

\begin{figure}[!ht]
\vspace*{-0.09in}
\centering
\hspace*{-0.22in}
\begin{subfigure}[b]{0.25\textwidth}
                \includegraphics[width=1.1\textwidth]{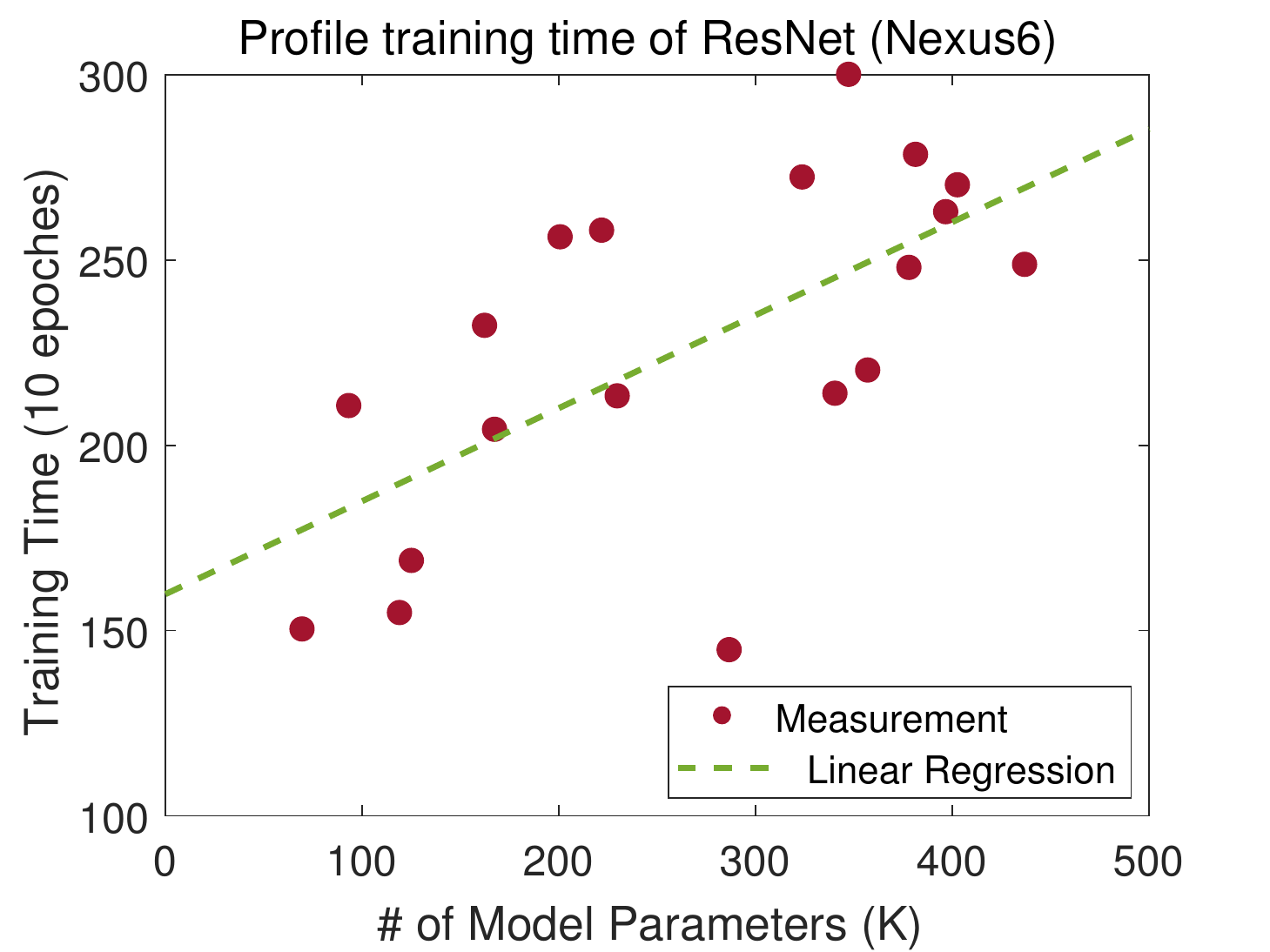}
                \vspace{-0.2in}
                \caption{}
\end{subfigure}
\hspace*{-0.02in}
\begin{subfigure}[b]{0.25\textwidth}
                \includegraphics[width=1.1\textwidth]{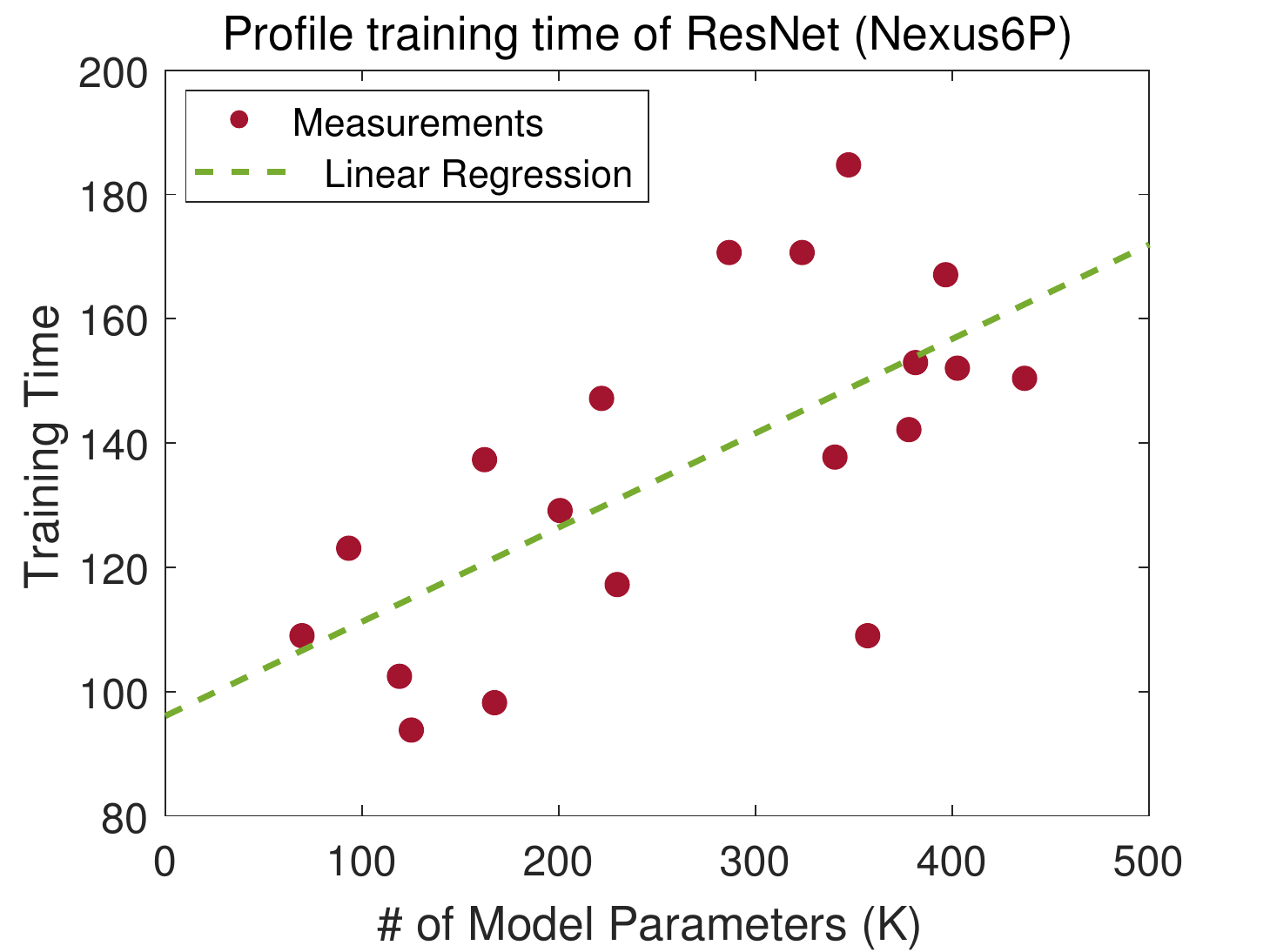}
                \vspace{-0.2in}
                \caption{}
\end{subfigure}
\hspace*{0.05in}
\vspace{-0.15in}
\caption{Profile ResNet architectures on the Nexus 6/6P.}
\label{profile_resnet}
\end{figure}

\begin{figure}[!ht]
\vspace*{-0.09in}
\centering
\hspace*{-0.22in}
\begin{subfigure}[b]{0.25\textwidth}
                \includegraphics[width=1.1\textwidth]{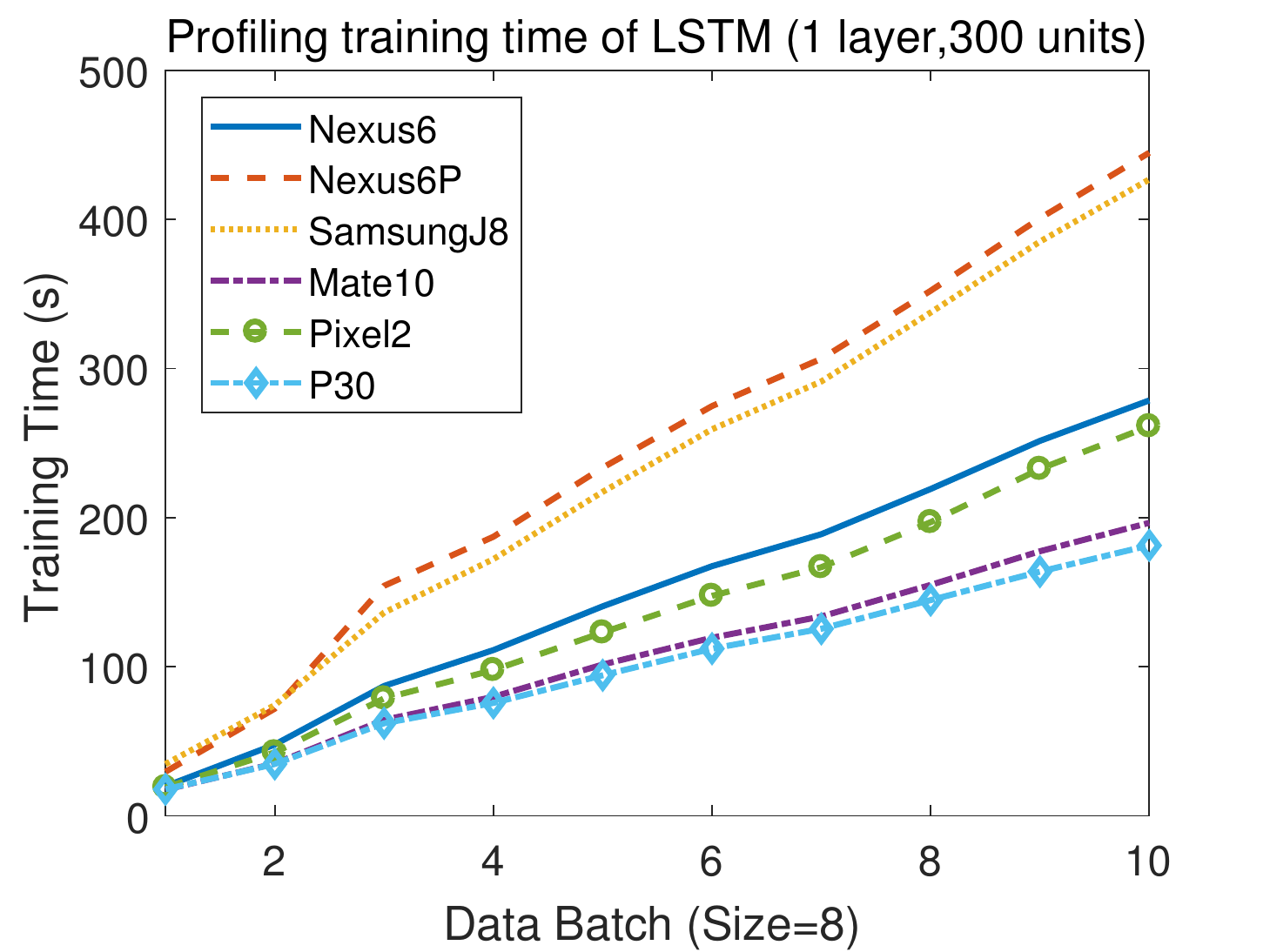}
                \vspace{-0.2in}
                \caption{}
\end{subfigure}
\hspace*{-0.02in}
\begin{subfigure}[b]{0.25\textwidth}
                \includegraphics[width=1.1\textwidth]{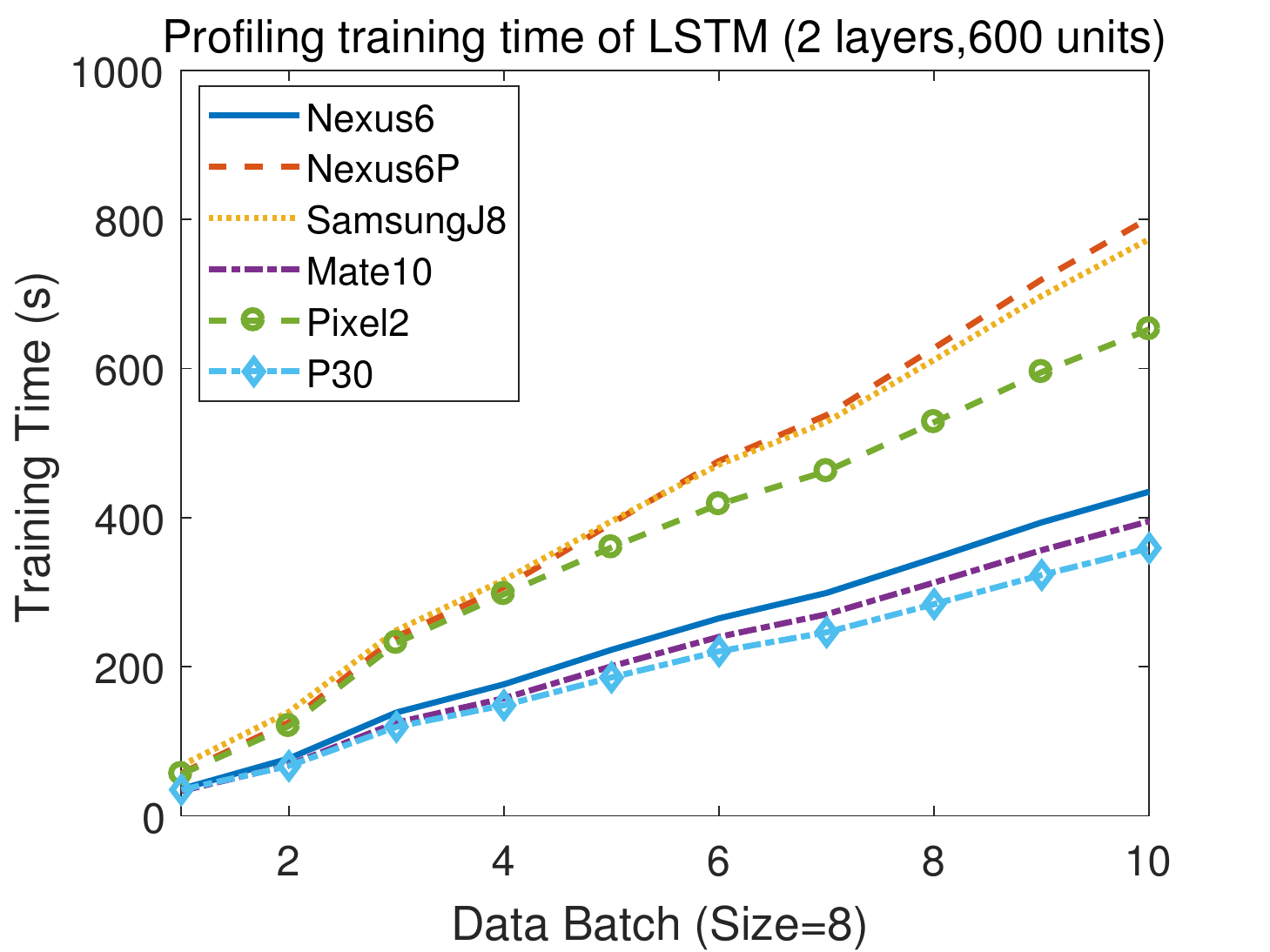}
                \vspace{-0.2in}
                \caption{}
\end{subfigure}
\hspace*{0.05in}
\vspace{-0.15in}
\caption{Profiling LSTM training time. (a) Layer-LSTM, 300 units (b) 2 Layer-LSTM, 600 units. }
\label{lstm1_fig}
\end{figure}

Fig. \ref{profile_resnet} plots the training time vs. the number of model parameters of ResNet architectures by adding/removing convolutional layers, adjusting kernel size and number of filters (on the two representative lowgrade devices). We can see that the linear relation generally holds. Note that adding/removing convolutional layers may have resulted similar number of parameters with adjusting the number of filters, but their training time may have noticeable difference, i.e., adding a convolutional layer typically leads to more training time due to increased cross-layer memory access. We have also tested Inception-ResNet and LSTM and their results reveal similar linearity relations. Next, we confirm that the linear relation also holds between the training time vs. the number of data batches in LSTM network as shown in Fig. \ref{lstm1_fig}. Some interesting phenomenons also worth discussion: 1) the execution time is much longer than the CNN with similar number of parameters. This is consistent with training LSTM on cloud GPU as it has very different dataflow/computation patterns from the CNN. 2) The memory demand is much higher per data batch. It triggers the background Garbage Collection activities quite often, which pause the training threads multiple times per epoch to free memory. Thus, in addition to the increased time, there are noticeable fluctuations on both curves.

\begin{figure}[!ht]
\vspace*{-0.09in}
\centering
\hspace*{-0.22in}
\begin{subfigure}[b]{0.25\textwidth}
                \includegraphics[width=1.1\textwidth]{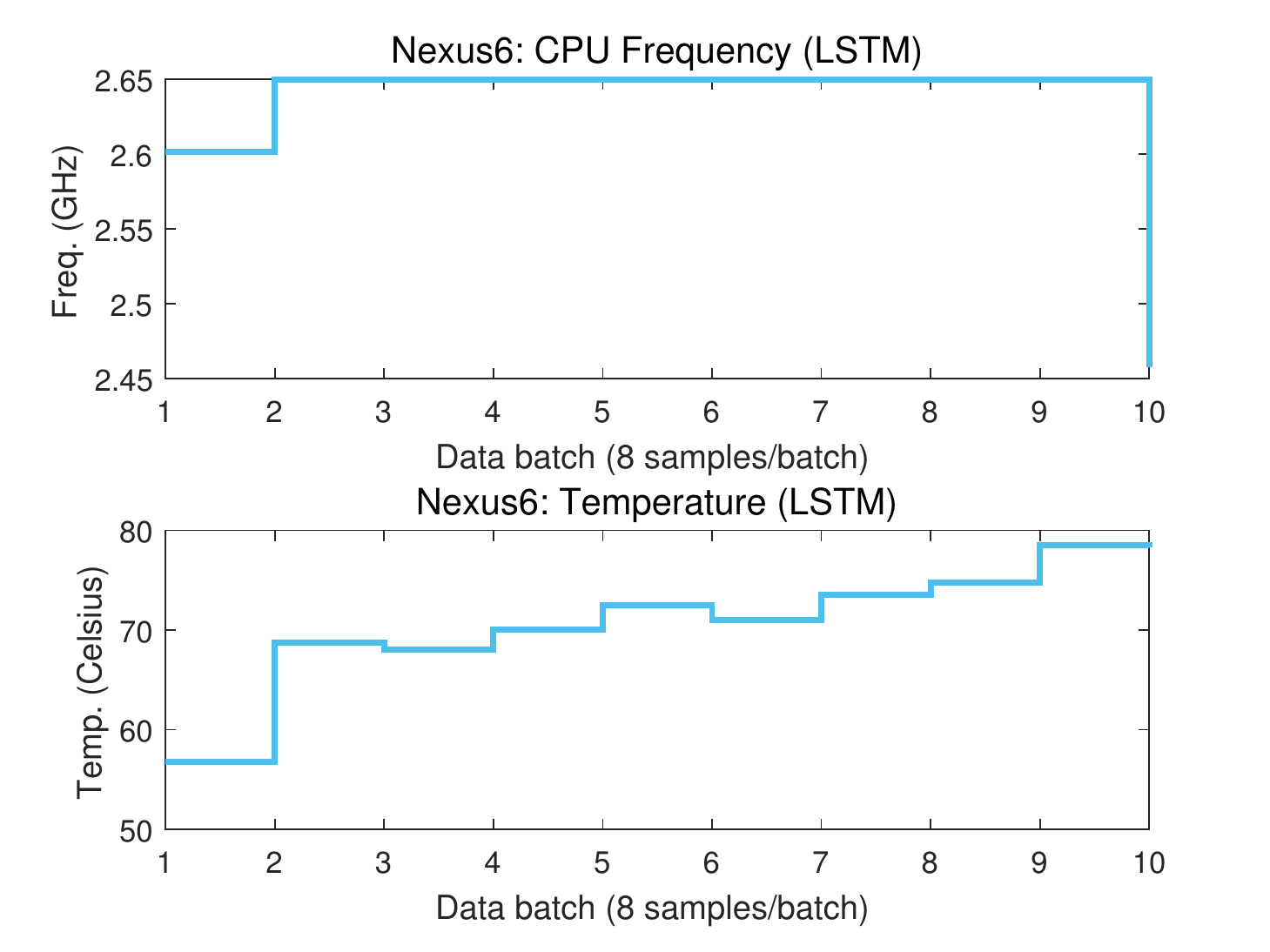}
                \vspace{-0.2in}
                \caption{}
\end{subfigure}
\hspace*{-0.02in}
\begin{subfigure}[b]{0.25\textwidth}
                \includegraphics[width=1.1\textwidth]{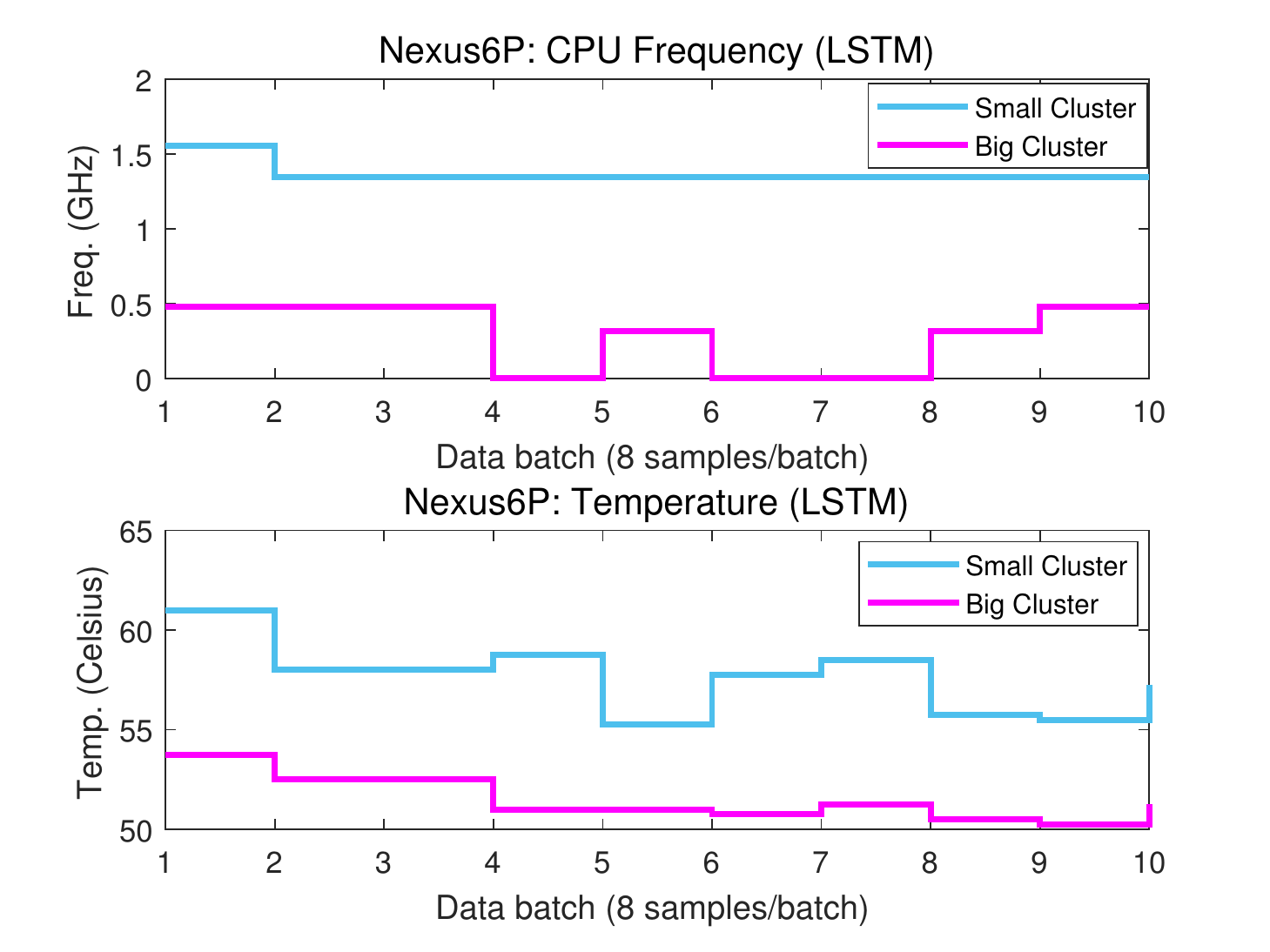}
                \vspace{-0.2in}
                \caption{}
\end{subfigure}
\hspace*{0.05in}
\vspace{-0.15in}
\caption{Profiling CPU frequency vs. temperature. (a) Nexus 6, (b) Nexus 6P. }
\label{lstm2_fig}
\end{figure}

On the positive side, the CPU can take a breath during memory free and the time spent in memory access. This is validated in Fig. \ref{lstm2_fig} that we show the trace of two devices that are prone to thermal throttling. For Nexus 6, if we compare the increase of temperature in Figs. \ref{cpu_temperature_fig}(a-b) with Fig. \ref{lstm2_fig}, the temperature curve of LSTM on Nexus6 obviously climbs much slower than training the CNN, thus creating more opportunities to run at high clockspeed; in contrast, the Nexus 6P still keeps shutting down the big cores to maintain their temperature below 55 degrees, which is more conservative compared to its predecessor.

For the non-IID properties discovered in this paper, they stem from the basic principles of deep learning which should also hold for both CNN and RNN networks. That is, high class imbalance, missing categorical information lead to accuracy loss and unseen samples help generalize the model. Due to space limit, we do not include a thorough evaluation, but the proposed mechanism is model-agnostic and applicable to a wide range of mobile devices on the market.

\section{Conclusion} \label{sec:conclusion}
In this paper, we study efficient scheduling of federated mobile devices under both device and data-level heterogeneity. We motivate this work by showing drastically different processing time under the same workload and opportunities to employ user selection to improve non-IID learning accuracy. We develop two efficient near-optimal algorithms to schedule workload assignments for both IID and non-IID data, and visualize the solution space analytically. Our extensive experiments on real mobile testbed and datasets demonstrate up to 2 orders of magnitude speedups and a moderate accuracy boost when data is non-IID. We also demonstrate cases when our algorithm deviates from the optimal region due to estimation gaps in the profiling process.

\section*{Acknowledgments}
This work was supported in part by the US NSF grant numbers CCF-1850045, IIS-2007386 and the State of Virginia Commonwealth Cyber Initiative (cyberinitiative.org).

\begin{IEEEbiography}
[{\includegraphics[width=1in,height=1.25in,clip,keepaspectratio]{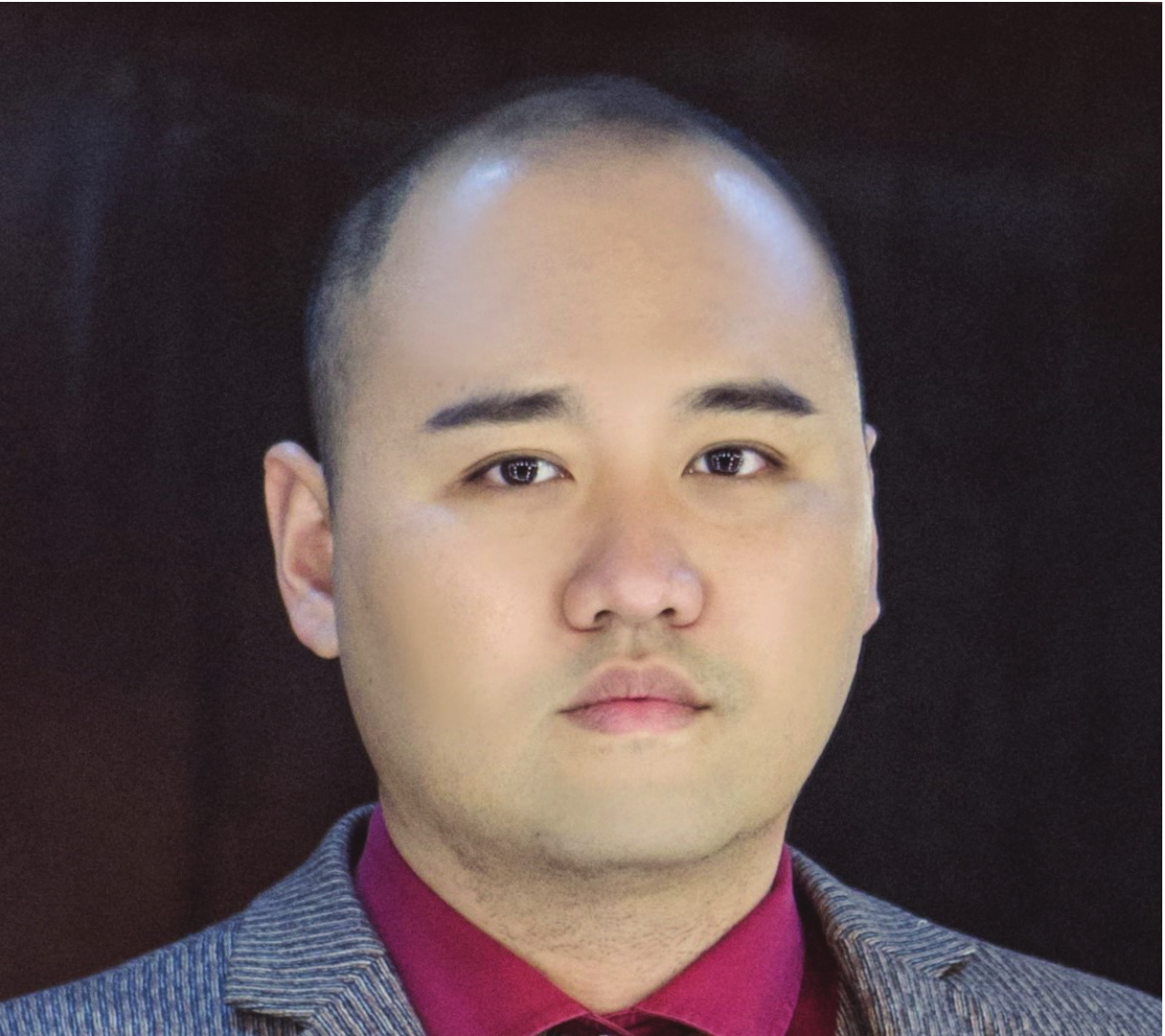}}]
{Cong Wang} received the B. Eng degree in Information Engineering from the Chinese University of Hong Kong in 2008, M.S. degree in Electrical Engineering from Columbia University in 2009, and Ph.D. in Computer and Electrical Engineering from Stony Brook University, NY, in 2016. He is currently an Assistant Professor at the Computer Science department, Old Dominion University, Norfolk, VA. His research focuses on exploring algorithmic solutions to address security and privacy challenges in Mobile, Cloud Computing, IoT, Machine Learning and System. He is the recipient of the Commonwealth Cyber Initiative Research and Innovation Award, ODU Richard Cheng Innovative Research Award and IEEE PERCOM Mark Weiser Best Paper Award in 2018.
\vspace{-0.1in}
\end{IEEEbiography}

\begin{IEEEbiography}
[{\includegraphics[width=1in,height=1.25in,clip,keepaspectratio]{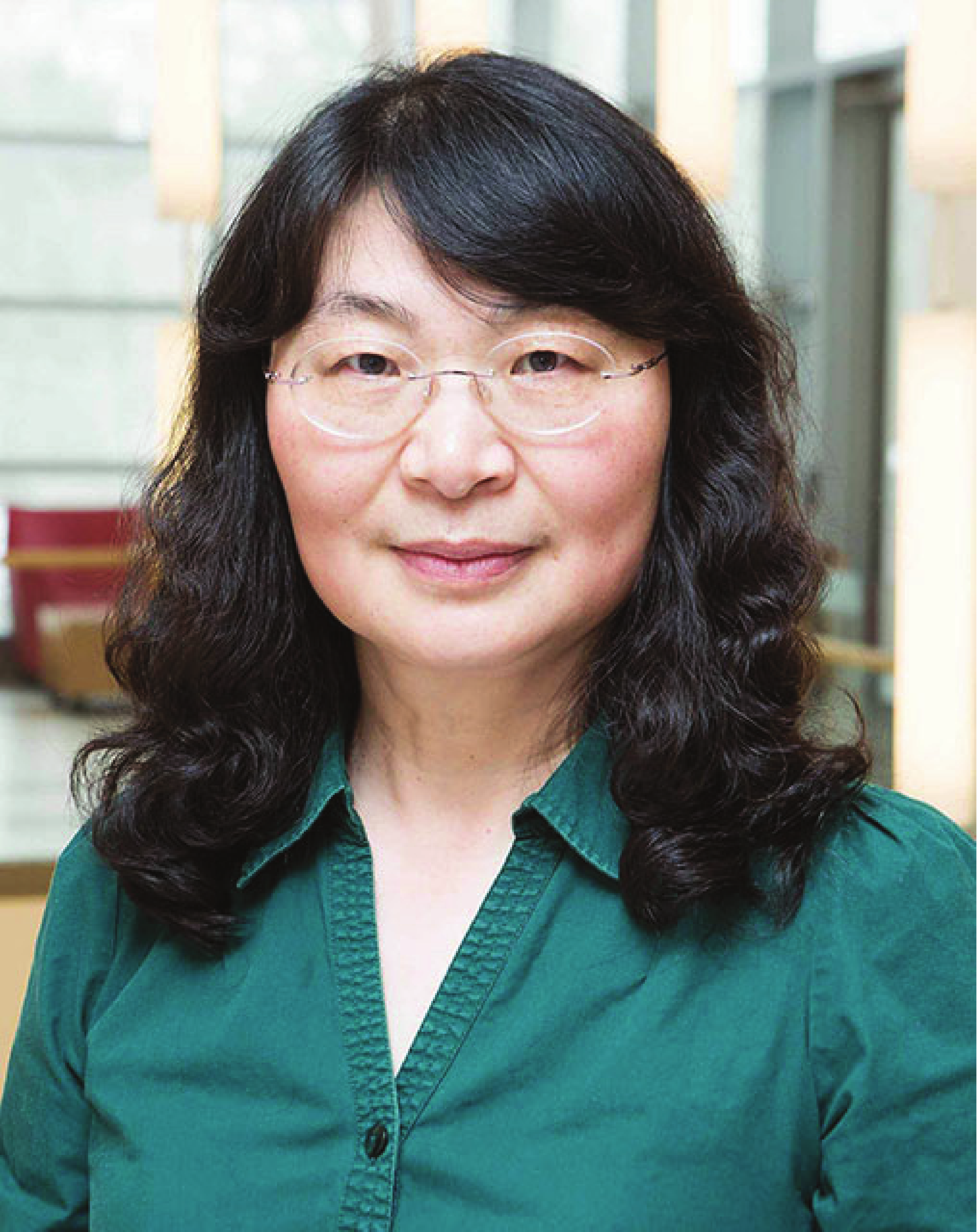}}] {Yuanyuan Yang} received the BEng and MS degrees in computer science and engineering from Tsinghua University, Beijing, China, and the MSE and PhD degrees in computer science from Johns Hopkins University, Baltimore, Maryland. She is a SUNY Distinguished Professor of computer engineering and computer science at Stony Brook University, New York, and is currently on leave at the National Science Foundation as a Program Director. Her research interests include edge computing, data center networks, cloud computing and wireless networks. She has published over 400 papers in major journals and refereed conference proceedings and holds seven US patents in these areas. She is currently the Associate Editor-in-Chief for IEEE Transactions on Cloud Computing and an Associate Editor for ACM Computing Surveys. She has served as an Associate Editor-in-Chief and Associated Editor for IEEE Transactions on Computers and Associate Editor for IEEE Transactions on Parallel and Distributed Systems. She has also served as a general chair, program chair, or vice chair for several major conferences and a program committee member for numerous conferences. She is an IEEE Fellow.
\vspace{-0.1in}
\end{IEEEbiography}

\begin{IEEEbiography}
[{\includegraphics[width=1in,height=1.25in,clip,keepaspectratio]{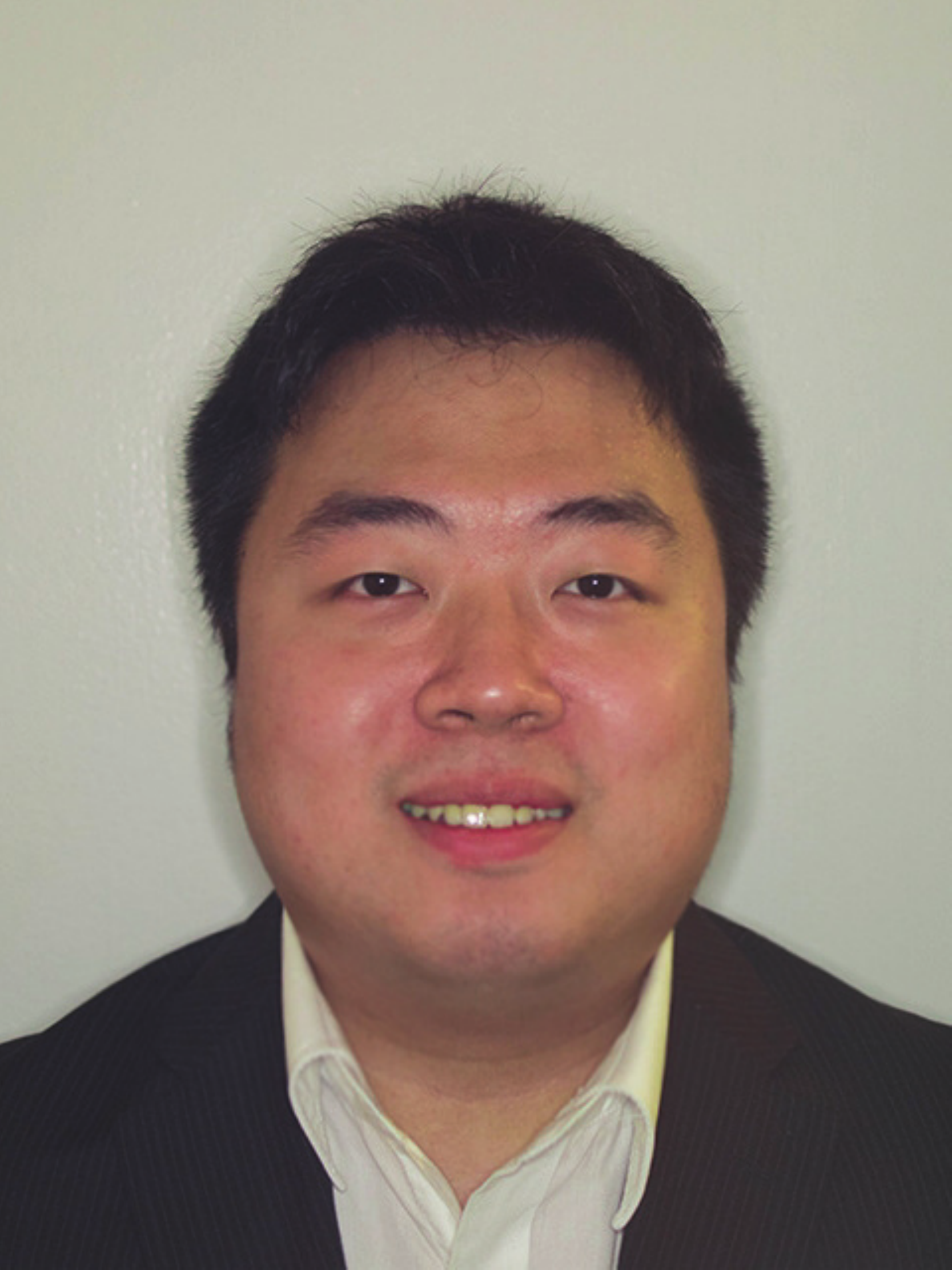}}]
{Pengzhan Zhou} received the Ph.D. degree in Electrical Engineering from the Department of Electrical and Computer Engineering, Stony Brook University, New York in 2020. Before that, he received the B.S. degree in both Applied Physics and Applied Mathematics from Shanghai Jiaotong University, Shanghai, China in 2014. His research interests include wireless sensor networks, performance evaluation of network protocols, algorithms, and artificial intelligence.
\vspace{-0.1in}
\end{IEEEbiography}

\end{document}